# IMPERIAL

# Predicting Extubation Failure in Intensive Care: The Development of a Novel, End-to-End Actionable and Interpretable Prediction System

Author

Akram Yoosoofsah

CID: 01334496

Supervised by

Dr Ovidiu Șerban

Dr Kai Sun

A Thesis submitted in fulfillment of requirements for the degree of
**Master of Science in Computing**

Department of Computing
Imperial College London
2024

# Abstract

**Background:** Predicting extubation failure in Intensive Care is challenging due to the volume and complexity of available data. Machine learning holds promise to revolutionise clinical decision-making; however, current research often overlooks the severe clinical consequences, failing to capture temporal patient trajectories and prioritize model transparency - underscoring the need for further exploration.

**Objectives:** This study aims to develop a novel, end-to-end, actionable and interpretable prediction system for extubation failure, focusing on temporal modelling with the use of both Long Short-Term Memory (LSTM) and Temporal Convolutional Networks (TCN), prioritising interpretability and collaborating closely with clinicians.

**Methods and Data pre-processing:** A retrospective cohort study was carried out on 4,701 patients undergoing mechanical ventilation from the MIMIC-IV database. Time series and static data were collected from the 6 hours before extubation. Features were selected and grouped into sets based on a novel cross-referencing exercise considering clinical relevance (using the WAVE study feature set as a benchmark) and literature popularity. Data was processed using novel techniques to address inconsistency and synthetic data challenges.

**Experimentation and Results:** Experimentation involved the iterative development of LSTM, TCN, and LightGBM models. The initial models trained on dynamic data resampled to the same frequency showed complete bias toward predicting extubation success. This behaviour was further reflected when we implemented advanced hyperparameter tuning, included static data, and tried different model architectures. Data was instead divided into three subsets within each group based on sampling frequency to minimise the impact of synthetic data. The data was processed through a novel fused decision system, which showed improved performance on the vanilla models. Overall model performance was comparable across architectures (AUC-ROC around 0.6 and F1 score consistently less than 0.5), with no clear advantage of architecture, including static data or adding features. Ablation provided insights into feature importance, with the change in AUC-ROC minimal for all features (close to 0 for the majority).

**Discussion:** We developed a novel prediction pipeline, addressing challenges in data pre-processing, model development and interpretation. Strategies were developed to mitigate model bias, including clinician-informed pre-processing, bespoke feature sub-setting and resampling strategy, stratified train/test split and masking. The requirements of the data guided Fused LSTM and TCN model development, which have not been deployed in this field previously, with static data integrated via a Feed-Forward Neural Network. Our results highlighted the challenge of synthetic data handling. Limitations of traditional interpretability methods such as SHAP and LIME were addressed through feature ablation. Although models showed modest predictive power, they were all comparable, demonstrating how synthetic data handling throttles performance regardless of architecture, static context, or the number of features.

**Conclusion:** This thesis provides a robust foundation for future work by bringing to light the influence of synthetic data in predicting extubation failure. While pioneering innovative, clinically relevant approaches to mitigate resulting model bias, it underscores the need for reliable, interpretable models to support clinicians and optimise patient outcomes in the ICU.

# Acknowledgments

In the name of Allah, the Most Gracious, the Most Merciful.

Firstly, I would like to express my sincere gratitude to my supervisors, Dr Ovidiu Șerban and Dr Kai Sun from the Data Science Institute at Imperial College, for their invaluable support, insightful advice and guidance throughout the project. Their expertise in data science and machine learning played a crucial role in ensuring the rigour this task necessitated.

My deepest thanks also go to Dr Mayur Murali and Dr Brij Patel, whose clinical insights were instrumental in aligning this project with the practical needs and standards of the Intensive Care Unit. Their contributions ensured that the research remained relevant to real-world clinical practices.

I am especially grateful to Dr Stephen Rees from Aalborg University for providing the WAVE dataset, which played a pivotal role in shaping the feature selection strategy for this study.

Finally, I would like to thank my family for their unwavering support and encouragement throughout this incredibly rewarding master's programme.

# Contents

# 1

# Introduction

## 1.1 Motivation

Decisions made by clinicians in the ICU are often life-or-death. With the sheer volume and complexity of data gathered in the ICU, clinicians are frequently inundated, and erroneous decisions are common. As such, when a patient's life is at stake, assistive tools are necessary for doctors to make the best-informed decision. This is where data science and machine learning have a vital role to play.

Within the ICU, ventilation management encompasses a multitude of decisions, the most important one being when to remove the patient from the ventilator - known as extubation. The optimal patient outcome is successful liberation from the ventilator, and thus, predicting extubation failure has been a keen focus of machine learning research. Applying machine learning to this area can potentially optimise patient outcomes by assisting clinicians, with whom the final decision ultimately lies.

However, the majority of general studies researching machine learning applications for extubation failure do not recognise the high-stakes nature of the final decision and for whom their assistive technology is intended. Moreover, most studies do not implement relevant model architectures to effectively capture temporal patient trajectories and fail to ensure sufficient model transparency to earn clinician trust, which often hinders clinical adoption.

To the best of our knowledge, only one peer-reviewed paper in this field employs relevant temporal models and focuses on interpretability. Nonetheless, this paper is still quite limited. Thus, there is currently no consensus approach to effectively deploy machine learning for predicting extubation failure, with essential considerations regarding data pre-processing, model design and interpretability not adequately explored.

## 1.2   Aims and Objectives

This thesis aims to explore the development of a novel, end-to-end actionable and interpretable prediction system that effectively leverages time-series data, empowering clinicians to confidently assess extubation failure risk.

Specifically, our objectives are to:

1. **Focus models specifically on predicting extubation failure, distinct from extubation readiness or weaning timing**

2. **Compare the performance of LSTM and TCN models with a non-temporal baseline, introducing TCNs as a novel approach for this prediction**

3. **Prioritise model interpretability for clinical transparency**

4. **Collaborate with clinicians to implement rigorous data pre-processing that preserves patient dynamics for real-world application**

By addressing these objectives, this thesis aims to contribute significantly to current research by advancing the creation of a clinically essential tool to improve patient outcomes.

# 2

# Background and Literature Review

## 2.1 The Opportunity of Data in Critical Care

The digitalisation of the healthcare system has revolutionised medical and clinical research. With the advent of electronic health records (EHRs), we can see a complete digital representation of a patient, from the recording of vital signs to demographic information. This has consequently resulted in an enormous amount of data being available, making the argument for the application of data science practices to healthcare particularly convincing.

This is no more so true than for critical care, which poses a unique challenge for data science. Data in the Intensive Care Unit (ICU) is complex and abundant, but, more importantly, decisions made are incredibly high stakes. Thus, the need for advanced, data-driven analytical tools as part of a clinically integrative approach to support decision-making is pressing.

## 2.2 Respiratory Failure and Mechanical Ventilation

Predicting extubation failure for patients receiving ventilation-based treatment is a critical area that can profoundly be impacted by data science. To fully understand the problem's importance and the need for a data science-driven assistive tool for clinicians, we must first explain the medical context of respiratory failure and mechanical ventilation.

**The need for effective treatment for respiratory failure**

Respiratory failure can be fatal. Azagew *et al.* notably outlined that the incidence of acute respiratory distress syndrome (ARDS – a severe form of respiratory failure) induced by COVID-19 was 32.2% with mortality ranging from 23% to as high as 56% [1]. Furthermore, respiratory failure has been shown to worsen with age; thus, with the World Health Organisation estimating that the number of people aged 60 and older will double by 2050, the advancement of means to assist in treating this condition cannot be ignored [2, 3].

Respiratory failure occurs when a person's lungs fail to perform their essential functions of exchanging oxygen and carbon dioxide (gas exchange) and hence cannot get sufficient oxygen to the blood [4]. What is perhaps most alarming is not only how sudden respiratory failure may transpire but also the profound long-term impact on survivors. While chronic instances exacerbate over time, ARDS can transpire within hours or days and is very difficult to foresee [5]. Moreover, even if patients survive, their livelihoods will likely no longer be the same. Survivors face several long-lasting consequences, including impaired cognitive ability, physical struggles, and not to overlook the emotional impact – all worsening quality of life [6].

**Mechanical ventilation as a cornerstone of treatment**

In current medical practices surrounding respiratory failure, comprehensive treatment management requires a multifaceted approach that tackles the underlying cause and provides supportive care for the patient. Mechanical ventilation is a cornerstone of all forms of treatment, with Matuschak *et al.* concluding that "the only treatment proven to improve survival is mechanical ventilation" [7].

Mechanical ventilation supports the essential task of gas exchange by alleviating the physical work of breathing when a patient's respiratory system is malfunctioning. While not considered a treatment per se, ventilation is a core part of an encompassing treatment plan that allows time for the underlying cause to be addressed [8].

One form of ventilation is positive-pressure ventilation [9]. This delivers oxygen-rich air into the lungs and can be invasive or non-invasive (NIV) [10]. Invasive ventilation is achieved either via the mouth (endotracheal intubation) or through a surgical opening in the neck (tracheostomy) [11]. Once an endotracheal tube is inserted, the patient is connected to the ventilator. Clinical professionals can precisely adjust ventilatory settings to enable adequate gas exchange whilst trying to curtail the risk of potential complications [12].

Clinicians select between control modes, where the ventilator controls breaths, and spontaneous modes, where breaths are initiated by the patient with some support provided by the ventilator. Figure 2.1, from Goligher *et al.* [8], highlights the patient-ventilator interaction. Decisions around ventilator strategy need to be taken with the utmost care, as inappropriate assistance can cause the patient severe distress and exacerbate their condition.

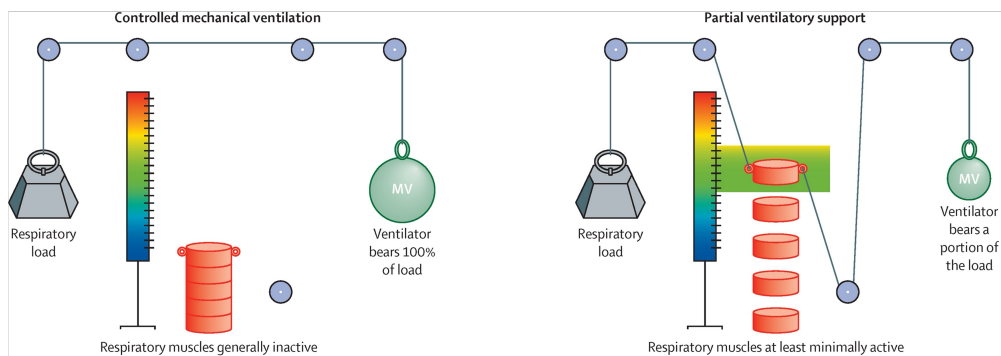

Figure 2.1: Interaction between patient and ventilator. When in control mode, the ventilator bears the entire load, whereas, in support or spontaneous mode, the respiratory muscles are active with the rest of the load taken by the ventilator. Taken from Goligher *et al.* [8].

Remembering the end goal of ventilation is essential – for the patient to no longer need ventilatory support. Ventilation is merely a tool to enable adequate gas exchange. For patients to fully recover, they must be able to carry out gas exchange independent of the ventilator. As the underlying condition begins to heal and respiratory muscle strength is regained, patients can be gradually withdrawn from ventilatory support – a process known as "weaning" [13].

### Liberation from mechanical ventilation

The exact definition of weaning is split into different opinions. Amongst the studies describing weaning, there has yet to be a clear consensus as to when weaning starts. This poses a fundamental problem for clinicians who need to determine when to reduce a patient's reliance on the ventilator. Some studies determine that weaning begins as soon as the patient is intubated. In contrast, others assess it when spontaneous breathing is initiated [13]. It is argued that the term "weaning", while apropos in terms of the reduction in ventilator reliance, is a touch misleading and that "liberation" is likely a more descriptive term – emphasising the complete removal of ventilatory support, which is the optimal patient outcome [14].

### Weaning decision making

Deciding when to start weaning is critical as it is estimated to comprise around 40% of the total duration of mechanical ventilation [9]. Most studies highlight that the primary method of assessing a patient's readiness to start weaning off ventilation is Spontaneous Breathing Trials (SBT) [13].

If an SBT is deemed successful, further factors need to be considered before the endotracheal tube can be removed – also known as "extubation". Such factors include airway secretions, the strength of coughing, and sufficient protection of the airway post-removal [14]. If clinicians deem conditions to be adequate, the endotracheal tube can be removed, otherwise the patient continues under ventilation.

In most ICUs, such decisions are attempted to be standardised through weaning protocols [14]. However, as concluded by Murali *et al.*, significant heterogeneity exists among clinicians regarding weaning decisions, even with current Clinical Decision Support Systems (CDSSs) and protocols [15]. Despite standardised protocols across ICUs, the extubation failure rate persists above 10% and has not significantly improved in the last few decades [16]. When considering the potentially fatal side effects of incorrect decisions, clinicians need to have as much assistive information beyond that of simple protocols.

### Risks associated with extubation

Liberation from the ventilator is one of the most pivotal decisions in the ICU. It impacts the patient's health at the time, and ill-advised decisions could negatively impact the patient for the rest of their life. The primary determinant of weaning success is successful extubation and, crucially, the timing of extubation [17]. Unsuccessful extubation has been reported to increase the likelihood of death seven-fold [18]. Furthermore, extubation failure typically occurs in 10-20% of ventilated patients, even if they are ready to wean [19–21]. As such, with extubation, timing is crucial.

If patients are extubated prematurely, they will need to be reintubated, prolonging their hospital stay unnecessarily. This can lead to complications such as airway loss, respiratory muscle degradation, and defective gas exchange [22]. Reintubation in unsuccessfully weaned patients increases the mortality rate and reduces the chance of patients returning home compared to those successfully weaned [18]. Conversely, prolonged ventilation increases the risk of complications such as pneumonia and diaphragmatic dysfunction, potentially further lengthening ventilator support [23]. These complications, coupled with potentially extended time under ventilation, increase the risk of death, with studies highlighting that the duration of ventilation commensurately increases the risk of extubation failure [24]. Therefore, the importance of determining the optimal timing of extubation cannot be understated. Reducing the duration of ventilation to only what is necessary for the patient is the primary way of suppressing complications [9].

**The need for assistive tools in extubation decisions**

The heterogeneity of clinical practice and patient conditions make clinician decisions significantly challenging. A clinician's judgement and experience are vital for successful extubation [14]. However, this makes the decision inherently subjective, and the clinician is inundated with vast amounts of data from several sources. As such, when a patient's life is at stake, assistive tools are necessary for doctors to make the best-informed decision, particularly when studies have shown that the success rate of purely physician-led weaning is estimated to be below 50% [25].

Clinical protocols attempt to normalise decision-making. However, the variability of ICUs, ventilator design, and patient conditions inhibit protocol generalisation, ultimately falling on the physician's subjective judgement [26]. The notion of personalised mechanical ventilation responsive to each patient's needs has been mooted. Still, with the number of permutations and inherent heterogeneity, there needs to be more high-quality evidence demonstrating its superiority over typical ventilation strategies [8, 27]. Moreover, the cost of implementing individual mechanical ventilation would be significant given the required resource intensiveness and clinical training.

Clinical decision support systems (CDSSs) have been introduced to help clinicians better understand a patient's state [28–31]. CDSSs are digital tools that match patient parameters to a digital clinical knowledge base from which recommendations are given to the clinician [15]. However, in many CDSSs, the clinical database is based on historical data, which limits their ability to inform patients about subtle changes in patient condition as the system does not learn from new patterns [32]. More importantly, doctors must trust the system, as with any technology adopted into medicine. Many CDSSs provide recommendations with little explanation of their reasoning, and the lack of transparency perturbs clinical adoption [33].

The need for assistive tools for weaning and extubation decisions is clear. While helpful, traditional protocols and CDSSs cannot capture the complexity of patient states, and the lack of transparency and intelligence hinders adoption. This paves the way for computationally intelligent, predictive models deployed in machine learning (ML) practices to make a meaningful impact, analysing large amounts of data and providing objective insights to bridge gaps in current clinical practice. However, like any technology, these tools will need to earn the physician's trust, with whom the clinical decision ultimately lies.

## 2.3   Machine learning in ventilation management

In the ICU, timely and accurate decisions are paramount. Unfortunately, clinicians face significant pressure, and erroneous decisions are common [34]. A study by Donchin *et al.* [35] reported an estimated 1.7 errors per patient per day in an ICU observed over four months. Moreover, the sheer volume and complexity of data, estimated to be 1,300 data points generated daily per ICU patient (10 times more than in any other hospital setting), further compounds the complex decision-making [36]. Consequently, healthcare professionals would be remiss in overlooking the need for advanced analytical tools, namely machine learning, to assist in timely and accurate decision-making in the ICU.

Ventilation management encompasses a multitude of decisions, from initiation to adjusting ventilator settings in response to the patient's rapidly changing condition [37]. With the high cost of making the wrong decision, several studies have naturally focused on employing ML in this field. For instance, Suo *et al.* demonstrated the use of ML to create an adaptive controller for mechanical ventilators, which was found to control the ventilator more accurately than the standard control feedback mechanism, potentially reducing the risk of patient injury [38].

Machine learning has similarly proven its utility in predicting ventilation duration [39] and even in predicting the need for a patient to require invasive mechanical ventilation hours in advance [40]. However, among the many critical decisions in this field, deciding when to wean the patient from ventilatory support and initiate extubation arguably holds the greatest significance. As previously stressed, this decision impacts the patient's immediate survival and long-term quality of life. The importance of determining the optimal timing of extubation cannot be understated, and the multi-faceted nature of this decision necessitates advanced assistive support, namely from machine learning.

**Machine learning for the prediction of weaning and extubation**

Several studies have investigated using machine learning to improve weaning outcomes in the ICU [26]. The distinguishing factor of these studies is the specific objective the models are designed to predict. When reading the "fine print", so to speak, several studies focus on predicting extubation readiness and not necessarily extubation success.

It is, therefore, essential to distinguish between extubation success and extubation readiness. Extubation readiness refers to the clinical evaluation before removing the endotracheal tube [41]. On the other hand, extubation success refers to the ultimate removal of the endotracheal tube, signifying liberation from ventilatory support [42].

The optimal patient outcome is successful liberation from the ventilator [14]. Thus, the focus regarding extubation studies hereinafter will be on literature concentrating on extubation success/failure as the objective of ML models rather than extubation readiness.

## 2.4 Literature Review - Learning from studies focusing on predicting extubation failure

Despite its clinical significance, research on predicting extubation failure remains relatively niche. We present a targeted review of key literature, excluding studies before 2015 due to advancements in ventilator management, machine learning technology and clinical guidelines, rendering comparison obsolete. This aligns with the documented evolving field of mechanical ventilation strategies [43].

The studies considered in this review explicitly state that the focus of model development was to predict extubation failure/success; hence, studies where this was not a stated objective were not considered. Furthermore, papers that collected data on neonates/infants, non-ICU conditions, and non-invasive ventilation were also excluded to ensure relevance to this study.

It should be noted that the studies (summarised in Appendix Table A.1) reviewed here are not exhaustive but offer a representative overview of typical studies in the field of predicting extubation failure and provide a foundation for the understanding of current research. These studies do not explicitly focus on time-series prediction or model interpretability but, nonetheless, contribute valuable insights and define a basis for more specialised work in those areas which will be reviewed later.

We organise these studies by their modelling approach, distinguishing between the development of single or multiple models. This cataloguing highlights the diversity of methodologies and facilitates more meaningful comparisons.

Four of the eight studies focus on developing a single ML model, and two aim to design artificial neural networks (ANN), with the other two investigating Gradient Boosting Frameworks (GBFs). The power of ANNs lies in their ability to comprehend complex data patterns (such as those inherent in healthcare data) and provide an optimal prediction. However, the primary downfall of ANNs is their main strength - their complexity. ANN training involves complex mathematical calculations, which are challenging for a human to interpret. This evokes the "black box" sentiment typically ascribed to Artificial Intelligence, a challenge any study looking into ANN architecture must address.

Kuo *et al.* [44] and Hsieh *et al.* [45] focused on designing ANN models for binary classification. Kuo *et al.*'s investigation predicted on a limited dataset (n=121), of whom 26% failed extubation. A literature review was used to select what they deemed relevant features, from which they narrowed to eight. The final model comprised 17 input neurons, a single hidden layer of 19 and 2 output neurons. The authors then carried out internal validation of their model, comparing its performance to traditional clinical predictors, including variations on RSBI (rapid shallow breathing index) and maximum inspiratory pressure against which it outperformed on an AUC-ROC (area under the receiver operating characteristic curve) basis [46]. AUC-ROC is a standard performance measure used in ML studies that evaluates independent of a decision threshold [47]. Markedly, no validation was undertaken against other ML architectures or externally with other ICUs or patient databases.

The work by Hsieh *et al.* [45] followed similar procedures. The clinical dataset comprised 3,602 patients with planned extubation, significantly more than in Kuo *et al.* The proportion of patients who had failed extubation was 5.1%, lower than the reported general range of 10-20% [14]. A final selection of 37 input features comprising a mix of clinical parameters and demographic information was used to train the model. However, they do not explicitly divulge how they came to select these features, stating they are chosen based on their "wide availability in ICUs," which is inherently vague. Validation in this study was also only carried out internally, with performance compared to individual predictors including RSBI, TISS (Therapeutic Intervention Scoring System – the number of intensive care treatments given to patients) and maximum expiratory pressure, against which it outperformed on an AUC-ROC basis [48]. Once again, no external validation was carried out. While the AUC-ROC reported by Hsieh *et al.* is higher than that in Kuo *et al.* (0.85 vs 0.83), a direct comparison is futile, given the differences in sample size and clinical settings.

Zhao *et al.* [49] and Chen *et al.* [50] interestingly decided to abstain from traditional ANNs and explore gradient-boosting frameworks for predicting extubation success/failure. Gradient boosting frameworks (GBF) implement a gradient-boosting machine learning algorithm. They create multiple weak learning models (simple models that perform slightly better than random chance) where new models are iteratively targeted to correct the error in the prediction of the previous one [50]. Further intricate detail is beyond the scope of this study.

Zhao *et al.* [49] investigated the development of a CatBoost (Categorical Boosting) model. Cat-Boost is a GBF algorithm, and its power lies in automatically pre-processing categorical features without encoding [51]. Unlike previous studies, the MIMIC-IV database was used to extract data from 16,189 patients, of whom 17% failed extubation. The authors subsequently extracted a subset of these patients based on set exclusion criteria. Unlike the previous studies, rigorous feature selection was carried out and notably well-documented. Recursive feature elimination (RFE) was employed based on SHapley Additive exPlanations (SHAP) values. SHAP is a technique used in ML to improve interpretability and visualisation [52]. It is important to note that interpretability was not the focus of this study and is more of an addendum. Eighty-nine features were used to develop the primary model, with 19 selected for a more compact version. The final model was then compared against ten other ML models, primarily on an AUC-ROC basis and outperformed. This study also impressively carried out external validation on 502 patients in Zhongshan Hospital in China.

Additionally, the model was fascinatingly integrated into a web-based tool and deployed. Clinicians could enter the 19 patient features at a specific time, and a prediction and the SHAP values of the top 10 most influential features are provided for interpretation. This is the only study amongst the eight reviewed here where clinical deployment was actioned, outlining how rare clinical deployment is.

Chen *et al.* [50] took a slightly different approach, focusing on developing a Light Gradient Boosting Machine (LightGBM) model to predict extubation failure. LightGBM is an alternative GBF algorithm that has been reported to increase the speed of the training process by almost 20 times while maintaining accuracy [53]. Like Zhao *et al.*, this study uses patient data from the MIMIC database (MIMIC-III) and extracted 3,636 patients based on set exclusion criteria, of whom 17% failed extubation. Initially, 92 features were processed based on correlation analysis to avoid highly correlated features, resulting in a final set of 68 features. Feature importance analysis was invoked to identify the 36 most vital features by computing the number of times the feature is used in the model to make a prediction. For better interpretability (again, not the focus of this study), SHAP values were used to corroborate whether these features positively or negatively correlated with extubation failure. Model performance with the top 36 features is comparable to that with 68 features (AUC-ROC of 0.8918 vs. 0.8130, respectively), implying feature extraction was meaningful. The final model was also compared with the performance of more traditional ML architectures, including logistic regression (LR), support vector machine (SVM), and ANN but it should be noted that these come from external studies that are not based on the same dataset or clinical settings, weakening the comparison.

The remaining four studies focused on building multiple models employing varied ML architectures and selecting the highest-performing one.

Febregat *et al.* [54] compared SVM, Gradient Boosting and Linear Discriminant Analysis (LDA) models. The intricacies of these architectures are beyond the scope of this study. The authors gathered data from 1,108 patients, of whom 9% failed extubation. Twenty features were chosen, but the selection rationale was not clear. Patient data was reported to be erroneous with incorrect time stamps, and thus, extensive pre-processing was required. Only legitimate extubations were extracted by validating ventilator readings with medical records to identify each extubation event. This narrowed down the working dataset to a relatively small 697 patients. After training, SVM displayed the best-reported performance on an AUC-ROC basis.

Otaguro *et al.* [42] and Huang *et al.* [55] followed suit by developing and testing three ML models.

Otaguro *et al.* sought to develop Random Forest, XGBoost and LightGBM models. Data was taken from 117 patients, of whom 11% failed extubation. The authors chose 58 features to train the model based on those available in electronic health records. LightGBM was shown to have the highest AUC-ROC. Compared to the LightGBM model developed by Chen *et al.*, Otaguro *et al.*'s model had superior performance on an AUC-ROC basis (0.9502 vs. 0.8198, respectively). Feature importance analysis revealed that the most crucial feature was time under mechanical ventilation, however, the lack of SHAP values hinders complete comprehension of this importance.

Fleuren *et al.* [56] took a slightly different angle by focusing on extubation failure within COVID-19 patients. From an initial 2,421 patients, 863 formed the working dataset after a number had been excluded for data quality and intubation criteria. In contrast to all studies presented, a broad list of features was initially gathered from the literature. Qualified physicians assisted in feature selection to ensure clinical relevance (resulting in a final list of 40). LR, decision trees and XGBoost models were trained with the ease of identifying feature importance in mind. The XGBoost algorithm had the highest AUC-ROC, with SHAP values subsequently generated to highlight the importance. While this is one of the first studies of its kind to predict extubation failure in COVID-19 patients across several ICUs, the model's performance cannot be compared with the other seven studies as it only involved patients suffering from COVID-19; thus, patient conditions are inherently different. External validation was also not performed in this study.

Finally, Huang *et al.* [55] trained logistic regression (LR), random forest (RF) and SVM models on data taken from 233 patients, of whom 12% failed extubation. Ventilator parameters were processed into datasets per 1, 30, 60, 120, 180 and 300 seconds, and RFE was employed to identify the most essential features. Six of the twelve features were optimal in the 180s split dataset and used to train the models. The RF was seen to be the model with the highest AUC-ROC. Alluringly, with the variables sampled every 180 seconds, the model can predict extubation success every 3 minutes, making it more dynamic compared to all studies evaluated thus far. SHAP values were used more for statistical analysis and internal interpretation rather than clinical explainability. The authors' preprocessing step of using "the averaging method" effectively transforms the data into aggregated features, potentially obscuring true temporal dynamics [55]. Consequently, there is conjecture as to whether their models best represent the inherent time-dependent patterns in the data.

## 2.5  Evaluation of general studies focusing on predicting extubation success

The reviewed studies must not be taken at face value. Here, we outline a summary evaluation based on key points considered when deploying ML in the medical field. It should be noted, however, that these studies are all based on different patient datasets and study settings, so comparing model performance is not valid. However, the author's choices and study design can certainly be scrutinised with clinical relevance in mind.

**Features**

Selecting the right features is crucial in machine learning. Excessive features can lead to overfitting, preventing generalisation. Conversely, too few or uninformative features can lead to underfitting [57]. There is also a trade-off between model accuracy and interpretability [58]. The more features used to train the model, the more complex it is to identify the critical drivers of prediction.

The number of features used to develop their models differs significantly in the above studies. Zhao *et al.* [49] used 89 features to build their primary CatBoost model, the highest of all the studies presented in this report. At the other end of the spectrum, Huang *et al.* [55] and Kuo *et al.* [44] only used 6 and 8, respectively. Excessive features complicate interpretability, undermining clinician trust [59]. Interestingly, it seems the authors recognised this, as they chose to deploy their compact model (based on 19 features) with comparable performance in their web-based tool rather than their extensive 89-feature model, implying there were several irrelevant features initially.

Some studies employed explicit feature selection approaches (Zhao *et al.*, Huang *et al.* and Chen *et al.*) [49, 50, 60]. Both RFE and correlation analysis used alongside SHAP in these studies served as systematic, objective methods to narrow feature selection and help visualise model rationale. This methodology serves a twofold benefit of simplifying data collection to only the most influential features and improves interpretability. Such approaches signify the path to attaining clinicians' confidence, significantly more so than other studies that are vague in their selection process, such as those by Hsieh *et al.*

Ultimately, clinicians must be involved in feature selection, as Flueren *et al.* [56] have notably done. Combining robust feature selection methods with clinical expertise would provide a balance between subjectivity and objectivity, making data pre-processing more rigorous.

**Data sources and imbalance**

Data diversity, quality and balance, impact the accuracy and generalisability of ML models [61]. While studies using large databases such as MIMIC benefited from extensive patient data, they also faced challenges with noise and required more meticulous pre-processing. Notably, the data collected by Fabregat *et al.* required extensive pre-processing to handle erroneous values. Contrastingly, data based on clinical trials (as employed in Kuo *et al.*, Hsieh *et al.*, Fabregat *et al.*, Otaguro *et al.*, and Huang *et al.*) often had smaller, less diverse samples, limiting model generalisability. Focusing on a smaller cohort may have been symptomatic of patient availability at the time, but the amount of data is insufficient from an ML standpoint.

Data imbalance poses another challenge. Predictions will likely be skewed towards the majority class, rendering the model obsolete when implemented on unseen data. The dataset used by Hsieh *et al.* only had a 5% extubation failure rate. As such, any prediction made by their model would naturally be biased towards extubation success, which is unacceptable for clinical deployment. Unsurprisingly, the authors did not publish an accuracy measure recognising the severity of the imbalance. They also consequently state that their findings "may not be generalisable to other ICUs," frustratingly defeating the end goal of general clinical adoption and earning clinician trust. Mitigation is possible through oversampling (the minority class) or undersampling (the majority class), which Hsieh *et al.* surprisingly did not do.

Although oversampling and undersampling can reduce imbalance, their effectiveness varies, and their absence is particularly concerning. Chen *et al.* tried to employ SMOTE (synthetic minority oversampling) to increase the incidences of extubation failure and mitigate this. Still, results with or without were not noticeable and thus not included in that study.

Studies are ostensibly as good as their data, but authors have little control over the initial failure rate in the dataset. However, their choice of data source, clinical trials or large databases, is within their control and a vital aspect of methodological design. Appropriate techniques must be employed to deal with data errors and imbalances to ensure model generalisability, highlighting the importance of pre-processing.

**Data annotation**

Supervised ML models need representative ground truth values. In these studies, the label was either extubation success or failure. However, studies seem to have varied definitions of extubation failure. Kuo *et al.*, Chen *et al.*, Fabregat *et al.*, Zhao *et al.*, Fleuren *et al.*, and Huang *et al.* used a 48-hour reintubation window and others a 72-hour window instead. As mentioned, extubation success remains an ill-defined term, leaving researchers needing to decide. Most studies tend to define failed ventilator liberation as requiring reintubation within 24-72 hours; thus, the studies presented here are within the range typically used in the literature [18, 62, 63].

This highlights the need for a consensus definition if studies in the field are to improve iteratively and valid comparisons are to be drawn. Decisions made earlier may not capture late extubation failures, prematurely skewing the dataset towards success as a majority class and fuelling the inherent data imbalance. This would subsequently exacerbate model generalisability. Additional clinician guidance would be invaluable in facilitating a consistent definition, which would help model predictions be more reliable for clinical adoption.

**Model validation**

Interestingly, only Zhao *et al.* validated their model externally. Their compact model was deployed on prospective validation data from over 500 patients and performed comparatively to their internal validation set, encouragingly implying generalisability. Disappointingly, no other study employed external validation. Comparisons were made to other models developed by the authors or similar models' results in the literature. Internal comparison against other architectures does promote the performance of the model of choice; however, it fails to prove that the model is generalisable, with half the studies stating explicitly that they cannot be sure that their model can be applied elsewhere – namely, Hsieh *et al.*, Otaguro *et al.*, Fabregat *et al.* and Kuo *et al.* Robust external validation in any study looking into extubation failure is critical to establishing model generalisability and clinical efficacy, which should be standard practice in the future.

**Chosen model architecture**

The choice of model architecture is fundamental. If the wrong architecture is chosen, models will struggle to derive meaningful insights, impinging predictive capabilities. Researchers need to strike a balance between complexity, relevance to the data and interpretability to even give their model a fighting chance of clinical adoption.

Complex architectures, such as ANN, do not lend themselves to ease of interpretation. Simpler models are often more interpretable but could lack predictive power, whereas more complex models can learn patterns astutely but not be entirely understood by those using them. Feature importance in tree-based algorithms (such as GBF and RF) can be easily visualised with SHAP plots (as Zhao *et al.*, Chen *et al.*, Fleuren *et al.* and Huang *et al.* have done). On the other hand, Otaguro *et al.* carried out feature analysis but did not provide SHAP plots. Thus, whether these features positively or negatively impact extubation failure is unknown, with the authors explicitly stating, "It was unclear how some of the important features identified affect decision making" [42].

The lack of transparency creates a high barrier to clinical deployment, not to mention gaining clinician trust. As quoted in a study by Ennab *et al.* [64], "The lack of interpretability in artificial intelligence models (i.e., deep learning, machine learning, and rules-based) is an obstacle to their widespread adoption in the healthcare domain". While these models should never form the final decision, clinicians must utilise assistive tools such as these to assimilate multi-modal data to corroborate their own experience and reach a conclusion. Ill-informed choices will likely have catastrophic outcomes. As such, any study looking to create a clinically relevant model must prioritise employing model interpretability techniques.

Achieving both interpretability and accurate predictions necessitates careful consideration of the model architecture in relation to the data. Different data types compel different architectures designed to capture the intended types of patterns and produce the desired predictive output. Healthcare data comprises a heterogeneous mix of data types, from demographic data such as age and gender to temporal data such as heart rate readings. In the ICU, patient trajectories are one of the most important indicators used when making clinical decisions and are highly predictive of the patient's future course, helping dictate effective care [65]. Temporal data analysis is essential to capture patient trajectories effectively. Hence, in any study deploying machine learning for decisions in the ICU, a relevant architecture must be chosen that can learn from temporal dependencies.

The architectures used in these more general studies are not designed to learn from long-term dependencies that would naturally capture patient trajectories. Surprisingly, therefore, these general studies fail to recognise the importance of the decisions they are building their models to assist with. Patient vital signs and ventilator parameters are inherently longitudinal and continuously monitored, patterns these models would fail to capture.

The choice of model architecture is paramount in any well-founded machine learning study. An incorrect choice can thoroughly undermine an investigation, as the authors need to balance model performance, interpretability, and computational feasibility in the context to which the study is applied.

To be considered clinically relevant, competent models must be able to learn and predict from temporally dependent patient trajectories and be easily interpretable. These requirements stem from the choice of architecture.

**Conclusions from general studies on extubation failure**

It is disappointing that most studies fail to recognise the real-world end goal of their work and for whom the tool is intended. In the ICU, clinicians must make decisions that could have fatal consequences, and any assistive tool must be trustworthy and relevant. To capture this, models must be able to learn dependencies from temporal data and be easily explainable. If clinicians cannot understand why the model has reached a particular prediction, they will never be able to use it confidently.

As such, the explored studies may be interpreted more as informative technical studies than viable for medical adoption. While they may focus individually on making time-series-based decisions or have elements of explainability, none capture both with the imperativeness or effectiveness required.

Any competent study must ensure clinical relevance by involving clinicians from the start and designing their model to be dynamic and transparent so that it can seamlessly be incorporated into ICU practices and optimise patient outcomes. Bespoke model architectural choices, a focus on interpretability and clear, rigorous data pre-processing are unquestionable requirements for clinical adoption.

## 2.6 Predicting extubation failure from temporal dependencies

As has been established, the gap in "typical" studies employing machine learning to predict extubation success/failure lies in the lack of use of models that can interpret temporal dependencies. Now, we can look beyond these general studies and identify how research has tackled this issue.

Time series data has an inherent temporal dependence. A value at a given time point for a feature is typically influenced by its past values, creating an ordered sequential relationship [66]. This data exhibits trends and patterns that traditional ML models used in the aforementioned studies cannot handle, as a critical assumption typically used is that training and evaluation data are independent and identically distributed (IID), i.e. the value of one data point does not influence the value of other points [67]. Time-series data required to effectively predict extubation failure violates this assumption; hence, more specialised model architectures are needed.

Several architectures and commensurate studies have been designed to capture temporal dependencies across various domains of machine learning research, including Recurrent Neural Networks (RNN) and their variant architectures, Temporal Convolutional Networks (TCNs), Transformer-based Models such as Temporal Fusion Transformers (TFTs) and Autoregressive Integrated Moving Average (ARIMA) models [68]. Each has its nuances; however, to remain within the scope of this study, we will focus on reviewing those specifically deployed to predict extubation outcomes thus far (RNN variants and TCNs) and then narrow down to extubation failure. Given that these papers present the most relevant studies regarding the architectures used for this project, we will evaluate both the model choices and their approach to data preprocessing, which can significantly impact model performance.

As a brief introduction, RNNs are a specialised form of ANN designed to predict sequential data. Proposed in the 1980s and 90s, early models struggled with the "vanishing gradient problem" [69]. This led to the development of Long Short-Term Memory (LSTM) networks by Hochreiter & Schmidhuber in 1997 [70]. LSTMs have revolutionised several domains, from natural-language processing [71] to speech recognition [72]. A simpler variant of LSTMs known as Gated Recurrent Unit (GRU) was subsequently introduced by Cho *et al.* in 2014 [73]. RNNs versatility in handling sequential data is highlighted in its application to many fields, such as financial forecasting [74] and, naturally, healthcare.

A much more recent alternative is Temporal Convolutional Networks (TCN) created by Lea *et al.* in 2016 [75] based on Convolutional Neural Networks (CNNs). TCNs apply convolutional layers to extract patterns from time-series data and offer computational advantages over LSTMs over longer sequences, as convolutional layers can be processed in parallel [76].

Most work in extubation outcome prediction has been carried out over the last decade and even prior, but only in recent years have specialised architectures designed for temporal dependencies come to the fore. These studies represent cutting-edge research in this domain, leveraging sophisticated machine-learning techniques to address the intricacy of temporal data when predicting extubation failure.

The work by Catling *et al.* in 2020 [77] is a seminal paper that analyses multiple architectures capable of predicting based on longitudinal data concerning extubation and the ICU. The authors sought to develop a TCN model to learn from longitudinal data, fused with a Feed-Forward Neural Network (FFNN – a form of ANN) to encode demographic information. Data was collected over four years from 4,713 ICU patients in the UK. The study aimed to predict specific events in the following 1-6 hours for every hour of patient admission, depending on the event. Extubation was predicted within 6 hours. Despite the objective not specifically being extubation failure but rather the occurrence of extubation, given the minimal number of papers in this field, the evaluation of this study serves as essential context to this project. Hence, we will focus on the results surrounding extubation prediction. TCN-FFNN model performance was compared with an LSTM-FFNN model and baseline LR and FFNN models, which can only predict based on aggregated data. Fascinatingly, this paper demonstrates the effectiveness of hybrid models, creating a bespoke architecture that handles time-series data through the TCN/LSTM and static data through the FFNN, a novel use not seen in typical studies in the field. This would facilitate the best representation of the ICU data, enabling more relevant patterns to be learnt. It should also be noted that this study presents a novel data processing strategy. The authors recognised that imputing missing values to fill in data can be erroneous and, hence, created a "missingness indicator" to mask missing values, preventing the models from learning from missing data typically present in clinical recordings [77].

On an AUC-ROC basis, the TCN-FFNN and LSTM-FFNN outperformed the baseline models for extubation (0.903 and 0.893 vs. 0.859 and 0.797 respectively). This highlights their innate ability to capture long-term temporal dependencies, hence patient trajectories better, facilitating better predictive capabilities. Interestingly, the TCN-FFNN had slightly better performance than the LSTM-FFNN, but the difference is insignificant to deem superior. The authors quote, "a predicted extubation is more likely to be correct when made in the context of stable or improving respiratory parameters rather than a snapshot of these parameters at a single point in time" [77]. This indicates TCNs' and LSTMs' significant potential in predicting extubation events over traditional non-temporal models reviewed in typical studies.

While this study is undoubtedly encouraging, its limitations need to be addressed. Firstly, this paper has little clinical validation, if any. Decisions were made without consultation with ICU clinicians; hence, complete clinical relevance cannot be guaranteed. Furthermore, and likely due to the lack of clinical intervention, the specific objectives of model prediction are arguably irrelevant. Focusing on extubation, the study uses all extubation events as positive labels, regardless of whether the outcome was desired or deemed successful. There may be a case where extubation in the next 6 hours occurs because the patient has successfully been weaned.

Conversely, extubation may be carried out but unsuccessfully. Both events would be in the positive class but clinically be completely different situations, so the model cannot distinguish between good and bad clinical practice. As such, this study, akin to the papers reviewed earlier, fails to recognise the gravity of decisions made in the ICU. Knowing whether extubation will be successful is more important than knowing if extubation will occur, as clinicians can take appropriate precautions. This limits the relevancy of this paper, a fact which the authors admit, "However, it may increase the clinical utility of TCN-FFNN to explicitly limit positive labels to interventions deemed successful or appropriate" [77].

Moreover, interpretability is not one of the objectives of this study. The authors employ Partial Dependence Plots (PDPs) to understand the predictions' determinants. Still, these might be challenging for clinicians to interpret, given the technique's reliance on rather complex statistics and less intuitive visualisation [78]. Overall, this study presents an initial view into the potential TCNs and LSTMs have for predicting using time-series data in the field. Still, it would have benefitted from clinical validation throughout and more refined prediction events.

Now, focusing on predicting extubation success/failure, some attention should be given to the recent paper by Fenske *et al.* [79]. The authors developed LSTMs and RNNs, among other models, to predict extubation success the next day. Interestingly, data from midnight to 8 AM was aggregated to predict next-day success rather than predicting extubation success at the end of a specified period. The LSTM model performed the best on both internal and external datasets. A commendable feature of this study is the involvement of clinicians throughout. Physicians manually reviewed the hospital data to ensure the data was accurate. Moreover, the research group are experienced in annotating clinical data, which underwent extensive quality control. However, since this paper is a pre-print and has yet to undergo the rigorous scrutiny of peer review, it will be considered with some reservation.

Zeng *et al.* [80] published a paper in 2022 developing interpretable RNNs to predict extubation failure. They claimed to be the "first research to develop a prediction model for dynamically predicting extubation risk" [80]. Immediately, this study meets the requirements of focusing on both dynamic predictions using a relevant, specialised architecture and interoperability, marking a significant advancement in extubation failure prediction. This study developed LSTM and GRU models to predict extubation failure trained on 8,599 patients extracted from the MIMIC-IV database. Data was built to 4-hour windows, with 89 features selected (20 static and 69 dynamic) that had an observation frequency greater than 0.1 per 4 h. Numerical outliers were removed using "reasonable ranges". Unlike in the study by Catling *et al.*, missing values were not masked but rather imputed by carrying the last observation forward and filling any resultant missing values with the mean of the training set. Padding and packing were employed to atone for variety in sequence length between patients (as is common in healthcare data), ensuring padded values are not learnt by the model [81].

The model hyperparameters were tuned using five-fold cross-validation, and performance compared to baseline non-temporal models (LR, SVM, MLP, RF and XGBoost). The models were trained to predict the risk of extubation failure between a scale of 0 and 1 every 4 hours, with the output being the extubation failure risk. On an AUC-ROC basis, the RNNs (0.828 for LSTM and 0.829 for GRU) performed better than the LR (0.814), SVM (0.816), and MLP (0.812) but were comparable to RF (0.820) and XGBoost (0.823) [80]. This evidence is inconclusive as to whether the RNNs can capture the inherent patterns indicative of extubation failure compared to traditional models.

Zeng *et al.*'s work is inherently successful in utilising RNNs to capture temporal dependencies to predict extubation failure, and they are seemingly the first research group to do so. They impressively created models that could make a prediction every four hours using a comprehensive set of features (both static and dynamic). Furthermore, they employed SHAP to provide insights into the most impactful features and should be praised for recognising the importance of model transparency and interpretability for clinical adoption. However, several limitations need to be addressed. Firstly, we will start with data pre-processing. Data was grouped into 4-hour windows to balance data size and model updates. Taking data every 4 hours over the patient's stay is not necessarily a clinically relevant strategy, and clinicians should have been consulted when making this decision. Moreover, the data imputation strategy is questionable. Using the "last observation carried forward" method as a primary means can introduce bias and fail to capture the inherent temporal variability as several values will be the same over time, despite them setting a frequency threshold to limit data imputation, with a paper by Lachin (2015) stating that this method "should not be employed in any analyses" for these very reasons [82].

Focusing on model development and validation, LSTMs and GRUs are related architectures, with studies exhibiting comparable performance [83]. This similarity is evident here, with no significant difference between the models' performance. This raises the question of whether investigating two models with related architecture and precedented similar performance may be redundant. Instead, exploring a more diverse range of architectures may yield more insight into the strengths and weaknesses of various approaches. Outcome labelling also comes into consideration as only the last feature vector of a patient can be labelled since the outcome of extubation (success or failure) is only known after the event. This means that earlier feature vectors are not directly supervised, which could result in less precise intermediate predictions as the model does not receive explicit feedback at each 4-hour window before the last one. As such, capturing intermediate patient trajectories could be more challenging for the models. Finally, there is no significant performance difference between RNNs, RF, and XGBoost. It should be noted, however, that the baseline models were trained on data taken from the last 4 hours before extubation and not dynamically throughout the patient's IMV stay; hence, they do not have the benefit of providing real-time updates to clinicians and being continuously refined as data becomes available.

Despite these limitations, Zeng *et al.* and Catling *et al.*'s work (and, to a reserved extent, Fenske *et al.*'s work) represent a pivotal step in utilising temporal modelling for extubation failure and ICU event prediction, demonstrating the potential impact of specialised architectures designed to capture temporal dependencies. However, the identified gaps in both specialised work and general studies cement a pressing need for further research that addresses the precise challenges of predicting extubation failure using time-series data, which are as follows:

- **Explicit focus on extubation failure:** Many studies focus on weaning, extubation readiness and predicting the extubation event, but not all specifically focus on the crucial outcome of extubation success/failure – with the optimal patient outcome being successful liberation from the ventilator [14]

- **Exploration of diverse temporal architectures:** RNNs are an invaluable architecture for temporal modelling. However, the field would certainly benefit from exploring other specialised architectures used for time-series prediction, such as TCNs, for a more comprehensive comparison

- **Prioritisation of interpretability:** There is no question that model transparency is a requirement for clinical adoption. Most studies do not typically focus on interpretability and, therefore, would struggle to be considered clinically deployable. Interpretability is usually an addendum rather than a primary objective. A more profound emphasis on model explainability is crucial so clinicians can make the best-informed decision, usually with significant consequences

- **Clinically relevant and rigorous data pre-processing:** Data pre-processing can be a crucial determinant of model performance. Choices made during these stages must be made with clinical consultation to ensure the final prediction is clinically relevant and aids in models' applicability in real-world clinical settings. It has been reported that the majority of ICU AI models remain in testing and prototyping, with few being deployed clinically, highlighting the lack of evidence supporting clinical values in ML studies [84]

## 2.7 Objectives of this thesis

In light of these gaps, this study aims to explore the development of an end-to-end actionable and interpretable prediction system that assists clinicians in making decisions on extubation failure from time-series data.  Specifically, we will:

1. **Focus on extubation failure:** This study will explicitly tailor the objective of model prediction to be extubation failure, separating it from studies focusing on extubation readiness or timing of weaning

2. **Investigate diverse temporal architectures:** We will develop and compare the performance of LSTM and TCN models alongside a non-temporal baseline model.  While LSTMs have been deployed, TCNs have yet to be used to predict extubation failure, creating a novel avenue of comparison

3. **Prioritise interpretability:** We will employ relevant explainability techniques to ensure the work carried out in this study is transparent for the intended clinical use, ensuring clinicians review any results for relevant insight

4. **Implement clinically relevant and rigorous data preprocessing:**  We will closely collaborate with clinicians at all stages of the study, ensuring data processing and all relevant choices preserve the temporal dynamics of patient data and are as suitable as possible for real-world clinical environments

By fulfilling these objectives and addressing these gaps, we aspire to significantly contribute to the current literature and help develop a clinically relevant tool to aid clinicians in making more informed decisions that could have potentially fatal consequences, ultimately leading to improved patient outcomes.

## 2.8   Additional Background

Here, we will provide more detail about the specialised architectures used in this project: LSTMs and TCNs.  We will only go into as much detail as is required to understand why these architectures are able to capture temporal dependencies and, therefore, are relevant to predicting extubation failure.  Any additional or more complex mathematical information is beyond this project's scope.

### Neural Networks and Model Training

Before we delve into LSTMs and TCNs, we must briefly cover neural networks and activation functions.

Neural Networks (NNs) are inspired by biological neural networks in the brain [85, 86].  It comprises several layers of neurons, all with their own weights and bias terms.  Neural networks are typically trained via backpropagation and gradient descent to iteratively minimise the error between their predictions and the ground truth [87].  Once model training is complete, the refined weights and bias terms represent a complex arbitrary mathematical function that accurately represents the underlying data's inherent patterns [88].

### Activation functions

Activation functions add non-linearity to neural networks and enable them to model complex relationships within the data.

We briefly outline the key activation functions highlighted in this project [88]:

- **Sigmoid** – Produces an output between 0 and 1 commonly used in binary classification

- **Tanh** – Outputs a value between -1 and 1, particularly useful in hidden layers in models where negative values are meaningful

- **ReLU (Rectified Linear Unit)** – Returns either 0 if the input is less than 0 or the value itself for positive inputs. ReLU is popular in hidden layers due to computational efficiency and ability to avoid the vanishing gradient problem

- **Leaky ReLU** – An adaptation of ReLU designed to avoid the "dying-relu" problem by ensuring there is a small non-zero value even when the input is negative

**Long Short-Term Memory: LSTM**

LSTMs are specialised forms of neural networks. A simple LSTM unit comprises a memory cell containing an input gate, an output gate and a forget gate (as shown in Figure 2.2) [89]. The network acts like a conveyor belt for information, designed to process data sequences – remembering essential information from the past and discarding irrelevant information, which enables them to capture long-term dependencies.

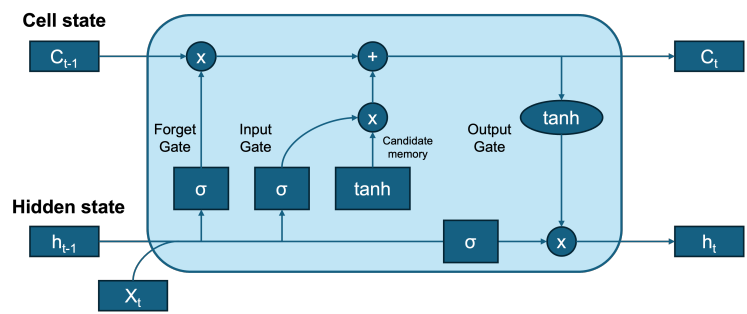

Figure 2.2: LSTM Memory Cell architecture. X: input, h: hidden state, C: cell state, $\sigma$: sigmoid activation, tanh: tanh activation, +: sum over all inputs, x: multiplication. Adapted from Van Houdt *et al.* [89]

The components operate simply as follows [89]:

- **Memory Cell:** Allows information to flow through the network and be modified via the various gates

- **Forget Gate:** Determines what information should be removed from the previous cell state

- **Input Gate:** Controls what new information should be incorporated into the cell state

- **Output Gate:** Controls what information should be output from the cell state

During a forward pass through the network, the LSTM cell processes the current input, the prior hidden state and the previous cell state. Gates control the flow of information, determining what to forget, what information should be added and what to output. Typically, an LSTM network would have multiple cells chained together. The flow of gates and activations enables the network to retain information from previous states, facilitating learning of long-term temporal dependencies, which is required for predicting extubation failure.

**Temporal Convolutional Network: TCN**

TCNs are inspired by CNNs. CNNs rely on convolutional layers, which apply filters to input data via convolution [90]. Convolution is used to extract features from data, typically used in image processing to identify edges and points of interest [75]. Convolutional layers are combined with other layers, such as pooling and fully connected layers, to perform feature extraction and produce a final output.

TCNs use convolutional layers to extract features and patterns from time-series data (which is one-dimensional compared to the typical two-dimensional image). To achieve this, two key adaptations are made to tailor to the unique characteristics of temporal data and temporal dependencies:

- **Casual Convolutions:** The output from a given time step is convolved only with information from the past, ensuring patterns are only learnt from current and past time steps and not the future. This maintains temporal order [91]

- **Dilated Convolutions:** Dilated convolutions are used to capture long-range dependencies. When a convolutional filter is "dilated", i.e. spread out over a larger range of data, it can "see" more data. When applied to one-dimensional, time-series data, dilations facilitate patterns to be captured across extended periods of time [92]

A representative TCN architecture is shown in Figure 2.3 adapted from Bednarski et al. [93].

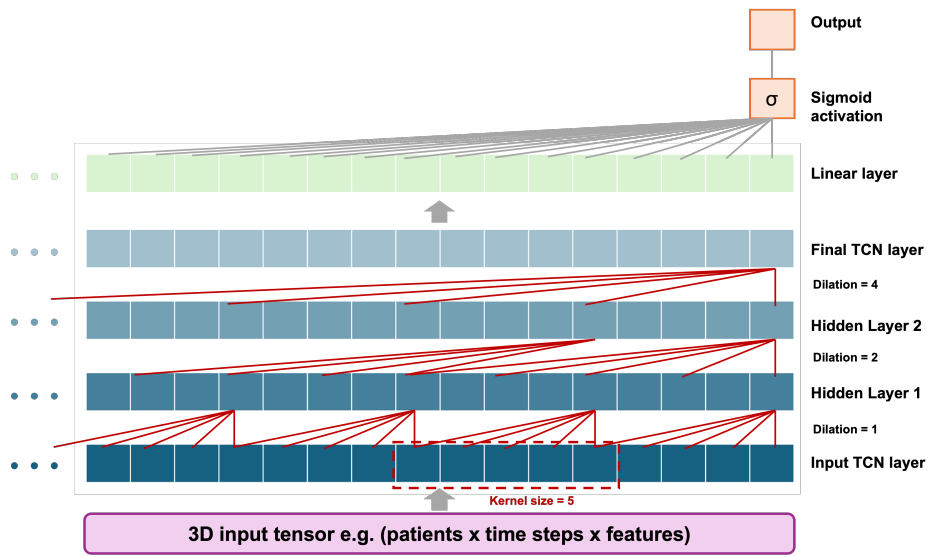

Figure 2.3: Exemplar TCN network for binary classification. The network employs a kernel size of 5 with exponentially increasing dilation factor with each hidden layer. Dilated, casual convolutions facilitate the learning of long-term temporal dependencies. Adapted from Bednarksi *et al.* [93]

In a TCN, data flows through convolutional layers, where casual and dilated 1-D convolutions are applied using a kernel. In each progressive hidden layer, the dilation is exponentially increased, facilitating the dilated convolutions to capture longer-range dependencies with each layer. As a result, kernel size, dilation factor and the number of hidden layers are key parameters in TCN architecture.

The combination of dilated, casual convolutions allows TCNs to process longitudinal data, capturing both short- and long-term patterns and, therefore, temporal dependencies – necessary for predicting extubation failure.

# 3

# Data Preparation and General Pre-processing

## 3.1 Data Preparation

**Clinical assistance**

From the start, it should be noted that all data pre-processing steps detailed in this section were clinically validated with Dr Murali, an intensive care doctor at Barts Health NHS Trust, London, UK. This ensured that this project adhered as closely as possible to real-world clinical practice and considerations, with a view towards clinical adoption.

**Data sources**

This cohort study was conducted retrospectively on data from the Medical Information Mart for Intensive Care IV (MIMIC-IV) [94–96]. MIMIC-IV is a publicly available database containing electronic health records of over 65,000 patients admitted to the ICU and over 200,000 patients admitted to Beth Israel Deaconess Medical Centre in Massachusetts, USA, between 2008 and 2022.

It is important to note that the database was deidentified per Health Insurance Portability and Accountability Act (HIPAA) standards, meaning all names, dates, and times are protected and represented by ambiguous IDs or values. Furthermore, upon review by the Institutional Review Board at the Beth Israel Deaconess Medical Centre for database creation, a local ethical review board (ERB) application was achieved. Thus, Imperial College London did not require ERB approval for this project.

The secondary database available for this project was from the Weaning Algorithms for mechanical VEntilation (WAVE) study, an unpublished randomised control trial building on prior work by Rees *et al.* [97] and Patel *et al.* [98] investigating the development and implementation of the Beacon Caresystem. While the WAVE dataset was too small for independent model development, its features represented recent real-world clinical practice and proved valuable in guiding our feature selection process.

**Patient cohort extraction**

We now needed to derive the relevant patient set from the over 200,000 in MIMIC-IV to form the training and test sets for model development. As such, we initially analysed a representative sample of studies in the ICU and mechanical ventilation domain that used MIMIC-IV or previous versions to identify their Inclusion Criteria and Exclusion Criteria (summarised in Appendix Table B.11). To define these two terms, we classify Inclusion Criteria as specific conditions a patient must meet to be considered for inclusion in the initial dataset. Exclusion Criteria are conditions that would disqualify a patient from being included in our research and are applied to the initial inclusion set. These criteria minimise confounding factors and ensure the study cohort is homogenous and representative of the target population of patients undergoing IMV.

From the start, it was clear that these criteria needed clinical justification and were representative of the considerations intensivists would use in the ICU.

As such, with the assistance of Dr Murali, we developed the following inclusion criteria, as well as their rationale and implementation, for future replicability. Appendix Table B.1 outlines each criterion's rationale and implementation.

- **Criterion 1: Needs to be in the ICU**

- **Criterion 2: Needs to have undergone invasive mechanical ventilation**

- **Criterion 3: Needs to have been extubated**

Once applied, the initial dataset consisted of 14,315 unique patients and 16,243 ventilation events with an associated extubation. The larger number of ventilation events reflects the situations where the same patient had multiple ventilation treatments.

Next, we needed to derive a set of Exclusion Criteria to ensure we had a representative dataset for model development. Once again, with the assistance of Dr Murali, we selected the following criteria, with the rationale and implementation outlined in Appendix Table B.2.

- **Criteria 1: Age < 18 and > 89 – excluding paediatric and extremely elderly patients**

- **Criteria 2: Duration of ventilation < 24 hours or > 30 days**

- **Criteria 3: Excluding repeated admissions**

- **Criteria 4: Patients under palliative care and end-of-life patients**

- **Criteria 5: Patients that died during ventilation**

- **Criteria 6: Patients who had major head or neck surgeries**

Once these criteria were applied to the initial inclusion set, the resultant patient cohort was 5,970. This cohort was then taken further to extract time series data for these patients to create the training and test sets. Figure 3.1 outlines the entire cohort selection process.

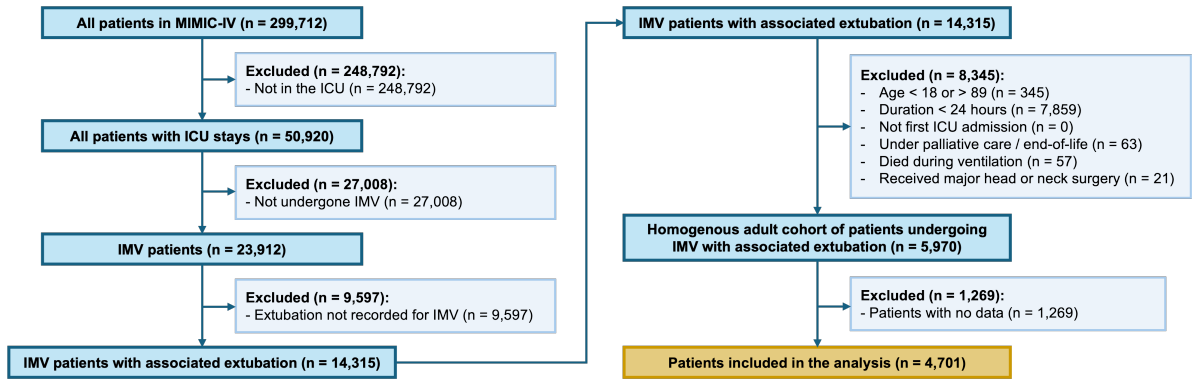

Figure 3.1: Patient cohort extraction flow implementing clinically determined inclusion and exclusion criteria.

### Data annotation

As reiterated, this study's desired outcome for model prediction is extubation failure/success. However, there is no consensus definition in the medical domain.

Therefore, in this study, we sought to establish a clinically relevant definition of extubation failure. The following criteria for success/failure were developed in close collaboration with Dr Murali:

Extubation failure is defined as:

- **Re-intubation within 48 hours of extubation**

- **Mortality within 48 hours of extubation**

- **If neither were the case, failure is logged if a patient was placed on ventilatory support within 6 hours post-extubation**

The patient cohort was annotated using these criteria, with those where the criteria applied being labelled 1 (failure) and 0 (success) for the rest. This framed our study as a binary classification task.

The re-intubation criterion was applied by searching for any patients with a recorded invasive ventilation event whose start time is within 48 hours of the end time of extubation of their first ventilator stay, resulting in 595 patients classed as failures. Mortality was determined using the recorded death time where provided, with any patients whose death time is within 48 hours after extubation end time being classed as failing extubation, with 487 mortalities within 48 hours. Finally, we needed to check if the remaining patients had been placed on ventilatory support. Ventilatory support was taken to be NIV, receiving $O_2$ flow, CPAP and any BiPap events recorded for that patient in the *chartevents* table within 6 hours of extubation end time. Where these were applied, the patient was classified as having failed extubation.

This resulted in 1,911 patients in the failure class and 4,059 in the success class, implying an extubation failure rate of 32.0%.

### Feature selection

Once we had derived our annotated patient set, we needed to select the relevant features from which to extract data. The number of features used in MIMIC-based studies varies in the literature;

thus, there is no consensus set of features most indicative of extubation failure, highlighting the aforementioned heterogeneity around extubation decisions in the ICU.

To select features for this study, we first comprehensively analysed those commonly used across related literature (shown in Appendix Table B.12. As per Dr Murali's recommendation, these features were grouped into clinically relevant categories: Demographics, Vital Signs, Laboratory Results and Ventilator Information. The features under "Other" were excluded from further analysis due to their adjudged lack of clinical significance. It should be noted that the studies in this review include extubation failure-specific papers and papers in the domain of ventilation and weaning, more generally, where features used were relatively easily identifiable to get a broader spectrum. The top fifteen features based on the number of papers they were used in from Vital Signs, Laboratory Results and Ventilator Features were taken where available to see which features were most popular in the wider literature. Demographic features were selected and placed into feature sets based on clinical recommendations.

Recognising that data collection can vary across ICUs, several features may be prevalent in literature but are not typically collected in real-world scenarios. While we did not conduct a primary study, the WAVE data represented a real-world clinical study and was leveraged as a benchmark (features are outlined in Appendix Table B.13). By cross-referencing our literature-derived features and those from the WAVE study, we created four cross-reference sets of features based on the overlap in popularity in literature and clinical availability. For reference, clinical availability refers to a feature being present in the WAVE data.

- **Cross-reference Set 1:** Features that are popular in literature and clinically available

- **Cross-reference Set 2:** Features that are popular in literature but not clinically available

- **Cross-reference Set 3:** Features that are not popular in literature but clinically available

- **Cross-reference Set 4:** Features that are not popular in literature and not clinically available

The feature sets are ordered by perceived study relevance, with Set 1 being the most relevant. Set 2 was prioritised over Set 3 as its popularity in the literature suggests strong potential for predictive value, acknowledging the evolving nature of clinical practice where research may drive new data collection practices. Set 4, comprising features that are not popular and not clinically available, was not considered further to ensure that only potentially informative were included in model development.

These three sets used in this study and their availability in MIMIC are outlined in Appendix Tables B.3, B.4 and B.5. Not all features from either literature or WAVE are available in the MIMIC database and, therefore, could not be extracted. Demographic/static features were assigned based on clinical consultation to indicate which data is available as standard in a typical ICU setting, e.g., age and weight, compared to those not always immediately available, e.g., Charlson score. Moreover, where features had the same popularity score in the literature, their assignment was carried out with clinical recommendations.

Despite its popularity, ventilation duration was not used as a feature since we controlled the range with our Exclusion Criteria, thus perturbing the natural variations that would occur. Heart failure and Respiratory failure were also not used despite their popularity.

**Data extraction**

After selecting the relevant features, both time series and static data needed to be extracted from the MIMIC-IV database.

To extract time-series data, we needed to decide on a relevant time window. In collaboration with Dr Murali, we chose the last six hours before the start of extubation. This six-hour window represents a clinically recommended timeframe within which data likely best informs decisions about extubation and its success or failure.

The relevant data for each feature set was extracted separately from the *chartevents* file, where the logged time of the event is within a 6-hour time window of the patient's extubation start time. It should be noted that upon data extraction, the resultant patient cohort was reduced to 4,701 as there were 1,269 patients for whom no data was recorded for any features across all feature sets. As such, our new and final cohort size was 4,701, and the extubation failure rate was 32.8%, which was not too distant from the original 32.0%.

All 15 features in Set 1 and all 19 features in Set 2 were extracted; however, for Set 3, no data was available for any patients for Resting Energy Expenditure and Vd/Vt ratio.

Time-series data was then reformatted to define the relative time from the start of the 6-hour window for each data point. This ensured that data could be represented chronologically, essential for input into models that learn from ordered temporal dependencies such as LSTMs and TCNs.

Static features were extracted from the *patients* and *chartevents* file. For Set 1, age was taken as *anchor_age* from the *patients* file, as were gender and ethnicity. Ethnicities where the value was "UNABLE TO OBTAIN" or "UNKNOWN" were placed in the "OTHER" category to ensure features were as relevant as possible and to reduce dimensionality. Weight, height, and BMI were extracted as the first recorded values for daily weight and height (cm) in the *chartevents* file. Where there were no recorded weights or heights for a patient, the values were set as the average of the patient population.

Table 3.1: Charlson Score weights. Adapted from Yang *et al.* [99]

| Comorbidity | Score |
| --- | --- |
| Prior myocardial infarction | 1 |
| Congestive heart failure | 1 |
| Peripheral vascular disease | 1 |
| Cerebrovascular disease | 1 |
| Dementia | 1 |
| Chronic pulmonary disease | 1 |
| Rheumatologic disease | 1 |
| Peptic ulcer disease | 1 |
| Mild liver disease | 1 |
| Diabetes | 1 |
| Cerebrovascular event | 2 |
| Moderate-to-severe renal disease | 2 |
| Diabetes with chronic complications | 2 |
| Cancer without metastases | 2 |
| Leukemia | 2 |
| Lymphoma | 2 |
| Moderate or severe liver disease | 3 |
| Metastatic solid tumour | 6 |
| Acquired immunodeficiency syndrome (AIDS) | 6 |

The only additional feature for Set 2 that could practically be derived from MIMIC was the Charlson Comorbidity Index score. As this required more information about the patient's medical history, it was placed in Set 2 as it was deemed not to be easily clinically available. The Charlson score is a weighted index predicting the risk of death within one year of hospitalisation created by Charlson *et al.* in 1987 [100]. The score for each patient was determined by extracting the relevant diagnoses and conditions and calculating the weighted score as per Table 3.1 from Yang *et al.* [99] summed with the age-dependent weights (50-59: 1 point, 60-69: 2 points, 70-79: 3 points and 80 years old or more: 4 points).

These three feature sets served as the basis for incremental model training to analyse whether increasing the number of features impacts performance. Appendix Table B.6 summarises basline statistics from all features in the three sets of our patient cohort.

Overall, the statistics between the success and failure cohorts are comparable, with a few features having significant differences. There seems to be a bias towards older patients, which is expected. There is also certainly a dominance of patients classed as Ethnicity (White), likely reflecting the population in the area surrounding Beth Israel Deaconess Medical Centre in Boston.

## 3.2 General Data Pre-processing

The following section outlines the general pre-processing methodology applied to all dynamic and static data regardless of the intended model architecture.

Hereinafter, the feature sets will be treated as incremental feature sets that combine the Cross-reference Sets outlined previously. They will be known as Feature Set 1, Feature Set 2 and Feature Set 3. For reference, Feature Set 1 contains only features from Cross-reference Set 1, Feature Set 2 includes all features from Feature Set 1 plus those in Cross-reference Set 2, and Feature Set 3 includes all data from Feature Sets 1 and 2 plus those from Cross-reference Set 3. This was to facilitate model training with increasing numbers of features.

Since Feature Set 1 was the most clinically relevant feature set, we tailored our methodology to its characteristics and applied that same methodology to the other two sets for consistency.

**Dynamic data general pre-processing**

Temporally variant (dynamic) features were pre-processed by removing low-observed features. This was achieved by calculating the average observation frequency of all features per patient in the training set (importantly, not the test set to prevent data leakage) and excluding features falling below a specific set threshold.

For Feature Sets 1 and 2, this threshold was set as 0.5 observations in the 6-hour window to balance the amount of data needed to be synthesised and the density of data used to train the model. However, for Feature Set 3, this threshold was lowered to 0.15; at 0.5, there was only one additional feature compared to Feature Set 2, which was deemed insufficient to justify the extra computational cost and likely not meaningful. Thus, the threshold was lowered to allow us to train the model on more features. The observation frequencies of the training set of Feature Set 1 are shown in Table 3.2 for reference, with Feature Sets 2 and 3 outlined in Appendix Tables B.7 and B.8.

Table 3.2: Average feature sampling frequency for Feature Set 1 in the training set. Where features are greyed out reflects their removal having fallen beneath the threshold of 0.5

| Feature | Average Train Set Sampling Frequency |
|---|---|
| Respiratory Rate | 6.615 |
| $O_2$ saturation pulseoxymetry | 6.611 |
| Inspired $O_2$ Fraction | 2.104 |
| Tidal Volume (observed) | 1.600 |
| Minute Volume | 1.598 |
| Peak Insp. Pressure | 1.524 |
| Tidal Volume (spontaneous) | 1.368 |
| Ventilator Mode | 1.204 |
| PH (Arterial) | 0.535 |
| Arterial $CO_2$ Pressure | 0.525 |
| Arterial $O_2$ pressure | 0.525 |
| Hemoglobin | 0.247 |
| EtCO$_2$ | 0.189 |
| Plateau Pressure | 0.160 |
| Negative Insp. Force | 0.008 |

Data was split into the train and test sets using an 80/20 split as standard. Notably, the split was stratified based on the calculated proportion of synthetic data required to resample the data to 30-minute intervals (the highest resampling frequency used in this study) based on the data in Feature Set 1. This ensured that the amount of synthetic data was comparable between the train and test sets.

This resulted in a final training set of 3,760 patients and a test set of 941 patients, with an average synthetic data proportion of 5.37 new data points per patient in the training set and 5.40 new points per patient in the test set. The class distribution between the two sets was also comparable, with 67.0% success and 33.0% failure in the training set, respectively, and 67.4% success and 32.6% failure in the test set. These 3,760 and 941 patients were maintained throughout the study as the definitive training and test sets for consistent comparison and evaluation. A comparison of dynamic feature statistics between the train and test sets is showcased in Appendix Table B.9, highlighting their similarity and noting Arterial Base Excess, which was later constrained through outlier removal.

Upper and lower bounds were set to remove extreme outliers that would otherwise skew the dataset. MIMIC helpfully provides high and low normal values for some time-series features; where present, these were extracted and used. Otherwise, a standard range of mean $\pm$ 3 standard deviations was used (or 0 as the minimum value where negative values for a feature are nonsensical). The derived bounds for all features are highlighted in Appendix Table B.10. Notably, the bounds were determined using the training data and applied to both the training and test data to avoid data leakage.

**Static data general pre-processing**

Static features were compiled into the relevant incremental feature sets and split by patient into the predefined train and test sets. Age, Gender and Ethnicity were treated as categorical values. Age was categorised to APACHE II score (used in ICUs for mortality assessments) bins [101]: $\leq$44, 45-54, 55-64, 65-74, and $\geq$75. Outliers were removed for numerical features (weight and height) using the interquartile range as lower and upper bounds, with removed values instead filling the training set mean, and BMI recalculated. Bounds and mean values are taken only from the training data and applied to train and test data where necessary. Numerical features were min-max scaled between 0 and 1, and categorical features one-hot encoded to create the static data for Feature Set 1.

Once calculated, the Charlson score was treated as a numerical feature per clinical oversight, and min-max scaled alongside the numerical data from Feature Set 1 to create the static data for Feature Set 2. The static data for Feature Set 2 was also used for experiments involving Feature Set 3.

This general pre-processing resulted in static and time-series data ready for further architecture-specific pre-processing steps, detailed in the following section.

# 4

# Model Development and Experimentation

This chapter presents a walkthrough of how our model development process evolved, illustrating the inherent inter-connected relationship between data processing and architectural choices.

This project aimed to develop an end-to-end prediction system that can assist clinicians when they need to make an extubation-related decision by predicting extubation failure. The models used had to capture patient trajectories effectively and, therefore, needed to use architectures that could effectively learn temporal dependencies. For this study, we employed LSTM and TCN models. LightGBM models served as a baseline comparison as they were observed to be among the best-performing in the literature but crucially less computationally intensive, which was essential given the limited resources for this project [53].

We needed to train separate models on the three feature sets to compare the additive effect of including features. Furthermore, we also wanted to determine whether adding static data was informative compared to just training the model on dynamic data. As such, two models were trained for each of the three architectures, one on dynamic data only and another on static and dynamic data. In summary, for each of the three feature sets, we needed to create architectures for and train the following six models: LSTM – dynamic data only, LSTM – dynamic and static data, TCN – dynamic data only, TCN – dynamic and static data, LightGBM – dynamic data only and LightGBM – dynamic and static data.

Predicting extubation failure with time-series data necessitated carefully considering how the data was treated and how that guided the architectural design. This chapter outlines our iterative approach to addressing these challenges and the interim results that guided our decision-making throughout.

## 4.1 Initial time-series pre-processing and model development

As before, we decided to develop our pre-processing strategy based on Feature Set 1, the optimal feature set, and then apply it to Feature Sets 2 and 3 for consistency. As such, any interim results or data, where shown, relate to Feature Set 1.

The field of predicting extubation failure using temporal architectures lacks a standardised approach. While Zeng *et al.* [80] is the most relevant peer-reviewed paper, their pre-processing methods are not directly applicable to our study since they do not focus on a fixed time window. Therefore, pre-processing data extracted within the 6-hour window would require some novelty while ensuring clinical relevance.

LSTMs and TCNs need the final input data sampled at a uniform frequency. Each feature must contribute a value at constant time steps within the 6-hour window to ensure input shape consistency. If the time steps do not align across features, it would be difficult for models to learn meaningful patterns. Inherently, this presented several challenges. The data extracted from MIMIC was by no means consistent. We determined the average sampling frequency on the training set for each feature per patient over the 6-hour window, as shown in Table 3.2. On average, the features are sampled at differing rates, from c. 6.6 times in the 6-hour window for $FiO_2$ to a minimum of c. 0.5 times for $PaO_2$. Furthermore, even within the same feature, data was not collected consistently and sometimes not at all, making it initially unsuitable for LSTM/TCN input. Therefore, careful consideration was needed regarding our data resampling and interpolation strategy to best represent inherent patient trajectories.

Looking at the literature, the aggregation technique used by Fenske *et al.* [79] was also not relevant here as they had one time step based on aggregated data for each day rather than within a defined window. Zeng *et al.* [80] used a padding and packing technique to handle variable sequence lengths. However, since their data was collected over a much longer timeframe using windowing and inherently had more data, this technique was irrelevant to our fixed 6-hour window.

Excessive interpolation for features with low sampling rates can introduce noise, but sufficient temporal resolution is required for practical model training. As such, in collaboration with Dr Murali, we opted for a 30-minute resampling interval to strike a balance between data density and minimising synthetic data generation. The finer granularity would allow more detailed information about the patient's trajectory to be captured and help the model detect subtle patterns and trends that might be missed with, say, 1-hour intervals. Furthermore, as corroborated by Dr Murali, many physiological parameters can change significantly within an hour, and clinicians review data relatively frequently. By mirroring clinical practice, we would ensure our models are trained on data reflecting real-world decision-making as closely as possible.

To interpolate the data to 30-minute intervals, simple forward fill or backward fill would not be clinically relevant, and in some cases where there was no data for a patient, it would not be possible. Several interpolation strategies exist, but we initially decided to use a combination of spline and linear interpolation in a tailored imputation strategy. We intended to use linear interpolation for features with clinically typical frequent changes or those that do not require a smooth curve and spline interpolation for features that are clinically known to change smoothly and continuously. However, it became clear that spline interpolation would not be possible as it requires at least k+1 data points where k is the spline order. As such, linear interpolation was chosen with clinical validation due to its simplicity and ability to capture temporal relationships, avoiding the computational expense of more complex methods such as KNN Imputation.

Linear interpolation requires values to interpolate between. Suppose there is no data to interpolate from, which would be the case for many patients, the resultant interpolated data would be set to meaningless NaN values. Thus, with validation from Dr Murali, we devised the following logic to create start and end values for each patient to ensure that there are at least two values between which data points could be interpolated. This created start (0 min) and end values (360 min) for all features, addressing potential issues with missing data that could arise from direct interpolation without boundary values.

- For each patient and each feature, if there was already a data point at 0 minutes (the start of the window) or 360 minutes (the end of the window), then that value was kept

- If there was no such value, if there is a value within the first or last 15 minutes (half the 30-min interval), then that is treated as the start / 0 min value or the end / 360 min value, respectively

- If neither are present, the average value for that feature across all patients in the first/last 15 minutes will be taken (ensuring the mean values are calculated from the training data and applied to the training and test data to prevent data leakage)

Following conversations with Dr Rees about research into extubation, upon data resampling, we created new clinically relevant features by dividing values for respective features at the same time point, creating the $SpO_2$:$FiO_2$ ratio and the $PaO_2$:$FiO_2$ ratio features. Both are informative clinically and from a research perspective.

As a categorical feature represented in numerical form, Ventilator Mode had to be handled separately. Instead, where the average was used for numerical features to create start and end values, we used the mode values since the mean is illogical. If no value was present at the start, the most common ventilation mode across all patients in the first 15 minutes was taken, as this is likely the start mode. If no value was present at the end, the most common ventilation mode was taken across all patients in the last 15 minutes before extubation. Moreover, linear interpolation would also be meaningless since values need to be integers. Thus, each 30-minute value was imputed using the mode in a 30-minute window (15 on either side) across all patients. The modes were calculated from the training data and applied to the test data to prevent data leakage.

The time-series data was then normalised via min-max scaling between 0 and 1 and converted into temporal sequences in chronological order from window start to end. The final output was a 3D NumPy array of the shape (no. patients, no. time steps, no. features) for both the train and test data.

**Initial LSTM model development**

We began our model development with the LSTM architecture, primarily due to the existence of a direct precedent which does not exist for TCNs in this domain.

One challenge we had to address from the start was that our dataset was inherently imbalanced (the extubation failure rate was c. 32%). Although, from a clinical perspective, this number is reasonable and, as confirmed by Dr Murali, observed in the real world, it may be problematic for machine learning as ML algorithms tend to favour the majority class [102]. As a result, models can struggle to learn patterns to predict extubation failure and incorrectly classify such patients, which would not be apropos for clinical adoption.

We created a bespoke sampling hyperparameter to address class imbalance with three options: normal, undersample and oversample. "Normal" maintained the current class balance, "undersample" uses the RandomUnderSampler and "oversample" uses SMOTE (Synthetic Minority Oversampling Technique) (manual over and under sampling were employed where the packages were not compatible with the training process and achieved the same effect) [103]. Note that only the training data was over- or under-sampled; the validation and test data remained at the original class balance to evaluate model generalisability.

We also created a bespoke loss hyperparameter that provides the option for weighted loss alongside a standard binary cross-entropy loss. This would be used only where the sampling method was "normal" and the loss parameter was "weighted", as applying it on top of over or under-sampled data is redundant.

The intention with the initial model architectural design was to keep it simple to establish a baseline performance with the knowledge that simpler architectures are generally easier to interpret. A simpler architecture also isolates the effects of data pre-processing strategies from model complexity, allowing us to diagnose areas of improvement more quickly.

As such, our LSTM model consisted of one LSTM layer, a fully connected (linear) layer to map the LSTM output to a one-dimensional scalar and sigmoid activation for binary classification. The hidden and cell states of the LSTM were initialised to zeros, as is standard practice.

The LSTM model was trained using a systematic approach involving hyperparameter tuning and evaluation. Optimal hyperparameters were determined using five-fold cross-validation (CV) on the training set. Specifically, a Grid Search was performed on the hyperparameter grid, as detailed in Table 4.1.

Table 4.1: Initial hyperparameter grid used for LSTM cross-validation with grid search

| Hyperparameter | Values |
|---|---|
| Hidden dimension | 32, 64, 128, 256 |
| Layer dimensions | 1, 2, 3 |
| Dropout probability | 0.25, 0.5 |
| Optimiser learning rate | 0.001, 0.0005, 0.0001 |
| Batch size | 32, 64 |
| Number of epochs | 20, 30, 40 |
| Sampling method | 'normal', 'undersample', 'oversample' |
| Loss type | 'normal', 'weighted' |
| L2 weight decay | 0.0001, 0.001, 0.01 |

During training, model performance was evaluated on the validation set split from the training set after each epoch. AUC-ROC was chosen as the characteristic metric due to its common use across literature and robustness to imbalanced data [104]. Early stopping was employed to prevent overfitting, with the minimum delta set to zero and patience set to five. Dropout and L2 regularisation were also utilised to avoid further overfitting, which is common in deep architectures. The training was carried out using mini-batch gradient descent, with the Adam optimiser implemented for adaptive updates to the model weights. Model weights were initialised using PyTorch's default methods for LSTM and linear layers.

Since Grid Search can exploit computational resources when there are several combinations, we initially employed three random searches with 50 iterations to try and narrow down the hyperparameter set. Where common values appeared across the random searches, the range could be more focused, resulting in the refined parameter grid (Table 5). The hyperparameters from the random searches are shown in the Appendix Table C.1.

Grid Search was then employed to determine the optimal LSTM hyperparameters based on this refined parameter grid. The optimal hyperparameters were as follows: 'hidden_dim': 32, 'layer_dim': 2, 'dropout_prob': 0.5, 'learning_rate': 0.0005, 'batch_size': 64, 'num_epochs': 20, 'sampling_method': 'undersample', 'loss': 'normal', 'weight_decay': 0.0001 and had a validation AUC-ROC of 0.6295, representing the best performance across all hyperparameter combinations. At first glance, it would suggest the model struggles to accurately classify certain instances and potentially overfitting. To confirm this on the test set, the final model was trained on the entire training set sampled according to the sampling method hyperparameter. Performance was evaluated using standard metrics of Accuracy, Precision, Recall (Sensitivity), F1 score, AUC-ROC and Specificity with the standard classification threshold of 0.5. Hereinafter, the classification threshold will be 0.5 unless otherwise stated. This produced metric values of Accuracy: 0.3273, Precision: 0.3273, Recall (Sensitivity): 1.0000, F1 Score: 0.4932, AUC-ROC: 0.4388.

Table 4.2: Refined hyperparameter grid based on Random Searches using hyperparameters from Table 4.1

| Hyperparameter | Values |
|---|---|
| Hidden dimension | 32, 64 |
| Layer dimensions | 1, 2, 3 |
| Dropout probability | 0.25, 0.5 |
| Optimiser learning rate | 0.001, 0.0005 |
| Batch size | 32, 64 |
| Number of epochs | 20, 30, 40 |
| Sampling method | 'normal', 'undersample', 'oversample' |
| Loss type | 'normal', 'weighted' |
| L2 weight decay | 0.0001, 0.001 |

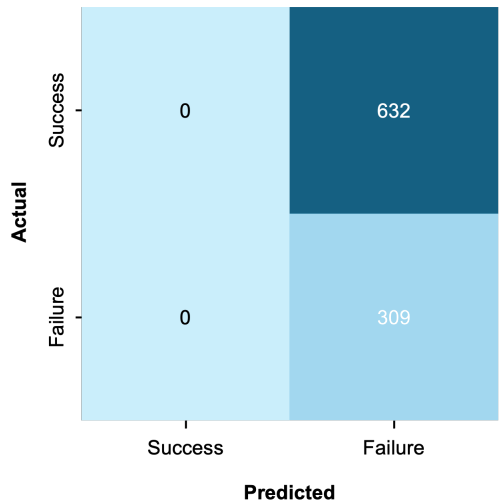

Figure 4.1: Confusion matrix for initial LSTM model with Grid Search

The results were surprising. Accuracy is less informative on an imbalanced dataset, so this will be ignored. However, precision, F1, AUC-ROC and specificity were all incredibly low, and recall was oddly perfect. Low precision suggested the model has a high rate of false positives, and perfect recall suggests the model correctly predicted all extubation failure cases. However, this could be misleading if the model classifies all patients to be in the failure class, as indicated by the specificity being 0. To confirm, the confusion matrix was determined in Figure 4.1, and it was immediately clear that the model was biased towards the positive class.

This behaviour was particularly odd, considering the failure class is the minority class in the original data. Thus, one would expect predictions to be biased towards the success class. This suggested that the model was inherently confused, and despite the training data being undersampled as per the optimal hyperparameters, the results suggest a significant class imbalance. Upon inspection, the under-sampling and over-sampling functions created equally balanced classes, so the issue was not with these functions. One possible explanation is that the model may have overfitted to the failure class during training despite early stopping, L2 regularisation, dropout and a relatively simple architecture being implemented. This can occur when the hyperparameters are not optimal, with suboptimal hyperparameters potentially exacerbating overfitting [105]. As such, we hypothesised that the hyperparameter tuning strategy was not producing the optimal results as it was not effectively exploring all possible combinations. Random search is known to be fallible as it can ignore several combinations [106]; thus, the refined grid in Table 4.2 is not representative enough.

**Trying Bayesian Optimisation to ensure optimal hyperparameters are obtained**

Consequently, we decided to start with a much more extensive set of hyperparameters, as outlined in Table 4.3. The values used are common across machine learning research.

Table 4.3: Extensive hyperparameter grid for LSTM tuning using Bayesian Optimisation

| Hyperparameter | Values |
| --- | --- |
| Hidden dimension | 32, 64, 128, 256, 512 |
| Layer dimensions | 1, 2, 3, 4 |
| Dropout probability | 0.0, 0.25, 0.5, 0.75 |
| Optimiser learning rate | 0.01, 0.001, 0.0001, 0.00001 |
| Batch size | 16, 32, 64, 128 |
| Number of epochs | 10, 20, 30, 40, 50 |
| Sampling method | 'normal', 'undersample', 'oversample' |
| Loss type | 'normal', 'weighted' |
| L2 weight decay | 0.00001, 0.0001, 0.001, 0.01 |

The issue here was that this grid contains 153,600 combinations, which would take days to work through with standard Grid Search, and we did not have the computational resources to carry this out. As such, we employed Bayesian Optimisation to help narrow down the hyperparameter space more efficiently using the widely used Optuna package [107]. The minutiae of Bayesian Optimisation are complex and beyond the scope of this study, but a functional explanation will be provided. Bayesian Optimisation is an alternative method of hyperparameter tuning, which does not have the computational expenses of Grid Search, and it has been shown to outperform global optimisation methods and Random Search, yielding models with slightly higher accuracy than both Grid Search and Random Search [108–110]. Simply, it is a more intelligent search algorithm that efficiently finds the optimal hyperparameters by using a probabilistic model to search the most promising parameter space and, thus, requiring fewer iterations than Grid Search, which searches exhaustively [111]. Here, employing Bayesian Optimisation allowed us to have a much larger hyperparameter space so that we could search significantly faster, and the smarter search would likely ensure we no longer had sub-optimal hyperparameters.

Using the Optuna package, we ran Bayesian Optimisation on the larger hyperparameter grid with 100 trials (the same number as Zhao *et al.* used in their CatBoost paper [49]), five-fold cross-validation as before and to maximise the AUC-ROC. The optimal parameters were: 'hidden_dim': 256, 'layer_dim': 2, 'dropout_prob': 0.5, 'learning_rate': 0.0001, 'batch_size': 32, 'num_epochs': 10, 'sampling_method': 'undersample', 'loss': 'normal', 'weight_decay': 1e-05. The final LSTM model was then trained using these hyperparameters and evaluated with the results indicating once again that the model was predicting all patients in the failure class: Accuracy: 0.3273, Precision: 0.3273, Recall (Sensitivity): 1.0000, F1 Score: 0.4932, AUC-ROC: 0.4667, Specificity: 0.0000 and confusion matrix Figure 4.2.

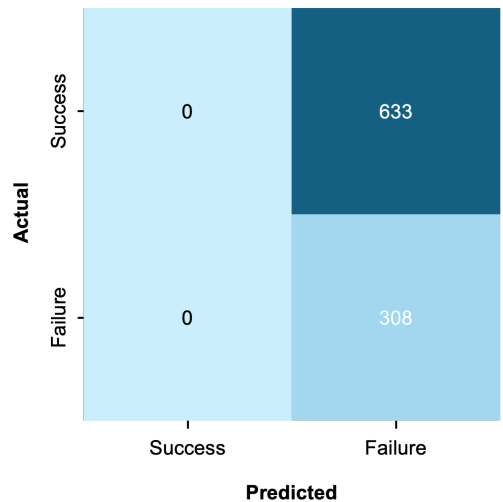

Figure 4.2: Confusion matrix for initial LSTM model with Bayesian Optimisation

To recap, we have explored various avenues to optimise our model's performance. First, we tried Grid Search narrowed with Random Search but hypothesised that this approach was not yielding optimal hyperparameters and likely overfitting. Subsequently, we employed a more sophisticated hyperparameter tuning algorithm, Bayesian Optimisation, to more intelligently search a broader hyperparameter space and obtain the optimal combination balancing exploration and exploitation. However, this resulted in the same outcome. Despite our efforts to prevent overfitting by implementing dropout, L2 regularisation, early stopping and bespoke sampling methods, and optimising hyperparameter tuning, the model was still clearly confused and very likely overfitting. As a result, we hypothesised that since the performance is invariable to the hyperparameters used, the issue does not lie within the model architecture or training logic but rather within the data itself or the way it had been pre-processed.

**Changing resampling strategy to better capture the original data**

Upon further inspection of the time-series data, we realised it contained many copied values. It seemed that resampling the data every 30 minutes did not utilise all the values in the original data and merely replicated the synthesised start values, creating a dataset not representative of underlying time-series patterns. We hypothesised that it was highly likely that the flawed data representation significantly contributed to the model's performance and likely overfitting. The copied values would introduce artificial patterns and hinder the model's ability to learn generalised temporal patterns. As such, we had to establish an alternative interpolation strategy.

The smallest unit of time recorded in the *chartevents* file is measured in minutes. The challenge here was that standard linear interpolation on a resampled sequence does not consider values in between intervals. For example, if a data point was recorded at 20 minutes and we chose to resample to every 30 minutes from the start, the data point would be removed – hence, we would lose a lot of relevant data. We noticed that many data points in the original dataset were recorded at intervals that are not multiples of 30 minutes and hence would not have previously influenced the final interpolation.

To capture as much of the original data as possible, after imputing start and end values for each feature, we initially resampled the data every 1 minute, capturing all original data points. Then, linear interpolation was applied to fill the values, and the data was finally re-sampled at the pre-defined target interval of 30 mins. If any values could not be interpolated, the mean value for that feature for that patient was used. This ensures that all data points, no matter where they were recorded, are considered when data is synthesised to better represent the patients' trajectories and temporal patterns. All other steps of pre-processing were as previously described.

It should also be noted that, at this stage, Ventilator Mode was removed from all datasets upon recommendation from Dr Murali. Individual ventilation modes here are clinically not that informative. Ideally, one would want to segment into control and support modes, but that would not be possible without full domain knowledge of the MIMIC study and ventilators used. Patients are mainly extubated once they are trialled on support mode, but it is not possible to discern this from the raw data. Hereinafter, Ventilator Mode was no longer considered.

Once the data had been resampled and interpolated, we determined the optimal hyperparameters via Bayesian Optimisation as before using the broader hyperparameter grid in Table 4.3. The resultant parameters with a validation AUC-ROC of 0.7027 were 'hidden_dim': 128, 'layer_dim': 1, 'dropout_prob': 0.0, 'learning_rate': 0.001, 'batch_size': 64, 'num_epochs': 50, 'sampling_method': 'undersample', 'loss': 'normal', 'weight_decay': 0.0001. The final trained model produced the following results on the test set: Accuracy: 0.3284, Precision: 0.3284, Recall (Sensitivity): 1.0000, F1 Score: 0.4944, ROC AUC: 0.5043, Specificity: 0.0000 and identical confusion matrix as in Figure 4.1.

While the accuracy, precision, F1 and AUC-ROC are marginally better than the previous experiment, as can clearly be seen, the model is still completely biased towards the failure class. Despite our efforts to establish more natural temporal patterns by fully utilising the underlying data, this behaviour suggests that the model may be latching onto specific features or patterns highly correlated with the failure class.

There is an argument that model complexity could encourage overfitting [112], with the number of hidden layers in the last two runs being 128 and 256. However, the first experiment had a hidden size of 32 and a relatively low number of epochs (20), yet it still exhibited the same behaviour, hence, this is likely not the case.

Ostensibly, there were deeper reasons for model behaviour outside architecture, hyperparameter tuning and bespoke resampling. As outlined in Blum *et al.*, feature relevance is a vital consideration in machine learning [113]. At this stage, we addressed this by identifying features that were important in the literature and clinically available, with Feature Set 1 representing theoretically the most informative set of features. However, despite the alleged importance of these features, it may be the case that they may not be sufficiently informative to enable the model to differentiate between success and failure alone. As such, we hypothesised that the current cohort of dynamic features was insufficient to prevent potential overfitting and model bias. As in the literature, incorporating static patient data could provide a more comprehensive picture for each patient.

**Testing with the addition of static data**

A large amount of static data is collected upon admission to the ICU and can contribute to clinical decisions. Static factors such as age have been demonstrated to require particular consideration from physicians in intensive care [114].

The current architecture can only take in dynamic data as the input is an LSTM layer. There are a number of ways in which static data can be handled alongside time-series data. Zeng *et al.* assume static values stay constant over the time window and, therefore, repeat them across all time steps, feeding the static data into the LSTM/GRU alongside the dynamic data [80]. However, including the same value multiple times does not provide new information at each time step. The LSTM model would inherently try to learn temporal patterns on data that does not change over time while creating unnecessary computational overhead. Repeated static values could also dilute the temporal patterns from dynamic features, especially if the static features are dominant in their importance. Separate handling of dynamic and static data could mitigate this.

To minimise computational overhead, the derived architecture simply added a linear layer to the existing LSTM architecture. Static data would be passed into the linear and mapped to the same dimensionality of the LSTM hidden state. Both outputs are concatenated together and passed through the sigmoid activation to generate a prediction (Figure 4.3). This architecture maintains simplicity while focusing on the temporal dynamics, which are the core of patient trajectories – the static features are treated as additional context to enhance model interpretation. It should be noted that the order of patients in the static data was equivalent to that of patient sequences in the dynamic data, so outputs matched when concatenated.

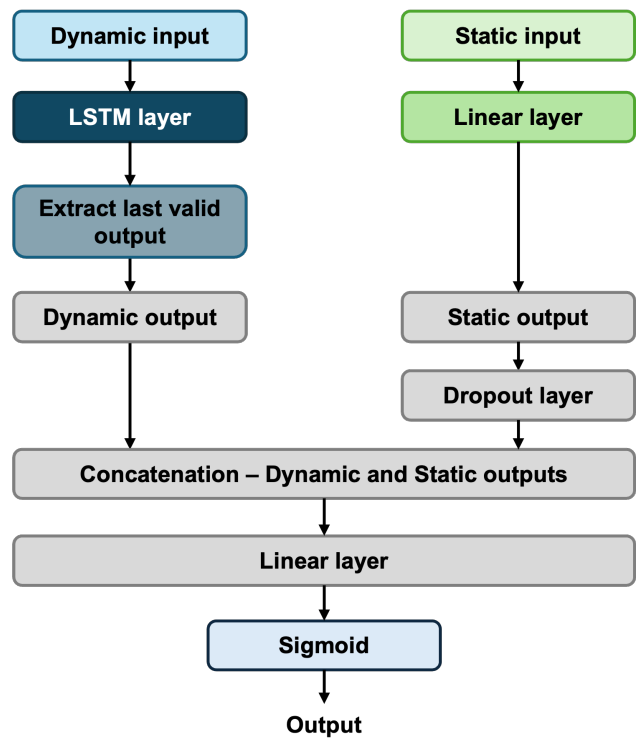

Figure 4.3: Architecture of the LSTM-Linear model for dynamic and static data processing. Dynamic data is passed through an LSTM layer while static data is passed through a linear layer. Relevant outputs from both are concatenated and passed through a final linear layer and sigmoid activation for binary classification output.

Hyperparameters were tuned as before using Bayesian Optimisation for 100 trials with the same parameter grid in Table 4.3. The optimal combination with a validation AUC-ROC of 0.7024 was: 'hidden_dim': 128, 'layer_dim': 3, 'dropout_prob': 0.25, 'learning_rate': 0.001, 'batch_size': 32, 'num_epochs': 40, 'sampling_method': 'normal', 'loss': 'normal', 'weight_decay': 0.0. Once trained, the optimal model was evaluated on the test set and yielded the results as follows: Accuracy: 0.6716, Precision: 0.0000, Recall (Sensitivity): 0.0000, F1 Score: 0.0000, ROC AUC: 0.5605, Specificity: 1.0000 and Figure 4.4.

Confusingly, the model was now demonstrating complete bias towards extubation success instead of extubation failure. Clearly, the additional context of static features has had a significant influence, but rather than balancing the data, it has biased it to the opposite extreme. This suggested that the static features likely introduce bias strongly correlated with the success class.

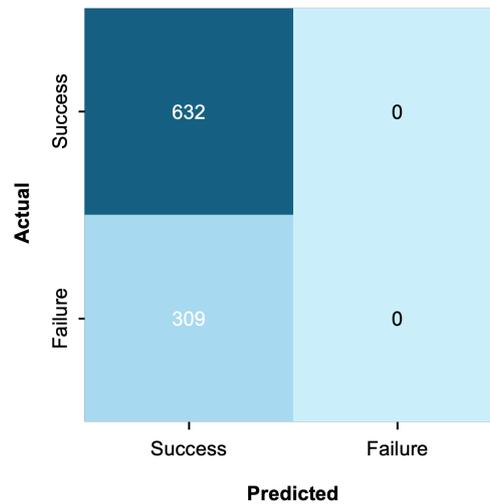

Figure 4.4: Confustion matrix of LSTM with static data included

To address this issue, it became essential to determine whether the behaviour we were seeing was specific to the LSTM architecture or a more general problem with the data. Comparing performance to a model with a fundamentally different architecture may provide valuable insights. This would be less applicable if we only used related architectures such as LSTM and GRU, as Zeng *et al.* did [80]. TCNs inherently do not rely on recurrent multiplications but rather casual, dilated convolutional layers to learn long-term dependencies. As such, TCNs avoid the pitfall of recurrent architectures potentially propagating null gradients and bias through time [75].

Pivotal work by Bai *et al.* demonstrated that TCNs outperform RNNs, LSTMs and GRUs across a broad suite of tasks and datasets [115]. As such, we hypothesised that the LSTM's unnatural behaviour may be solved by using TCN architecture instead.

**Developing TCN architecture to predict extubation failure**

TCNs have yet to be specifically employed to predict extubation failure. Thus, there was no direct precedent to instigate architectural design. However, both LSTM and TCN shared the exact requirement for sequence data. Thus, both architectures used the same pre-processed data (resampled to 1 min, linearly interpolated and then resampled to 30 mins).

The TCN architecture used in this study was inspired by the work of Bai *et al.*, which has been extensively used in machine learning research [115]. In their paper, a generic TCN architecture is outlined, which was "informed by recent research, but is deliberately kept simple, combining some of the best practices of modern convolutional architectures", thus providing a solid foundation for this report [115]. Figure 4.5 outlines the complete TCN architecture for handling solely dynamic data, which comprises the following components.

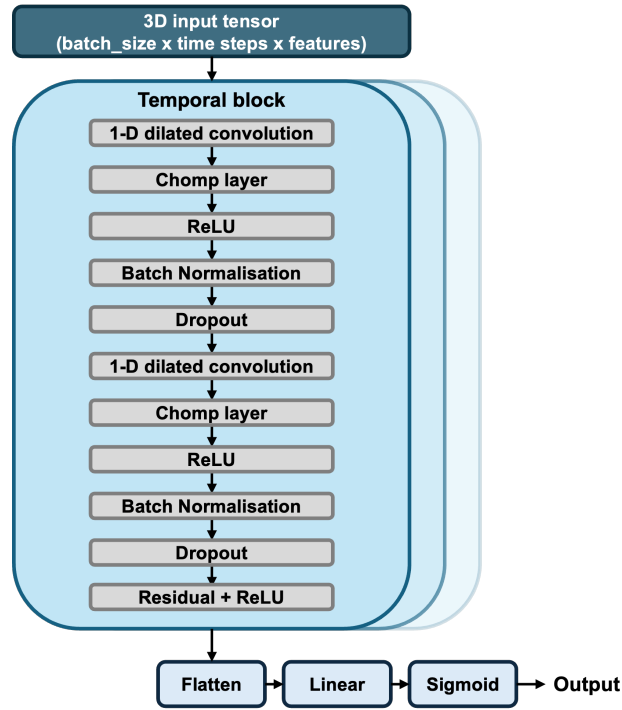

Figure 4.5: Architecture of the Temporal Convolutional Network (TCN) model. Input time series sequences in the shape of (batch_size, time_steps, no_features) is passed through multiple Temporal Blocks, each consisting of two dilated Conv1d layers, Chop1d, ReLU activation, Batch Normalisation and Dropout layers. A residual connection adds the block input to the output of the second convolutional layer, with a 1x1 convolutional downsampling applied if the input and output shapes differ. The final block flattens the output and applies a Linear layer followed by a Sigmoid activation to produce binary predictions across the batch. Adapted from Bai *et al.* [115]

- **Temporal Blocks**

  - Dilated convolutions: Each block comprises two 1D dilated convolutional layers. The network captures long-range dependencies via the dilation factor, which increases exponentially with each block ($2^0$, then $2^1$, then $2^2$ and so on), increasing the receptive field

  - Chomp layer: The Chomp layer trims excess padding added during the dilated convolution and removes elements that would otherwise allow the model to look ahead in time, preserving temporal alignment

  - Activation and normalisation: ReLU activation followed by Batch Normalisation is applied to add non-linearity and stabilise the training process. It should be noted that the Batch Normalisation layer is not present in the architecture developed by Bai *et al.* but was included as it has been shown to accelerate deep network training [116] and is included in the reviewed paper by Catling *et al.* [77]

  - Dropout: Spatial dropout is included to prevent overfitting

- **Residual Connections**

  - Each Temporal Block has a residual connection that adds the block's input to the convolutional layers' output. Crucially, if the dimensions of the input and output do not match, a 1x1 convolutional layer is applied

- **Network Structure**

    - Layer composition: The full TCN consists of sequentially stacked Temporal Blocks, with the number of blocks a hyperparameter to be tuned. Casual convolutions are implemented by setting the padding argument for the convolutional layers within the Temporal Block to be (kernel_size − 1) * dilation and the stride to 1 to ensure each output only depends on current and past elements and the output length is the same as the input length

    - Output layer: The output is flattened and passed through a linear layer and sigmoid activation for binary classification

- **Weight Initialisation**

    - All convolutional layers are initialised using Kaiming Normal initialisation. Several initialisation methods exist but this is known to be particularly effective for layers with ReLU activations and maintains the variance of gradients in the network, as outlined in seminal work by He *et al.* [117]. Biases are initialised to zero

- **Final Output**

    - The final output is a probability between 0 and 1, representing the likelihood of extubation failure for a patient

The training for this TCN architecture is nearly identical to the rigorous process of the LSTM, including five-fold cross-validation with the primary metric being AUC-ROC, early stopping, L2 regularisation, bespoke sampling methods and loss. The key differences are the inclusion of TCN-specific hyperparameters (number of channels and kernel size) shown in Table 4.4 and the use of gradient clipping with max norm set to 1.0 to prevent the exploding gradient problem (gradients becoming too large during training) that can occur in deep networks [118].

Table 4.4: Extensive hyperparameter grid for TCN tuning

| Hyperparameter | Values |
|---|---|
| Number of channels | '16,32', '32,64', '64,128', '128,256', '16,32,64', '32,64,128', '64,128,256' |
| Kernel size | Range 2 - 11 |
| Dropout probability | 0.0, 0.2, 0.4, 0.6, 0.8 |
| Optimiser learning rate | 0.01, 0.001, 0.0001, 0.00001 |
| Batch size | 16, 32, 64, 128, 256 |
| Number of epochs | Range 10 to 100 in steps of 10 |
| Sampling method | 'normal', 'undersample', 'oversample' |
| Loss type | 'normal', 'weighted' |
| L2 weight decay | 0.00001, 0.0001, 0.001, 0.01 |

The optimal parameters were determined to be 'num_channels': '32,64,128', 'kernel_size': 6, 'dropout_prob': 0.6, 'learning_rate': 0.001, 'batch_size': 64, 'num_epochs': 60, 'sampling_method': 'normal', 'loss': 'normal', 'weight_decay': 1e-05 with a validation AUC-ROC of 0.7093. The final model was then trained and evaluated using the test set and produced the following results: Accuracy: 0.3284, Precision: 0.3284, Recall (Sensitivity): 1.0000, F1 Score: 0.4944, AUC-ROC: 0.4743, Specificity: 0.0000 and an identical confusion matrix as in Figure 4.2.

As we can see, the performance was consistent with the LSTM trained on dynamic data. Once again, despite various optimisation attempts, the model was completely biased towards one of the classes. This indicates that the issue was not model-specific but rather most likely within the nature of the data itself, motivating us to revisit our pre-processing strategy to address this persistent behaviour.

## 4.2   Devising a new approach to time-series data pre-processing and model development

Thus far, we have concluded that the observed model performance is likely due to the nature of the dynamic data. In addressing this, we returned to first principles, carefully considering the original MIMIC-IV data and the pre-processing methodology implemented. Again, we had to ensure clinical relevance was maintained.

As we know from the original data, the average observed frequency for all features was, for the majority, very low (Table 3.2). In real-world terms, this means that data recording during the MIMIC study was inconsistent for all features and across time, with some features reported approximately once an hour ($O_2$ saturation pulseoxymetry) to others seemingly once every 12 hours (Arterial $O_2$ pressure). This does not provide sufficient data to train complex neural networks such as LSTMs and TCNs — Althnian *et al.* [119] outline how the lack of data is common in the medical domain. By looking into the impact of dataset size on classification, they concluded that "the overall performance of classifiers depend on how much a dataset represents the original distribution" [119]. The data points within the 6-hour window in the original dataset used in this project would not represent the underlying patient data since these features continuously vary over time. Sampling them at such infrequent rates results in inadequate and poor-quality data. Unfortunately, this was not something we had control over, given this is a retrospective secondary study.

Currently, all data synthesis occurs in the resampling and interpolation step outlined at the start of this section, where data was resampled, regardless of the sampling frequency, to a 30-minute interval. However, the quantity of synthetic data would have differed for each feature. For example, on average, poorly observed features such as $PaCO_2$ and $PaO_2$ would have linear interpolation based almost entirely on synthetic data. Conversely, for $O_2$ saturation pulseoxymetry and Respiratory Rate, the interpolated data points would be based on a greater proportion of real data, hence the temporal patterns likely more reflective of innate patient trajectories. The model would be learning from some features where the temporal pattern is more representative and others where the pattern was primarily synthetic, which we hypothesised was likely causing the biased behaviour across both LSTMs and TCNs. Thus, this variation between features needed to be addressed.

Since the determining factor of the representativeness of synthetic data created was the average sampling rate in the original data, our solution proposed grouping the features into subsets based on their observation frequency. Again, we determined the strategy for Feature Set 1 as the most clinically relevant feature set and then replicated it for the other two. For simplicity, we split the features into low, medium and high-frequency subsets around the following criteria:

- **Low Frequency Subset:** average frequency < 1 in 6-hour window

- **Medium Frequency Subset:** 1 < average frequency < 3 in 6-hour window

- **High Frequency Subset:** average frequency > 3 in 6-hour window

This split the features in Feature Set 1 as follows:

- **Low Frequency:** PH (Arterial), Arterial $CO_2$ Pressure, Arterial $O_2$ pressure

- **Medium Frequency:** Inspired $O_2$ Fraction, Tidal Volume (observed), Minute Volume, Peak Insp. Pressure, Tidal Volume (spontaneous)

- **High Frequency:** $O_2$ saturation pulseoxymetry, Respiratory Rate

We then split the patients into the train and test sets based on the original stratified split earlier in the chapter. The first problem was that not all patients had data for all features, particularly the low observed features. This meant each subset had a different number of patients for its respective train and test set than the original dataset. The low (1,399 train and 350 test), medium (3,722 train and 930 test) and high (3,760 train and 940 test) all varied compared to the original 3,760 patients in the full training set and 941 in the test set. Remember, the primary goal of this exercise is to minimise the amount of synthetic data created. As such, where a patient had no data for a feature, we were reticent to sanction creating entirely synthetic data to include all patients, which would undoubtedly impinge model training. Thus, we decided to employ masking.

Masking is a technique used in machine learning to tell the model to ignore specific values in the input data [120]. The eventual LSTM and TCN model architectures designed to handle these feature subsets would need to detect where values have been masked and selectively ignore them when propagating data through the model. The simplest way to achieve this is to set any values to be masked as NaN.

We decided to maintain all patients throughout each subset to avoid further complexities and dynamic masking in the resultant model. To achieve this, if the patient had no data for a feature, a value was created within the window at an arbitrary time point and set to NaN. This maintains the patient in the subset but provides a platform to fill with NaN values for that feature and mask when passed through the model. This was applied to both the training and test sets.

Next, we had to determine the resampling and interpolation approach. This is pivotal as it determines the proportion of synthetic data created, hopefully somewhat counterbalanced by masking.

Naturally, the target sampling frequency needed to be apropos to each subset while ensuring sufficient data to train the temporal models. There exists no recommended strategy for this. Thus, to keep it simple, we chose the following resampling rates incorporating clinical expertise:

- **Low frequency:** every 2 hours (sequence length = 4)

- **Medium frequency:** every 1 hour (sequence length = 7)

- **High frequency:** every 30 mins (sequence length = 13)

The resampling and interpolation logic was then implemented as follows. Importantly, where a patient had no data for a specific feature, we did not impute that data but set all time steps for that feature as NaN for later masking.

- **Calculate start and end values**

  - Compute mean values for each feature in the training set within the first and last halves of the target sampling window, e.g. for the low-frequency subset, the first and last hour

- **Fill in the Start and End values**

  - If a value exists at 0 / 360 minutes, set as the respective start or end value

  - Else, check for a value within the first/last half of the target sampling window

  - If found, use the earliest / latest value within this window

  - Otherwise, use the mean start/end value calculated from the training data (applied to both train and test data)

- **Resample and interpolate**

  - Resample data to the initial interval (1 minute) to ensure a uniform timeline

  - If there is any original data for a patient, apply linear interpolation to fill missing values

  - If no data is present for the feature, fill the entire series with NaNs

  - Resample the data to the target interval (e.g., 30 minutes) to obtain the desired frequency

This ensured all patients had values for all relevant time steps in each feature subset, whether actual values or NaNs, to be later masked.

Next, features were scaled using min-max scaling between 0 and 1 as before and sequences in the form of 3D arrays were created for each feature subset to capture the different sampling rates for LSTM/TCN input. The process was repeated for Feature Sets 2 and 3, with the subset groupings shown in Appendix Tables C.2 and C.3.

For Feature Sets 2 and 3, categorical features (GCS—Eye Opening, GCS—Motor Response, Richmond-RAS Scale) needed to be dealt with. In collaboration with Dr Murali, the decision was made to treat these values as numerical values to reduce dimensionality and recognise that these features are inherently numerical. However, these features represent scores concerning a patient's consciousness or sedation state and thus must be integer values. Therefore, we set values for these features to whole numbers for clinical relevance and ensured they were within the bounds of the respective scoring system. Before min-max scaling, we rounded and clipped the RAS score between -5 and +4 to limit the values to clinically interpretable scores [121]. Similarly, GCS – Eye Opening and Motor Response were rounded and clipped between 1 and 4 and 1 and 6, respectively [122]. Features were then scaled, and sequences created for model input.

Note that since the numerator and denominator were sampled at different rates, so we did not create $SpO_2$:$FiO_2$ and P/F ratio features in the new strategy.

**Implementing a fused decision architecture for LSTM**

Each feature subset has been resampled to a bespoke rate that ideally facilitates a better representation of the original data while balancing data density and data synthesis. Therefore, the subsets have different sequence lengths and number of features, so the shape of each subset input data differed. Our previous LSTM and TCN architecture for dynamic data was designed to handle all sequences with the same length and number of features, rendering them redundant.

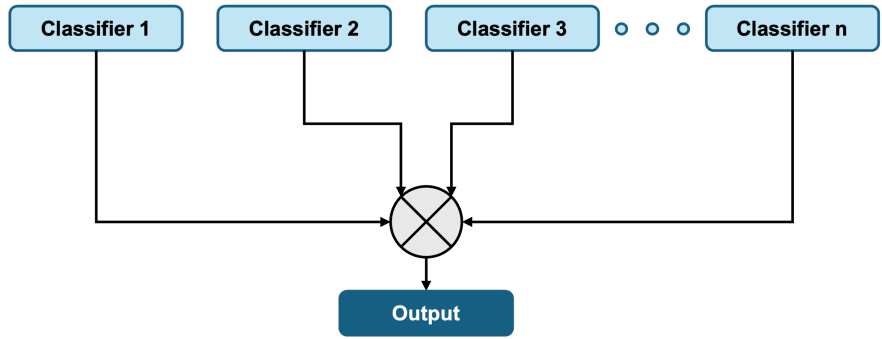

Figure 4.6: Decision fusion. All classifiers process input data in parallel and combine their outputs via decision fusion to provide the final output. Adapted from Chandola *et al.* [123]

A fused decision system was naturally the primary choice for handling the new input data. Simply, fused decision systems combine outputs of multiple classifiers to achieve a consensus decision to enhance the overall classification performance [124, 125]. In this instance, a parallel decision fusion system was required, where each subset is passed through a separate LSTM or TCN, and each output is fused to result in a final prediction, as outlined in Figure 4.6 taken from Chandola *et al.* [123].

As before, we started with the LSTM. The Fused LSTM architecture was implemented in PyTorch and consists of the following key components (highlighted in Figure 4.7).

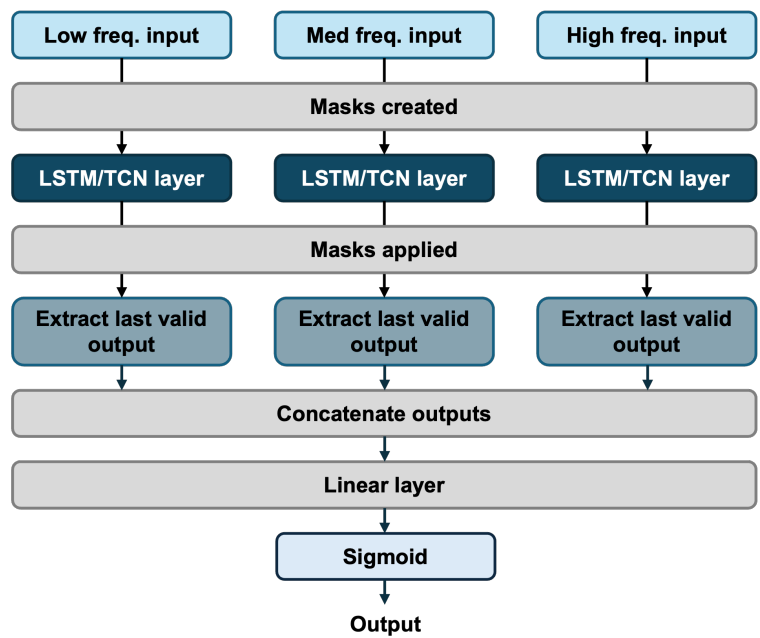

Figure 4.7: Novel architecture of the Fused LSTM/TCN model for extubation failure prediction. Input data from three distinct feature subsets are each passed through a separate LSTM/TCN layer. Before processing, each sequence undergoes masking to missing values (NaNs), which are set to zero for computation. The LSTM/TCN outputs are masked, and the last valid output for each sequence is extracted and concatenated. A fully connected layer fuses the outputs with a sigmoid function applied to generate the binary classification output.

- **LSTM layers**

  - The model contains three LSTM networks, each processing sequences from one feature subset. Each LSTM network is essentially the same as that implemented earlier in this chapter but with bespoke input dimensions reflecting each of the subset sequence shapes

  - Each LSTM layer's hidden state and cell state are initialised to zero

- **Masking**

  - In the forward pass, prior to passing data into the LSTM layers, masks are created by identifying NaN values and then setting those values in the data to zero to ensure data can pass through the model

- **Sequence Processing**

  - Each feature subset is passed through its respective LSTM layer

  - After this step, the created masks are mapped to the outputs to zero out the values corresponding to the original NaN values, hence missing data. This ensures only valid data is taken into consideration for the final prediction

  - For each sequence, the last valid output in the time sequence is extracted

- **Fused Decision**

  - The outputs of the three networks are concatenated and passed through a fully connected layer to reduce to a single output dimension

  - The final output is passed through a sigmoid activation function to generate the extubation failure probability

A decision needed to be made regarding whether to have local hyperparameters for each layer or a global set of hyperparameters. While having individual hyperparameters for each layer would likely help produce the best possible performance, it would require significant computational resources, which were not available in this project. We used the same global set of hyperparameters as before (Table 4.3) and tuned them using Bayesian Optimisation.

All other elements of the LSTM training, i.e. early stopping, dropout, L2 regularisation, sampling method, and loss, were identical. Bayesian Optimisation was run for 100 trials to determine the following optimal combination: 'hidden_dim': 64, 'layer_dim': 2, 'dropout_prob': 0.0, 'learning_rate': 0.001, 'batch_size': 64, 'num_epochs': 40, 'sampling_method': 'normal', 'loss': 'normal', 'weight_decay': 1e-05. The final model was trained and evaluated on the test set: Accuracy: 0.6982, Precision: 0.6812, Recall (Sensitivity): 0.1521, F1 Score: 0.2487, AUC-ROC: 0.6567. Specificity: 0.9652 and Figure 4.8.

Model performance was no longer completely biased. Precision is over double, recall and F1 scores are much lower, but AUC-ROC and Specificity are higher. Looking at the confusion matrix, the model can now somewhat differentiate between success and failure. While the model is still biased towards success, not all patients are predicted to be in the same class. Moreover, the AUC-ROC being higher and above 0.5 suggests the model can predict better than random chance. Thus, ostensibly, the subset pre-processing and fused decision model has undoubtedly made a positive impact, proving our hypothesis that the issue was within the nature of the data and how it was pre-processed.

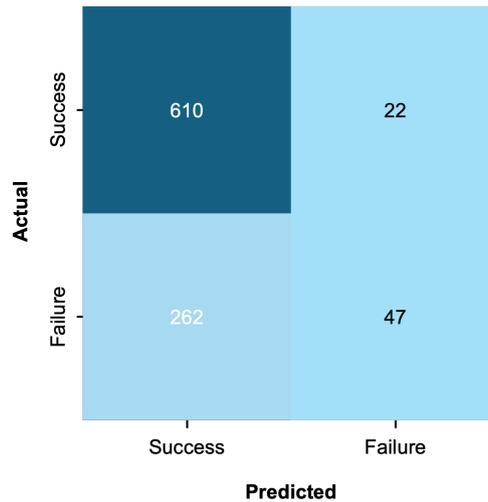

Figure 4.8: Confusion matrix for Fused LSTM post novel pre-processing

### Implementing a fused decision architecture for TCN

To confirm that this change was not model-specific and due to our bespoke pre-processing strategy, we developed a fused decision TCN architecture.

Like the Fused LSTM, we will implement a model that utilises three separate TCN layers (made of Temporal Blocks) as we have already created. The architecture of the Temporal Block and TCN is the same as previously described. However, in the forward pass through each TCN layer, instead of a linear layer, a 1x1 convolution is applied, and the output is fused with that of the other TCN layers. In the Fused TCN architecture, masks are created, and sequences are processed as with the LSTM model. The last valid output for each sequence is obtained from the output of each TCN layer and concatenated, fused and passed through sigmoid activation for binary classification. The architecture is summarised in Figure 4.7.

Once again, we decided to use the same global set of hyperparameters (Table 4.4) rather than local parameters for each TCN layer in the interest of computational efficiency. Training included gradient clipping and measures to prevent overfitting, as described previously.

Hyperparameters were tuned with 100 trials of Bayesian optimisation and determined to be: 'num_channels': '16,32', 'kernel_size': 4, 'dropout_prob': 0.2, 'learning_rate': 0.001, 'batch_size': 256, 'num_epochs': 10, 'sampling_method': 'normal', 'loss': 'normal', 'weight_decay': 0.0001. The AUC-ROC on the validation set for this combination was 0.5756.

Upon evaluation with the test set, we see similar performance to the Fused LSTM with a bias towards success but not a complete one: Accuracy: 0.6865 Precision: 0.6750, Recall (Sensitivity): 0.0874, F1 Score: 0.1547, ROC AUC: 0.6116 and Specificity: 0.9794 and Figure 4.9.

Data for Feature Set 2 and 3 were also pre-processed into subsets using the same thresholds as Feature Set 1 and run through the fused architectures, and full results are presented in Chapter 5.

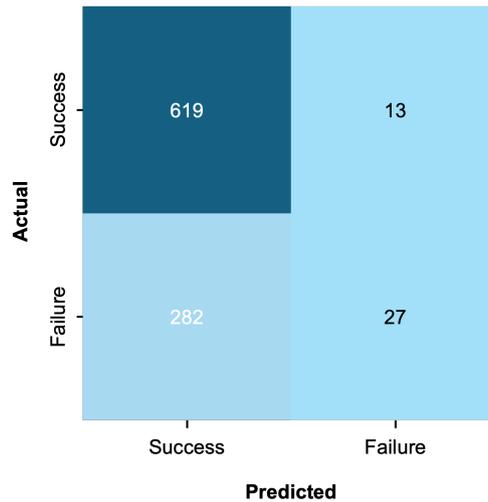

Figure 4.9: Confusion matrix for Fused TCN post novel pre-processing

**Adapting the fused architecture for the inclusion of static data**

It was clear that simply passing the static data through a linear layer was not informative enough to improve the performance of dynamic data in the vanilla models. As such, we wanted to be able to derive more information from the static data.

Inspired by Catling *et al.* [77], we used a Feed-Forward Neural Network (FFNN) to process static data, creating Fused LSTM-FFNN and Fused TCN-FFNN architectures.

For both LSTM and TCN, all elements of the Fused systems used to process dynamic data remained the same. However, both models incorporate a fully connected FFNN consisting of multiple linear layers followed by an activation function (ReLU, Tanh, Sigmoid or Leaky ReLU) and dropout. The number of layers, units per layer, activation function and dropout rate for the FFNN were all additional hyperparameters to be tuned alongside the LSTM and TCN hyperparameters (Appendix Tables C.4 and C.5, respectively). The output of the FFNN is transformed into a feature vector that matches the dimension of LSTM outputs and is concatenated. The final output is passed through a sigmoid activation function for binary classification. Fused LSTM-FFNN and Fused TCN-FFNN architectures are summarised in Figure 4.10.

Following Dr Murali's recommendation, ethnicity was categorised into the MIMIC-provided subgroups of ASIAN, BLACK, HISPANIC, WHITE, and OTHER to reduce dimensionality. Ethnicities were one-hot encoded and incorporated into the static data previously created.

**LightGBM − the baseline model**

LightGBM was observed to have one of the highest performances in the literature and is computationally efficient, so it was chosen as the non-temporal baseline in this study. The lightgbm package in Python was used to create the model [53].

Data was aggregated for all features over the 6-hour time window. For simplicity, we used mean aggregation to represent the average behaviour of a feature across the period. The train and test data was taken after general pre-processing prior to LSTM/TCN-specific pre-processing. Where a patient had no values for a feature, we mean imputed across the patient population to maintain consistency using average values. The means were calculated using the training set and applied to the train and test set to avoid data leakage, with the same set of 3,760 training patients and

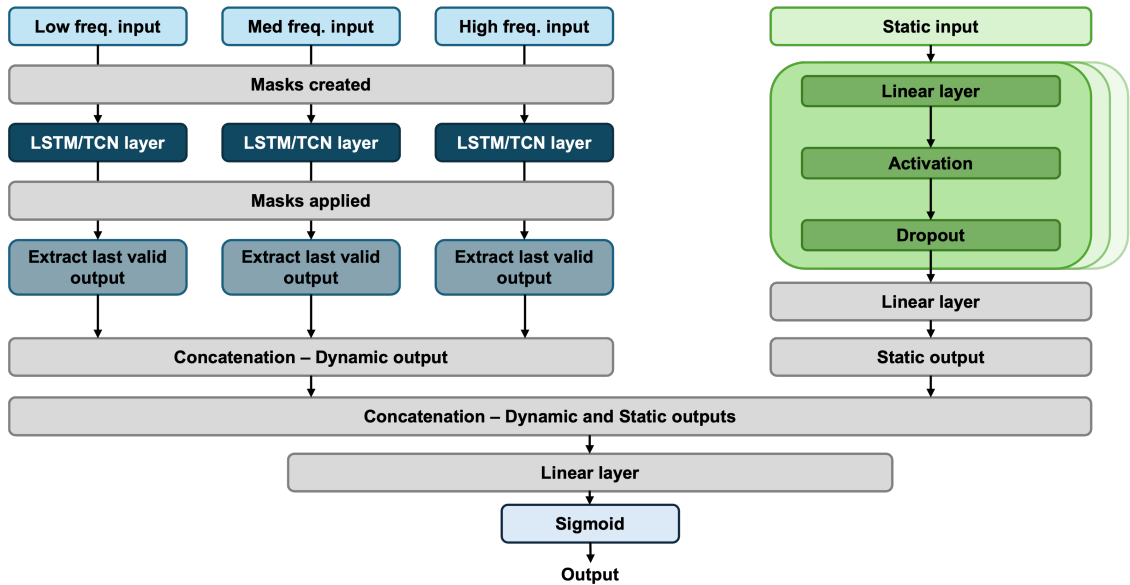

Figure 4.10: Architecture of the novel Fused LSTM/TCN-FFNN model for dynamic and static data processing. The model processes three distinct feature subsets of data, each with different resampling rates through corresponding LSTM/TCN layers. Masks are applied to handle missing values in the dynamic data, and the last valid outputs are extracted. Additionally, static data is passed through a fully connected feed-forward neural network (FFNN). This network consists of linear layers, activation functions and dropout layers. The outputs of the TCNs and FFNN are concatenated and passed through a final linear layer, followed by sigmoid activation to produce the final binary classification output

941 test patients used for consistency. Note that the features are not split into subsets as with the LSTM/TCN input data. This is fundamentally because LightGBM data needs to be aggregated; thus, the bespoke sampling rates become redundant. Moreover, $SpO_2$:$FiO_2$ and P:F ratios were not calculated to align with the dynamic data used.

This same pre-processing was applied to all feature sets for the dynamic data. Where static data was also included, the features were appended to the tabular dynamic data for each patient.

The comprehensive grid of hyperparameters used for tuning is shown in Appendix Table C.6.

As before, a Bayesian optimisation run with 100 trials was used, with AUC-ROC as the primary validation metric for five-fold cross-validation. Early stopping was implemented as a callback, with the number of stopping rounds set to 10. The final model was trained and evaluated on the test data.

This process was repeated for all feature sets. Once again, the categorical features in Sets 2 and 3 needed to be correctly handled. Aggregating GCS and RAS scores using mean aggregation is clinically meaningless. Instead, we used the mode as this would represent the most common score assigned to a patient over the 6-hour window, which is clinically more interpretable, as corroborated by Dr Murali. Where patients had no values for these categorical features in either the training or test set, it was filled with the mode of the training dataset to avoid data leakage.

**Model interpretation**

Feature ablation was employed to ascertain feature importance. For each subset (low, medium, and high), each constituent feature is systematically removed one at a time, and the model is trained on the ablated training data using the optimal hyperparameters that were previously determined. The ablated model is then evaluated on the test set, and the change in performance metrics compared to the baseline performance trained on all features is recorded [123]. Once again, AUC-ROC was used as the primary metric.

The same methodology was implemented with models processing static and dynamic data. Each static feature was systematically removed to create an ablated model. The results were then visualised to highlight the importance of each feature to the AUC-ROC score.

**Ensemble methods**

We created ensemble predictions to determine whether combining the outputs of multiple base models can achieve a better performance than any individual model.

In this study, we employed averaging and stacking. More sophisticated methods, such as AdaBoost, were considered, but we decided to implement more straightforward methods for simplicity and computational resource usage.

Averaging involves taking the mean of the probabilities for each patient across all classifiers and evaluating the performance with a threshold of 0.5. Whilst stacking uses a meta-model to learn how to best combine the predictions from base models with a simple Logistic Regression model used in this study [126].

The results of these ensemble methods, applied across all three feature sets, along with those of the final LSTM and TCN fused decision and LightGBM models, are presented and analysed in Chapter 5.

## 4.3 Implementation of this thesis

The link to the Github repository for this project can be found at: `https://github.com/YoosoofsahA/msc_computing_project`.

The project was constructed and run using Google Colaboratory, which enables the development and execution of Python notebooks within a cloud-based environment. All analyses were carried out in Python version 3.10.12, and Pandas, Numpy, Scikit-learn, and Matplotlib packages were extensively used for primarily data-based tasks and PyTorch for building and training machine learning models. All experiments were run on an Intel(R) Xeon(R) CPU @ 2.20GHz and an NVIDIA Tesla T4 GPU with 15360 MiB memory provided by the Colab environment.

# 5

# Results of final model architectures

The following section outlines the results obtained from the final models, whose architectural design was guided by the nature of the data once an effective pre-processing strategy had been settled on.

Table 5.1 reports all standard evaluation metrics (Precision, Recall, F1 score, Specificity and AUC-ROC) for the LSTM, TCN, LightGBM, and ensemble models across all feature sets to provide a holistic view of the performance of the MIMIC-IV retrospective cohort.

The table is structured with each row representing a different model and input data combination, while the columns highlight the respective metrics for each feature set. It should be noted that accuracy is deliberately omitted since the underlying dataset is imbalanced.

Given this study's extensive nature, multiple comparative points of analysis are necessary. These include model architecture, the inclusion of static data, and comparison across feature sets. As such, this final results section is organised to address each of these points systematically.

**Comparison of temporal architectures on time-series data**

This study developed Fused LSTM and TCN models to process time-series data split into subsets based on sampling frequency. LightGBM was employed as a baseline model based on its superior performance in the literature and computational efficiency.

Here, we will highlight the models' performance on the dynamic data, facilitating interpretation of their ability to capture temporal dependencies.

Focusing on the models designed to capture temporal dependencies (LSTM and TCN), as prioritised in this study's objectives, we compare their performance to determine whether the choice of temporal architecture is critical.

Initially, looking at Feature Set 1 as the most clinically relevant set of features, we can see that the performance of the two models is very similar overall, with values generally being in a narrow range of each other. Discrepancies can be observed in precision, where the LSTM model slightly outperforms the TCN (0.6812 vs. 0.6750) and recall (0.1521 vs. 0.0874, respectively). This suggests that the LSTM may be better at identifying true positives. Additionally, on the primary metric of AUC-ROC, the LSTM performs marginally better (0.6567 vs. 0.6116) implying superior discrimination between classes.

Table 5.1: Evaluation metrics for all models. Results are presented for all models across all feature sets. Metrics presented are: P = precision, R = Recall, F1 = F1 score, Sp = Specificity and AUC-ROC

| Models | Feature Set 1 | | | | | Feature Set 2 | | | | | Feature Set 3 | | | | |
|---|---|---|---|---|---|---|---|---|---|---|---|---|---|---|---|
| | P | R | F1 | Sp | AUC-ROC | P | R | F1 | Sp | AUC-ROC | P | R | F1 | Sp | AUC-ROC |
| Fused LSTM (dynamic only) | 0.6812 | 0.1521 | 0.2487 | 0.9652 | 0.6567 | 0.4362 | 0.4757 | 0.4551 | 0.6441 | 0.6994 | 0.4659 | 0.4207 | 0.4422 | 0.7642 | 0.6302 |
| Fused LSTM-FFNN (static + dynamic) | 0.7708 | 0.1197 | 0.2073 | 0.9826 | 0.6561 | 0.6106 | 0.2233 | 0.3270 | 0.9304 | 0.6742 | 0.4915 | 0.4660 | 0.4784 | 0.7642 | 0.6642 |
| Fused TCN (dynamic only) | 0.6750 | 0.0874 | 0.1547 | 0.9794 | 0.6116 | 0.6338 | 0.1456 | 0.2368 | 0.9589 | 0.6178 | 0.7200 | 0.0583 | 0.1078 | 0.9889 | 0.6226 |
| Fused TCN-FFNN (static + dynamic) | 0.4268 | 0.5566 | 0.4831 | 0.6345 | 0.6300 | 0.5079 | 0.4175 | 0.4583 | 0.8022 | 0.6696 | 0.4188 | 0.5922 | 0.4906 | 0.5981 | 0.6401 |
| LightGBM (dynamic only) | 0.4340 | 0.3722 | 0.4007 | 0.7627 | 0.5674 | 0.4975 | 0.3402 | 0.3898 | 0.8418 | 0.5811 | 0.5211 | 0.3204 | 0.3968 | 0.8560 | 0.5882 |
| LightGBM (static + dynamic) | 0.4220 | 0.4466 | 0.4340 | 0.7009 | 0.5825 | 0.5795 | 0.3301 | 0.4206 | 0.8829 | 0.6065 | 0.4975 | 0.6236 | 0.3922 | 0.8402 | 0.5819 |
| Ensemble – averaging | 0.6058 | 0.2039 | 0.3051 | 0.9351 | 0.5695 | 0.4835 | 0.1424 | 0.2200 | 0.9256 | 0.5340 | 0.4962 | 0.2104 | 0.2955 | 0.8956 | 0.5530 |
| Ensemble – stacking | 0.6500 | 0.2063 | 0.3133 | 0.9444 | 0.5754 | 0.6842 | 0.2063 | 0.3171 | 0.9524 | 0.5794 | 0.7200 | 0.2857 | 0.4091 | 0.9444 | 0.6151 |

For Feature Sets 2 and 3, the LSTM again has higher AUC-ROC, reinforcing a better discriminatory performance. Precision is lower for both feature sets in LSTMs, whereas recall is much higher, leading to a higher F1 score.

**Comparison of temporal architectures to the baseline**

For Feature Set 1, LightGBM (0.4340) lags behind LSTM and TCN in precision (0.6812 and 0.6750, respectively). This is also seen for AUC-ROC, where LightGBM has a lower value of (0.5674). Interestingly, LightGBM outperforms both temporal models in recall (0.3722 vs. 0.1521 LSTM and 0.0874 TCN), indicating that it may be better at capturing true positives, albeit with a trade-off in precision.

Across the larger feature sets, the performance of LSTM and TCN relative to LightGBM presents a more nuanced picture, with both having comparatively mixed performance across metrics. However, AUC-ROC is higher for temporal models compared to the baseline across all feature sets.

**Comparing performance when static data is included**

Static data provides context to extubation failure decisions in real-world clinical scenarios. If a performance improvement is seen consistently when static data is included, then it is likely that this data adds relevant context to the model.

Starting with the Fused LSTM, for Feature Set 1, precision improves when static data is incorporated (0.7708 vs. 0.6812). Recall (0.1197) and AUC-ROC (0.6561) are marginally lower. For Feature Sets 2 and 3, improved precision is seen. AUC-ROC drops for Feature Set 2 but increases for Feature Set 3. Adding static data to LSTM seems to have a mixed impact on performance as precision tends to increase but has varying effects on recall and AUC-ROC.

Looking at the Fused TCN, incorporating static data increases AUC-ROC (0.6300 vs. 0.6116). Furthermore, recall drastically improves (0.5566 vs. 0.0874), but precision drops (0.4268 vs. 0.6750). Similar patterns are seen across the larger feature sets, particularly with improved AUC-ROC and recall. Ostensibly, TCN models benefit more from the context of static data in terms of recall and AUC-ROC.

When the Fused LSTM-FFNN and Fused TCN-FFNN are compared to the LightGBM with static and dynamic data, an improvement in AUC-ROC is seen compared to the baseline. The impact on other metrics seems to vary, with most values somewhat comparable.

**Comparing the effect of adding more features**

Examining model performance across feature sets gives us insight into whether adding more features is informative for model prediction.

In the Fused LSTM, transitioning from Feature Set 1 to Feature Set 2 results in a slight increase in AUC-ROC (0.6994 vs 0.6567) and a notable drop in precision (0.4362 vs 0.6812). Recall increases (0.4757 vs 0.1521), leading to an improved F1 score (0.4551 vs 0.2487). This demonstrates the precision-recall trade-off and suggests that adding features helps the model capture more failure cases, albeit with less confidence. Analogous trends are seen when static data is incorporated. Increasing from Feature Set 2 to 3 caused recall to drop slightly (0.4207) and AUC-ROC marginally to fall (0.6302), as is also the case with the static data (0.6642 vs 0.6742).

The Fused TCN model trained on only dynamic data exhibits similar trends, with recall increasing from Feature Set 1 to Set 2 (0.1456 vs 0.0874), precision decreasing (0.6338 vs 0.6750), and little change in AUC-ROC (0.6178 vs 0.6116). Interestingly, the pattern is reversed with the static models as precision increases, recall drops, and AUC-ROC partially improves. From Feature Set 2 to 3, with the dynamic-only models, precision increases while recall falls, and the opposite pattern is seen for the static models.

**Performance of ensemble learning**

As a follow-up, we wanted to see whether ensemble learning could boost the performance of our models by combining their predictive power and comparing that to the performance of the individual models. The focus here is on analysing the temporal models' performance; thus, LightGBM will not be compared against the ensemble models.

Both averaging and stacking ensemble models had lower AUC-ROC than all temporal models. In Feature Set 1, averaging (0.5695) and stacking (0.5754) were lower than all LSTM and TCN models, whose values were over 0.6. The other metrics were more in line. The trend is generally seen across the other two feature sets, where precision, recall, and F1 are typically comparable, but AUC-ROC is consistently lower.

**Plotting ROC curves**

Plotting ROC curves for each model (true positive rate vs. false positive rate) highlights the trade-off between sensitivity and recall. Curves with a larger area to the diagonal line (45 degrees) display a higher AUC-ROC and are better discriminative models.

The ROC curves for models trained on Feature Set 1 is shown in Figure 5.1, with the curves for Feature Sets 2 and 3 outlined in Appendix Figures D.1 and D.2.

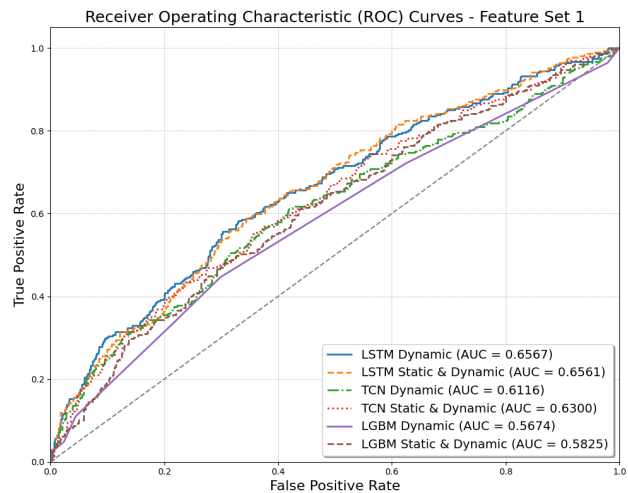

Figure 5.1: ROC curves for LSTM, TCN and LightGBM models trained on Feature Set 1

We can see little difference between the area under the curve for each model, reflecting the metrics in Table 5.1. This is seemingly the case for all feature sets, underscoring the similarity in model performance regardless of architecture, feature set and static data inclusion.

**Model Interpretation**

To understand the factors driving model predictions, we employed Feature Ablation. This technique removes one feature at a time systematically, trains the ablated model, and evaluates the impact on the primary metric of AUC-ROC.

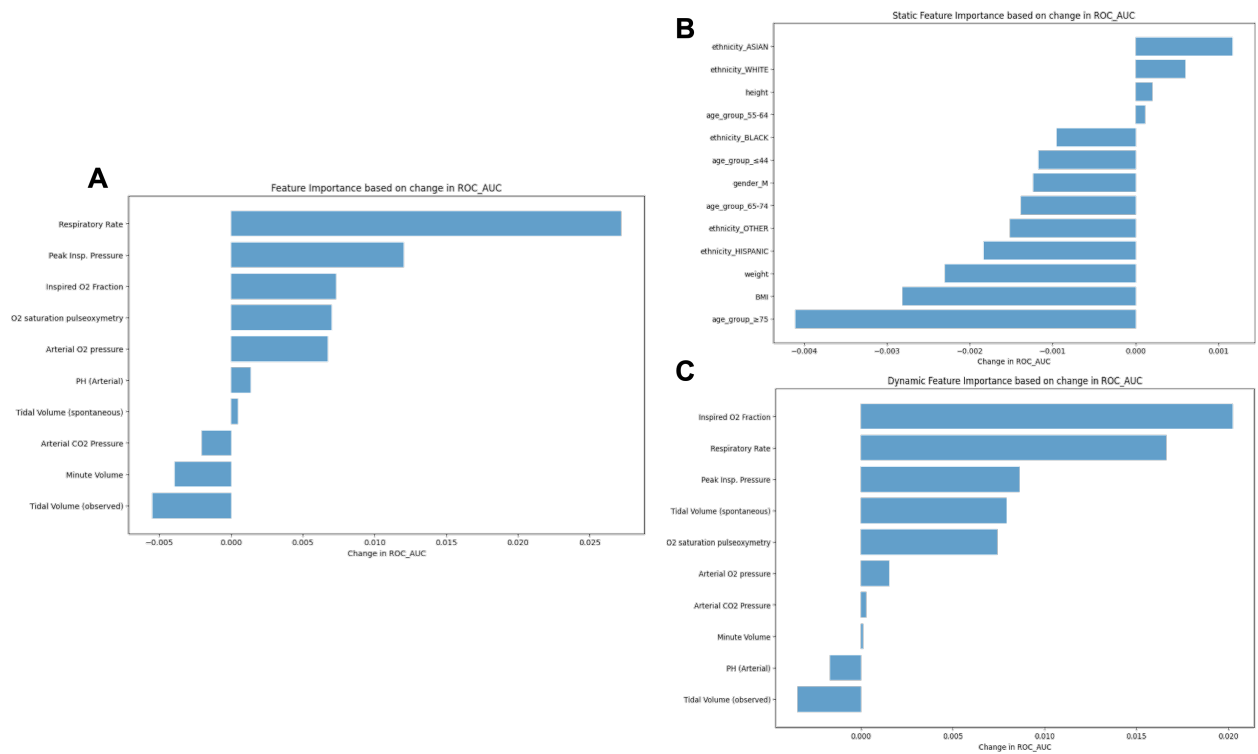

Figure 5.2: Feature ablation values for Fused LSTM on dynamic only data (A) and Fused LSTM-FFNN on static (B) and dynamic (C) features from Feature Set 1. The scores represent the change in AUC-ROC when that feature is ablated

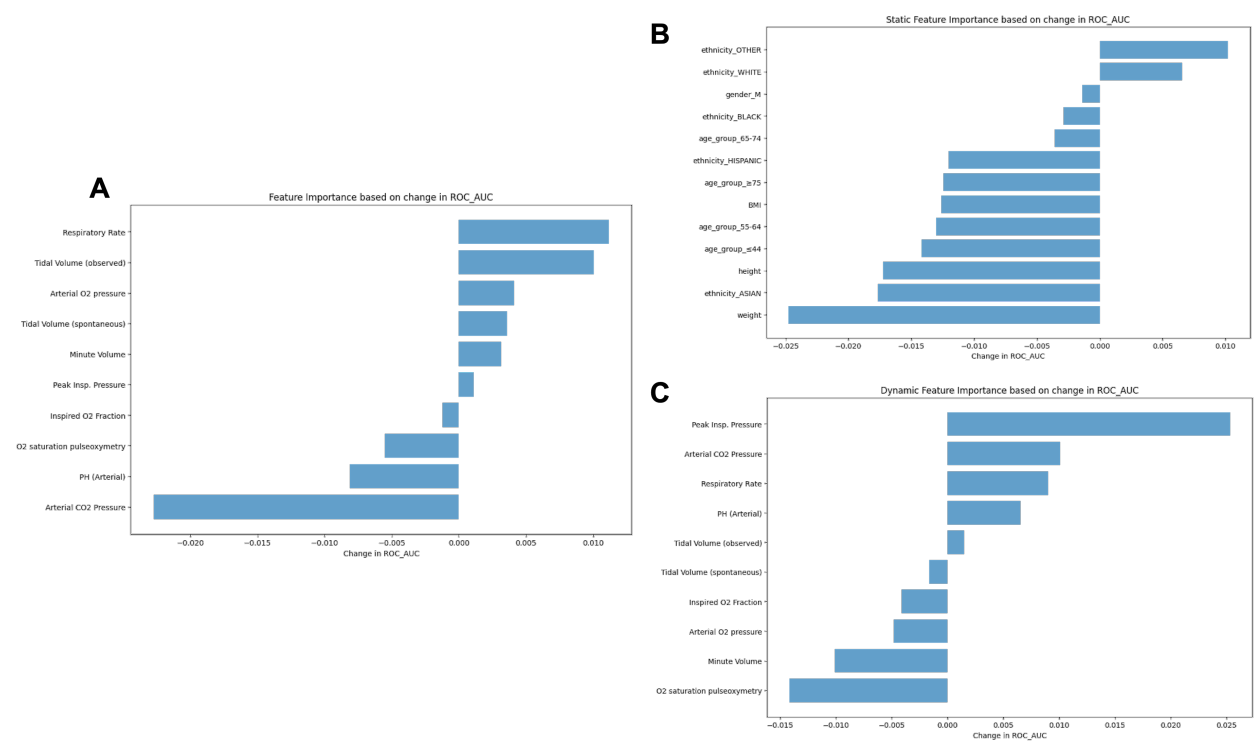

Figure 5.3: Feature ablation values for Fused TCN on dynamic only data (A) and Fused TCN-FFNN on static (B) and dynamic (C) features from Feature Set 1. The scores represent the change in AUC-ROC when that feature is ablated

The results are presented as bar charts plotting each feature and the change in AUC-ROC (also known as ROC AUC) resulting from that feature's ablation. The more negative the change in AUC-ROC for a specific feature, the more influential that feature was in model decision-making.

Figures 5.2 and 5.3 show the results of feature ablation for LSTM and TCN models trained using Feature Set 1 (both with and without static data). Given our goal to develop an interpretable model and the clinical importance of Feature Set 1, we prioritise these temporal model results to exemplify model transparency and insights into its decision-making process. Detailed ablation results for the other feature sets are provided in the Appendix (figures D.3, D.4, D.5 and D.6) to support these discussions.

To clarify, when we say that a feature is "important", we mean that its inclusion enhances the model's ability to discriminate between extubation success and failure.

For the LSTM model trained solely on dynamic data from Feature Set 1, respiratory mechanic measurements of Tidal Volume (observed) and Minute Volume emerge as the two most informative features, as their absence exhibits the most significant negative change in AUC-ROC. Conversely, Respiratory Rate and Peak Inspiratory Pressure appear to be the least informative.

Age $\geq 75$ seems to be the most helpful static feature for distinguishing between extubation success or failure, followed by BMI. Generally, ethnicities have varying influence. Interestingly, the importance of dynamic features used in the Fused LSTM-FFNN model seems to differ in the presence of static data. While some features exhibit similar behaviour (Tidal volume and Respiratory Rate), features such as PH (Arterial) have transitioned from a negative change in AUC-ROC to a positive change, indicating increasing importance when contextualised with static information.

Turning to the TCN models, Arterial $CO_2$ Pressure is by far and away the most important feature in the dynamic-only model as it has the most negative change, contrasting with its lesser importance in the LSTM. Respiratory rate is, intriguingly, once again, the least important feature. In the Fused TCN-FFNN model, weight was the most critical static feature – surpassing its importance in the Fused LSTM-FFNN model. Once again, ethnicity shows varying influence. In the dynamic feature set for the Fused TCN-FFNN, we see $O_2$ saturation pulseoxymetry as the most important feature – whereas, in the LSTM, it hindered discriminatory capability.

# 6

# Discussion

Respiratory failure can be fatal. Mechanical ventilation is a cornerstone of effective treatment and deciding to take a patient off the ventilator (extubation) is arguably one of the most important decisions a clinician must make in the ICU. With the risk of an ill-informed choice impacting not only the patient's immediate survival but also their long-term quality of life, the need for assistive tools to process large amounts of complex healthcare data and reliably predict failure of extubation cannot be understated. This is where the advent of machine learning is vital. However, any tool must be able to effectively capture patient trajectories by learning from temporal data. Moreover, transparency of prediction is paramount to earning the physician's trust, with whom the clinical decision ultimately lies.

Most previous work fails to recognise the real-world end goal of predicting extubation failure and for whom the predictive tool is intended. To our knowledge, only one of the peer-reviewed studies focusing on predicting extubation failure employs model architectures designed to capture temporal dependencies. Additionally, model interpretability is not given enough importance. Unfortunately, this seems to result from a lack of clinical validation throughout data processing and model development, rendering the majority of the work completed thus far somewhat inapt and, more disappointingly, not clinically deployable. However, these gaps provided several avenues for further exploration.

In this study, we explored the development of an end-to-end actionable and interpretable prediction system using time-series data to help clinicians make better-informed decisions about extubation failure. This system can be broadly split into three parts, as shown in Figure 6.1: data preprocessing, development of temporal model architectures and interpretability. Our work not only pushes the boundaries of research in this critical domain but also proposes novel approaches that serve as robust foundations for further investigation.

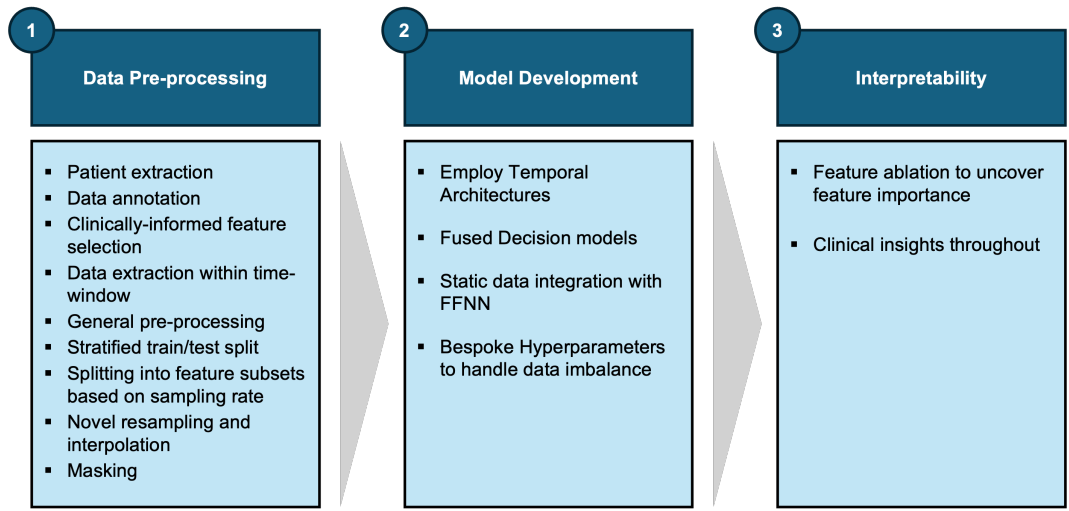

Figure 6.1: Novel end-to-end pipeline to predict extubation failure. The pipeline outlines steps taken in this thesis specifically relating to Data Pre-processing, Model Development and Interpretability

From the beginning, there were several practical challenges that we had to overcome upon review of the literature:

- **Limited Prior Research:** To the best of our knowledge, only one peer-reviewed paper existed where architectures geared to learn temporal dependencies have been used to predict extubation failure (Zeng *et al.* [80]). Therefore, there is no consensus on how architecture should be designed for this purpose, hindering best practice

- **Unexplored Model Architectures**: To our knowledge, no paper explored the potential of Temporal Convolutional Networks (TCNs) for extubation failure prediction, leaving a significant gap in our understanding of suitable model architectures

- **Lack of Standardised Pre-processing:** There is an absence of established guidelines for data pre-processing and feature engineering when using temporal data to predict extubation failure within a fixed window. The lack of standardisation can lead to inconsistencies and inadequate model performance

- **Heterogeneity of approach:** Where common areas are addressed in relevant studies, e.g. annotation of extubation, no consensus exists, which renders evaluation more challenging

As a result, much of what we had to achieve in this study necessitated innovative approaches and validation with clinicians throughout.

The discussion will be organised into the three components of the pipeline.

## 6.1   Data pre-processing

Starting with data pre-processing, as a retrospective study, data extracted from MIMIC-IV inherently posed its own challenges. While there were several time-series features to select from, data recording was inconsistent across features and time, which is common in medical data. Moreover, the size of the database meant that cohort establishment, feature selection, and data extraction strategies needed to be carefully conceived with clinician input.

Our approach to constructing the patient cohort closely resembles practices in literature based on the MIMIC database (Appendix B.11). Inclusion and exclusion criteria were outlined with direct input from Dr Murali to ensure that we had a homogenous, representative population that limited confounding factors. Interestingly, this specific combination of criteria has not been replicated in any literature using MIMIC to predict extubation failure. This is telling of the limited medical basis used in comparable studies and, therefore, where this study excels with the view of ensuring clinical relevance. Moreover, clinical justification was paramount when determining the definition of extubation failure for annotation. Studies investigating this in the past have disparities in their adopted classifications, as highlighted in Appendix A.1 and B.11. As one can see, most studies tend to define failed ventilator liberation as requiring reintubation within 48-72 hours. The definition used in this study is comparable but more comprehensive. Re-intubation and mortality within 48 hours is a standard definition used across the literature. However, the incorporation of a criterion surrounding post-extubation ventilatory support is rarer. To clinicians, a patient requiring NIV such as CPAP or BiPAP indicates that they have not fully recovered and would likely struggle with further intervention, symptomatic of extubation failure. This further highlights the importance of clinical oversight in any ML study as several patients in the literature would likely be misclassified, and, therefore, models led astray. Bringing this into the real-world context, if a model fails to recognise potential failure and the patient is extubated, the consequences of premature removal from ventilation could be life-threatening.

We employed a multifaceted feature selection approach, balancing existing literature insights with real-world clinical considerations. This systematic and clinically informed feature selection process aimed to identify relevant predictive variables, recognising that research practice may not always mirror clinical practice. While several features have been explored in past studies, the lack of a 'gold standard' set underscores the heterogeneity in the domain and the challenges of navigating the vast feature space.

Our cross-referencing strategy enabled us to construct distinct feature sets and prioritise those popular in literature and clinically available. Moreover, excluding Set 4, which comprised features neither popular in the literature nor clinically available, ensured our models focused only on potentially informative features.

To our knowledge, splitting features into sets of varying clinical importance has not been done before in this domain. Most previous work uses standard feature selection strategies such as Recursive Feature Elimination and correlation analysis, with many not even outlining their process explicitly. Our methodology provides a structured, transparent framework to navigate the vast feature space in medical data required to predict extubation failure. The intention of incrementally training models on more features to investigate the impact of the number of features on performance yielded insights that will be reviewed in due course. We strived for a balance between evidence-based selection and clinical pragmatism to enhance our predictions' generalisability. As we advance, studies may explore using primary data and more diverse real-world datasets to refine feature selection further.

We chose to predict extubation failure based on data from the last 6 hours prior to extubation. This was determined in collaboration with clinicians, who typically look at patient data in 6-hour windows for patients under ventilation. When deciding to extubate, it was emphasised that the most recent 6 hours weighed heavily in the decision-making process. Thus, regardless of the quantity of data present in MIMIC in that time window, we prioritised clinical relevance over data abundance.

This starkly contrasts with Zeng *et al.*, who captured data in 4-hour windows during ventilation [80]. From the end-user perspective, receiving a prediction every 4 hours may be too frequent for the clinician to make a meaningful decision. Windowing is a widespread technique when processing time-series data in ML, and its inclusion in the study ostensibly reflects the judgement of an ML researcher and not a learned clinician [127]. Moreover, even looking at the typical studies in the broader field, none seem to recognise the importance of a clinically relevant time window, regardless of whether the data is aggregated. It is hoped that this study can underscore the importance of clinically guided decisions throughout data pre-processing and collection.

An inclusion criterion based on the average observation frequency per feature was set to reduce potential imputation bias in later pre-processing. For Feature Sets 1 and 2, any feature below the threshold of 0.5 was removed, and for Feature Set 3, the threshold was 0.15. Clinical data often has irregular resampling due to the nature of patient monitoring and intervention, and features with low sampling frequencies might be insufficient for reliable predictions without significant imputation. Therefore, eliminating low observed features prior to resampling and interpolating is critical in any study using time-series data in this domain. However, the value of the threshold is up for debate. A balance must be struck between the level of data imputation and the number of features the model is trained on. If the threshold is too high, the model may have too few features and overlook important information, but too many features may introduce noise [60].

The main comparable study by Zeng *et al.* uses a threshold of an average observation frequency of 0.1 in a 4-hour window [80]. However, they can afford a lower threshold since the accumulation of data over many time widows can compensate for occasional sparsity, which was not relevant to this study. As demonstrated, no consensus around a clinically relevant threshold value exists. As such, while based on a decision to balance data synthesis and data density, it is difficult to know whether the thresholds used in this study were optimal. Moreover, the methodology of basing the threshold only on the most clinically relevant feature set and replicating it for others may not be ideal. It was shown that the threshold of 0.5 was insufficient to include a meaningful number of additional features in Feature Set 3 and subsequently had to be lowered to 0.15. It might be the case that the threshold needs to be unique to the feature set to balance data density and synthesis. Needless to say, a significant body of work should be undertaken to analyse model performance while employing different thresholds, potentially involving sensitivity analyses to quantify the trade-off between imputation and feature inclusion. In addition, clinical expertise should be sought to determine meaningful thresholds contingent upon the characteristics and context of each feature set.

Following this, we recognised that the amount of synthetic data would likely be a determining factor. Where synthetic data was imputed, later interpolation would generate artificial temporal patterns. Unfortunately, we were at the mercy of the quality of the medical data, and thus, synthesis was unavoidable. To ensure the magnitude of synthetic data patterns was comparable between the train and test set, we stratified the train/test split by the proportion of required synthetic data. If the training data were more artificial, the model would be biased towards the ill-representing patterns and unable to generalise to the test data. Admittedly, the calculation was based on a maximum resampling of 30 minutes, which does not apply to all data subsets created and likely requires a more refined strategy. However, this stratification was carried out before deriving the subset and fused architecture strategy, so it represents the most informed decision we could make at the time. Emphasis should be placed on this strategy's innovative nature, which has not been employed in the literature to the best of our knowledge. This reinforces the gap in current research. Most studies seem to disregard synthetic data's influence on model bias. We call for future studies to incorporate similar or more refined strategies to ensure a balanced distribution of synthetic data when splitting out the train and test sets.

A standard statistical approach was used to filter outliers using the mean $\pm$ 3 standard deviations where MIMIC values were not provided. We acknowledge that we should have developed these bounds in collaboration with clinicians for a more robust strategy. Their expertise in recognising typical values seen for adult patients in the ICU would ensure that the remaining values are clinically relevant. Unfortunately, due to limited time with clinicians and their demanding schedules, we focused their invaluable insights on more crucial areas that required medical domain knowledge, such as patient extraction, feature selection and result interpretation. While the current approach was sufficient for initial data cleaning, future iterations would benefit from clinically informed outlier detection.

## 6.2 Model development

In this study, we chose to use LSTM and TCN models to capture temporal dependencies and best learn patient trajectories in the ICU. These are not the only architectures that can handle time-series data, and naturally, further work would benefit from a more expansive exploration. However, these models are well established across several domains in healthcare and beyond when the task requires prediction based on temporal patterns from the past, such as stock prediction. This provided several valuable precedents for our models.

What should be emphasised, however, is the striking observation of the lack of exploration of such temporal architectures for extubation failure prediction. We highlighted how, to the best of our knowledge, only one peer-reviewed paper employs closely related LSTM and GRU models for this purpose. This accentuates the novelty of our work, particularly since the creation of a TCN model of any kind to predict extubation failure has not been achieved before. Moreover, our study uniquely compares two fundamentally different architectures (LSTM and TCN) trained on the same data to gain insights into their relative strengths and weaknesses in this domain. The choice to develop simple architectures was deliberate, as it aimed to place more weight on the data's quality and nature rather than the models' complexity.

LightGBM as a baseline was strategically selected to determine whether the complexity of temporal models translated to improved predictive performance. Speed and accuracy are vital in a clinical environment, and LightGBM is specifically designed to be lightweight and, thus, faster [53]. Knowing that clinical data is sometimes incomplete, if a model that can take aggregated data has comparable or superior performance than one whose data would require significant pre-processing to impute missing values, it is challenging to recommend deploying models that require substantial computational resources (LSTM and TCN).

Our model development process was one of continuous iteration, and it was not until we controlled the amount and influence of synthetic data by developing a novel pre-processing methodology that the models could discriminate between extubation failure and success.

Initially, we opted to resample all features to the same 30-minute interval regardless of the original observation rate. To achieve this, linear interpolation was chosen where, in fact, most literature on extubation failure, including Zeng *et al.*, use forward fill to impute data or missing values [80].

While briefly considered for simplicity, this is not clinically relevant as it would result in several values being repeated across consecutive time steps, creating artificial plateaus that misrepresent the dynamic nature of real-world patient trajectories.

Since no unified approach exists to creating medical sequence data for temporal models, we had to develop a novel, clinically validated approach. Our logic used to create start and end values is entirely unique. Still, crucially, we ensured that choices such as using a half-window cut-off (15 min for example) and taking average values were clinically plausible.

It should be noted that in real-world terms, taking an average value across a patient population is sub-optimal. Patients are inherently unique, with differing comorbidities, demographics and medical backgrounds, and, therefore, an average is not representative of the nuances of medical performance. However, this logic was designed to mask the inadequacies of medical data as we could not extract time-series values from MIMIC-IV at regular intervals without significant pre-processing. Ideally, future studies will have access to more consistently sampled data and can confidently investigate patient-specific imputation techniques or more advanced interpolation strategies to capture individual patient nuances better.

A simple LSTM model was developed to prevent overfitting (early stopping, L2 regularisation and dropout were also employed). Recognising the imbalance in the underlying data, we implemented a novel approach of creating bespoke hyperparameters to optimise training and avoid potential biases. This innovative approach stands in contrast to prevailing practices in the literature, which typically implement either under-sampling or oversampling (such as Chen *et al.* employing SMOTE [50]) or neither – which is concerning given that all datasets in the literature reviewed were imbalanced. However, the optimal strategy for a particular data set is unclear. Wongvorachan *et al.* [128] highlighted that the ideal sampling strategy depends on the level of imbalance. Through our approach, we tuned for the sampling strategy that uniquely provided the best performance on the validation set rather than a pre-determined, uninformed choice. We showcase our commitment to reducing model bias by moving beyond pre-determined sampling strategies. However, we must recognise the additional computational complexity added as a result. Adding extra hyperparameters increases the cross-validation time and, therefore, computational resources required. Further work here should take this into consideration, possibly even developing adaptive sampling methods that dynamically adjust to the inherent data during training.

No matter what we tried, be it more comprehensive hyperparameter tuning with Bayesian Optimisation, changing the resampling strategy to better reflect patient trajectories, or including static data context, the model was entirely biased to one class or the other. This issue of bias with LSTM architectures is not uncommon, which has been shown in work by Al-Selwi *et al.* [129], to sometimes struggle to learn long-term dependencies due to its inherent recurrent nature, particularly when the sequence length is not optimal. Bayer [130] highlights that LSTMs can still encounter the vanishing gradient problem outlined earlier, where repeated multiplications of their recurrent weights can cause gradients to vanish and bottleneck learning long-term dependencies effectively. Moreover, Chen *et al.* [131] emphasise that LSTMs can perform poorly when forecasting events using imbalanced time-series data. Ostensibly, the inherent design of LSTMs, while powerful in learning temporal patterns, may struggle to comprehensively learn long-term dependencies in this scenario where the inherent data is imbalanced, or the sequence length may not be optimal. Thus far, we had only experimented with LSTM architecture.

Therefore, we repeated the process with a novel TCN architecture. The model exhibited the same biased behaviour even with a completely unrelated architecture. Such outcomes had never been reported in the literature exploring extubation failure; thus, we were without a precedent solution and in unexplored territory in this domain.

Given the same performance was seen with two separate architectures, it was clear that the issue was with the data. At this stage, we recognised that only one study on extubation failure had to create temporal sequences and not just aggregate the data. As such, most other studies may not have reported similar problems as they did not have to deal with creating sequence data. It then became apparent that since the original MIMIC-IV data had several missing values, we had to manufacture a large quantity of synthetic data.

In general ML practices, synthetic data is significantly more straightforward to acquire than real-world data, particularly in scenarios with privacy issues [132]. Consequently, it is widely used in ML research in areas from Computer Vision [133], Natural Language Processing [134] and more poignantly, healthcare [135, 136], where patient confidentiality and privacy make acquiring data all the more challenging.

However, data synthesis comes with its risks. Artificial intelligence only learns patterns from the given data and cannot distinguish between real-world and synthetic data. Data synthesis often has implicit issues that impinge on the representativeness of data distribution [132]. Generating a large amount of synthetic data creates patterns that may not represent the underlying features, mainly where the original data is sparse. Consequently, as quoted in the review by Hao *et al.* [132], "the model may internalise erroneous patterns, thereby inducing biased predictions and undermining the overall performance and reliability of the model when confronted with genuine data". Moreover, synthetic data may not capture temporal patterns, which "may culminate in the ineffectuality of models in real-world applications" [132]. As such, it became clear that this could have caused the biased behaviour displayed by both the LSTM and TCN models – so we turned our attention to our data synthesis approach.

We developed an entirely unprecedented approach to proactively minimise the negative impact of synthetic data. By strategically grouping features into subsets based on observed sampling frequencies, we addressed the core issue of inconsistent synthetic data proportions across features. This ensured features were resampled at a rate commensurate with the amount of original data present. The resampling rates were guided by clinical insight, providing meaningful data handling and minimised artificial patterns. Furthermore, we employed masking to maintain data integrity by avoiding imputation and forcing the models to handle missingness.

This methodology represents a significant step forward in medical data pre-processing. We have not only established an initial framework for minimising the exacerbating effects of synthetic data but also maintained data integrity and provided sufficient information for model training. The true power of this innovative approach became apparent when we trained and evaluated our models on this newly processed data.

Hereinafter, our model architectural choices had to be guided by the nature of the data. Initially, we had chosen LSTMs/TCNs and engineered the data to meet the basic need for a consistent sampling frequency. However, since we had placed data into subsets, each with different dimensions, the architecture had to be adapted. This led to the creation of our "Fused LSTM" and "Fused TCN" models.

To the best of our knowledge, such a fused decision system has not been used in machine learning studies in the extubation domain, let alone extubation failure. Implementing these architectures and a simple FFNN incorporating static data yielded a remarkable transformation in model performance.

Upon evaluation, models could discriminate between extubation failure and success, mitigating previously demonstrated total bias. The Fused LSTM and TCN models no longer had perfect recall or specificity (indicators of complete bias), making it clear that splitting the data into feature subsets and using our bespoke resampling strategy had made a difference. Our findings validated the notion that the way synthetic data is created and handled was almost undoubtedly the determining factor contributing to model bias and overfitting. We implemented our strategy across feature sets and temporal model architectures, and the results enabled us to make observations across architectures, including static data and adding features.

Misclassification in this domain is unacceptable. This is mainly reflected in the high cost of false positives and negatives. On the one hand, a false positive indicates a patient expected to fail extubation but was ultimately going to succeed. If this were recommended to a clinician, the patient would undergo a prolonged period of ventilation, which increases the risk of ventilator-associated complications [23]. On the other hand, a false negative indicates failure to correctly identify a patient who will fail extubation. This could lead to severe complications as the patient has not had sufficient time to recover from their underlying condition and thus does not have the muscle strength to pick up the respiratory work, potentially with fatal consequences. As such, a balance between high precision (minimising false positives) and recall (minimising false negatives) is a clinically desirable requirement of any model deployed in the ICU. Consequently, we prioritised looking at the F1 score alongside the AUC-ROC to ensure our models were not only accurate but also potential candidates for clinical use.

An essential piece of analysis is evaluating the performance of temporal models in comparison to the baseline to ensure our temporal architectures are sufficiently informative. While the AUC-ROC was generally higher for LSTMs and TCNs, the LightGBM models typically had a higher or comparable F1 scores across the feature sets. One could interpret that this somewhat reinforces the importance of modelling temporal dynamics, with temporal architectures providing a more nuanced prediction capability. However, this is contingent upon the features used and model choice, and thus, the outperformance is not unequivocal at this stage.

Recall was seen often to bring down the F1 score in the temporal models. This would suggest that the model is failing to predict a large proportion of failure instances in the test data. A possible explanation for the higher AUC-ROC but low F1 score relates to the imbalanced nature of our dataset. Our temporal models seem to predict the majority class well but perform poorly on the minority class (failure), resulting in a lower recall and, thus, F1 score. Relating this to the performance versus the baseline, temporal models may struggle with class imbalance if they place greater importance on temporal patterns indicative of the majority class, which is seemingly the case here. Moreover, if the minority class samples do not exhibit consistent temporal patterns, temporal models likely will not capture them.

LightGBM, on the other hand, is a non-temporal model that implements a gradient-boosting framework. In the process of boosting, each model iteration focuses on correcting the mistakes of the predecessor by giving higher priority to misclassified instances [53]. As a result, LightGBM is likely better at handling imbalanced datasets, as outlined by Khan *et al.*, "LightGBM's ability to handle imbalanced datasets makes it an attractive option for a wide range of applications" [137]. This could explain the better precision-recall balance and resultant higher or comparable F1 score.

Furthermore, the decision-making process of LightGBM classifiers is significantly more straightforward than that of temporal models. While temporal models are conscious of past values influencing future predictions, LightGBM uses ensembles of decision trees to represent complex relationships in the data [53]. This may align more with our aggregated feature sets than the more complex subsets with bespoke sampling rates. The innate simplicity may result in improved generalisation, a better balance between precision and recall and, thus, at times, a higher F1 score.

In this study, the LSTM and TCN models performed relatively similarly. The LSTM models have marginally higher AUC-ROC, but the F1 scores are generally comparable, all under 0.5, suggesting a poor balance between precision and recall and room for improvement in their predictive competencies.

This observation is relatively surprising, given the findings reported by Bai *et al.*, that TCNs outperformed LSTMs across several tasks [115]. This conclusion is re-iterated by other authors, including Gopali *et al.*, who reported TCNs outperforming LSTMs when applied to anomaly detection [138]. However, when we focus on the extubation domain, the work by Catling *et al.* directly compared the performance of an LSTM-FFNN to a TCN-FFNN to predict extubation [77]. While this is not specifically extubation failure, their results show that the TCN model performed similarly to the LSTM, reflecting the behaviour in this study.

We also investigated the impact of adding static data context on model performance. Clinically speaking, static features provide essential context to decisions made in the ICU. As Dr Murali corroborated, factors such as age and BMI can indicate how likely a patient is to respond to ventilation and subsequent extubation successfully.

The results seem inconclusive regarding the impact of static data. For the LSTM models, there is very little difference in AUC-ROC between the models with or without static data, a trend also seen in the F1 score. However, for the TCN models, an increase is seen in both AUC-ROC and F1 scores across all feature sets. The marked rise in F1 score is driven by an increase in recall, suggesting that the context provided by static data may enable the model to capture more failure cases. Now, the reason why the impact of static data on TCN models was more notable is not apparent and most certainly warrants further investigation.

However, the ability to adjudge the informative contribution of static data to temporal data when predicting extubation failure has not been achieved before. Moreover, integrating an FFNN to learn patterns from the static data inspired by Catling *et al.* [77] has also not been applied in this domain. Typically, static data is passed directly into the LSTM/TCN layers by copying values across each time point (as done by Zeng *et al.* [80]). While computationally convenient, it overlooks the elementary distinction between static (tabular) and time-series data. Our design explicitly addresses this while leaving significant avenues for developing more sophisticated hybrid models. Moreover, our design more naturally reflects clinical decision-making, where insights from temporal and static data are recorded separately but combined to make an informative decision.

Our experimental design also uniquely enabled us to see the impact of incrementally adding features of varying clinical relevance. Adding features to the LSTM and TCN models had minimal impact on AUC-ROC. When we look at the F1 score, there seems to be a more notable improvement to the LSTM when more features are added than to the TCN models. Seemingly, neither LSTM nor TCN models have linear relationships between the number of features and their performance. Although additional features may improve specific metrics, such as recall, there is a trade-off of lower precision and potential overfitting, suggesting a complex interplay between data characteristics and model architecture.

The findings here may seem surprising when considering the clinical context of each feature set. Feature Set 1 represents the features at the cross-section between popularity in literature and clinical availability. Therefore, one would expect its respective models to have the best discriminatory power. However, our findings show that it either required more features for improved performance with the LSTM or its clinical relevance was not completely captured in the TCN. Interestingly, similar results across feature sets suggest that adding features did not encourage overfitting any more than could be underlying, possibly highlighting the likely limitations of relying on literature-based feature selection, particularly on studies lacking clinical input and not based on MIMIC data.

Having additional features does not always translate to improved performance when predicting extubation failure. As demonstrated by Zhao *et al.*, their lighter model based on 19 critical features performed to a similar level as their extensive 89-feature model, aligning with our findings in this study [49]. When considering the number of features to train a model, there is a balance between model performance and interpretability, particularly in a clinical setting. A parsimonious model with a controlled number of clinically relevant features might be preferable from a computational intensity and practicality standpoint.

However, assessing the impact of the number of features is a machine learning convention and not necessarily clinically driven. Clinicians typically focus on a select number of indicative features, but not all. Hence, the number of features is not really a consideration. We show that ML has the potential to process numerous complex features and learn patterns from those clinicians may not have considered.

Our work presents a transparent and clinically informed feature selection methodology, providing a structured platform for further work to explore more refined strategies. Categorising features into sets of varying clinical relevance has not been done before in this domain, and we believe it has the potential to provide unique insights into the usefulness of features that clinicians may not typically examine.

However, while analysing the minutiae of differences between metrics to gain insight into model behaviour is an important exercise, we must also take a step back. When looking at performances across LSTM, TCN, and LightGBM, it is intriguing that all models seem to perform at a similar underwhelming level. In other words, there does not seem to be a prominent "best-performing" model, demonstrated by the metrics in Table 5.1 and the similarity in the ROC curves (Figure 5.1 and Appendix Figures D.1 and D.2).

Regardless of model architecture, the inclusion of static data and an increasing number of features, no notable improvement or change in performance is seen. Values for specific metrics are seemingly within a narrow range of each other, and a trade-off is often seen between precision and recall. Furthermore, the AUC-ROC always seems to be around 0.6. While this is encouraging as it is slightly better than random chance (0.5), it is significantly lower than the acceptable levels for clinical consideration, which would be closer to 1. As such, one could class all models developed as being "weak learners" [139].

Comparing model performance against literature is arguably futile and reflects the novel nature of this study. The only near-comparable research is from Zeng *et al.* [80]. However, this study differs in how the data has been extracted, pre-processed, and annotated, how features were selected, and the level of meaningful clinical input, to name a few. So, while they report better performance on an AUC-ROC basis (0.828) for their LSTM model when predicting extubation failure, the foundation for comparison is weak. The observation is no criticism of that study but more a reflection of the disappointing lack of focus that has been placed on employing temporal architecture for this purpose in ML research more generally.

Ensemble methods were shown to help improve specific metrics but did not consistently outperform individual LSTM and TCN models. Suppose the ensemble's performance is comparable to that of the individual weak learners, as is the case here. In that case, it indicates that the nature of the data is throttling the performance of all models.

This analogous performance behaviour echoes our initial observations of the models before employing our innovative feature sub-setting and resampling approach. At that stage, models were consistently biased to one class or the other, and it was only when we refined our strategy regarding synthetic data that an improvement was seen. In this case, all models seem to be biased towards the extubation success class by a similar amount, but complete bias has now been mitigated. This general observation underscores the critical message that the data, its nature and the way it is pre-processed with regard to synthetic data are most likely the determining factors when predicting extubation failure.

The persistent model bias is fascinating. Even when we tried to minimise the impact of synthetic data through masking and bespoke resampling, the model performed better than random choice but was wholly inadequate for the clinical consequences of the task at hand. The primary reason is likely data quality. As outlined, the underlying MIMIC data was inconsistent across features and time. This, therefore, required significant pre-processing to achieve the necessary input for LSTM/TCN layers. Despite our efforts, ill-representative synthetic data patterns may still dominate the training data. This would undoubtedly be the case where a patient only had one data point for a feature with a high average observation frequency. Our masking only created NaN values where a patient had no original data for a specific feature. However, where they had at least one original data point, data for that feature would be resampled and interpolated, which would innately involve creating a considerable amount of synthetic data. Admittedly, a more refined approach building upon this novel methodology is required, with possible exploration into varying resampling logic to see what is optimal for the data at hand.

Moreover, we determined a fixed feature subset strategy based on Feature Set 1. As such, pre-processing was not adapted to the additional features in Sets 2 and 3 (save for the lowered threshold in Feature Set 3). It may, therefore, not be surprising that the performance between all feature subsets is similar given they were all pre-processed in a near identical manner.

However, it should be highlighted that our more considered efforts represented an improvement from our initial pre-processing attempt. Thus, our iterative discoveries still exhibit a problem with model bias but also present a potential solution toward clinically viable predictive models - within the influence of synthetic data.

## 6.3 Model interpretation

Interpretability is crucial to clinical support systems. Any prospective assistive tool in the field must earn the physician's trust through transparent and comprehensive explanations of predictions.

In this study, we used feature ablation to determine the impact of dynamic and static features on our model predictions, but this was not the only technique considered. As outlined in great detail in the review by Di Martino *et al.* [140], several methods exist to encourage the transparency of predictions when employing AI with healthcare time-series data. These techniques can be split into post-hoc and ante-hoc interpretability methods. Post-hoc refers to interpretations made of the model after training, while ante-hoc explanations offer transparency during training.

The primary examples of post-hoc methods commonly used across machine learning are SHAP and LIME. SHAP offers global (across the whole patient set) and local (for one patient) feature importance, and LIME focuses on local explanations only [59, 141]. Both, however, would be useful for clinicians. Understanding how the model interprets features across all patients would enable the clinician to cross-check that predictions are based on medically relevant features. Moreover, at the individual level, each patient has different medical dispositions, and thus, a local understanding of which features seem to be contributing to a prediction of extubation failure could assist in a more tailored treatment plan.

SHAP and LIME are the state-of-the-art methods for explainable AI, and SHAP has seemingly been successfully implemented in many extubation failure studies, including Zeng *et al.* [80]. As such, we attempted to explore these first using the available Python packages. However, it quickly became apparent that these techniques would prove technically challenging to implement due to our models' and data's novel and more complex nature. SHAP values are based on feature importance in a single input space; our architecture leverages multiple inputs representing each feature subset, and thus, SHAP would struggle to calculate a unified value. Furthermore, SHAP does not naturally account for temporal dependencies. Furthermore, the implementation of masking to maintain data integrity complicates matters. SHAP struggled to understand these masks' impact and how missing values should be implicated in the final output.

To see whether this pattern only applied to the SHAP and LIME libraries, we attempted to employ a couple of techniques mentioned in the paper by Martino *et al.* [140] and readily available through the Python Captum library: DeepLIFT and Integrated Gradients (145). Unfortunately, similar challenges were incurred here as well.

The factors that set our work apart – feature subsets, bespoke resampling and masking – rendered traditional interpretability methods inadequate. As such, we decided to pivot and use feature ablation, which can be performed without a package, unlike most post-hoc strategies. While ablation may not offer the complexity of interpretation or popularity of the use of SHAP or LIME, it allowed us to assess the overall impact of features on model predictions.

Our models seem to pick up clinically relevant features but lack alignment in their importance. In the LSTM model trained on Feature Set 1 without static data, Tidal Volume (observed) and Minute Volume were towards the upper end of feature importance. This aligns with clinical understanding, as a low observed tidal volume might indicate respiratory fatigue, and a low minute volume is indicative of insufficient gas exchange. Tidal volume represents the air inhaled and exhaled with each breath, with the minute volume representing the total air breathed in one minute [142]. A low observed tidal volume might indicate respiratory fatigue and a low minute volume indicates insufficient gas exchange, which could lead to the patient being considered a failure upon extubation as they cannot pick up the breathing. Interestingly, minute volume is calculated by dividing tidal volume by respiratory rate. As we can see, the respiratory rate was the least helpful feature; thus, the importance of the minute volume is slightly confounding. It may suggest that the positive predictive impact of tidal volume can compensate for the limitations of respiratory rate, but this is challenging to conclude without further work.

Incorporating static data was intended to enrich the models' understanding. Briefly observing the static features for the LSTM trained on Feature Set 1, age $\geq$75 and BMI being the most indicative features, resonates from a clinical standpoint. Geriatric (older) patients have less muscle mass and reduced lung elasticity and are, therefore, less likely to pick up the work when extubated from the ventilator. Moreover, a high or low BMI is a crucial indicator of patient outcomes, as corroborated by Dr Murali. Patients with a high BMI will have significantly more breathing work due to excess weight on their chest wall and diaphragm. Moreover, those with low BMI often present muscle weakness, impairing their ability to sustain adequate breathing effort. As such, it is interesting that our model seemed to pick up on these relevant relationships. Age $\geq$75 was consistently the most important static feature across LSTMs in each feature set. Tidal Volume (observed) and Minute Volume were also deemed relatively informative across LSTMs trained on only dynamic data on the first two feature sets. This reflects a potential level of consistency in LSTM model prediction regardless of the number of features.

However, the order of these importances was not reflected in the TCN model on the same data. Arterial $CO_2$ pressure ($PaCO_2$) was the most important feature in the model without static data for Feature Set 1. Elevated $PaCO_2$ reflects inadequate alveolar ventilation, indicating the patient is struggling to breathe efficiently, possibly resulting from respiratory muscle fatigue and increasing the risk of extubation failure. In the static data, weight was the most influential feature in the first feature set and has the same logic as BMI.

Another intriguing observation across feature sets and architecture is the varying importance of ethnicities. As explained by Dr Murali, ethnicities are not typically looked at when deciding to extubate. Still, the varying importance between ethnic groups is a notable finding, particularly in light of the insights gained from the COVID-19 pandemic. Ethnic minority groups were disproportionately affected by the virus, with Black and Asian patients over-represented in those requiring advanced respiratory support in the UK as a result of socio-economic and medical dispositions [143]. Naturally, a vast amount of research was conducted around the pandemic to discover such trends. While the evidence in this study is inconclusive regarding which ethnicities are most indicative of extubation failure, it warrants further investigation.

However, it should be noted that feature importance across architecture and feature sets is, for the most part, inconsistent. Adding static context also seems to influence the perceived importance a model places on the dynamic features compared to the same features in the dynamic-only model. This inherently raises questions about the robustness of these interpretations.

While ablation offered valuable glimpses into model behaviour, once again, we must take a step back. Looking at the change in AUC-ROC values more closely, one can see that all values across all feature sets and model architectures are minimal. All models display feature importances close to zero, with the largest no higher than $\pm 0.05$.

While feature ablation offers key insights into model behaviour, the minimal changes in importance suggest that model performance is not heavily dependent on any single feature, regardless of the architecture. Interestingly, this sentiment was reflected in discussions with clinicians who highlighted that all the features considered across the feature sets are clinically relevant. Several of them are linked; therefore, they present a holistic representation of a patient used to make a decision, using a range of features to make predictions. Our models seem to capture this clinical reality.

Besides the clinical reasons, several technical reasons could explain this behaviour. Feature ablation, while simple to implement, is inherently flawed as it does not capture the interactions between features in the way SHAP can, only assessing the impact of a single feature in isolation. Hence, it may be that all features are highly correlated (as clinically expected) and, therefore, share predictive power, making it harder to isolate. To confirm this, correlation analysis should be carried out on features before training to gauge potential relationships and remove highly correlated features, as implemented by Chen *et al.* [50]. Furthermore, our model architecture may be too complex, which could result in overfitting to noise or spurious correlations rather than interpreting a true signal.

More fundamentally, the models may struggle to leverage features to make robust predictions. Since this occurred across two unrelated architectures, the explanation likely again lies within the nature of the data. Where synthetic data is rife, the impact of truly important features may be diluted if their temporal patterns are artificially obscured. This would result in small changes in AUC-ROC and an underestimation of feature importance. Particularly where the amount of synthetic data varies between features and patients, ablation results would also vary between models and experiments, potentially explaining the inconsistencies. Through interpretation of model predictions, we have again shown that synthetic data is likely the determining factor in model behaviour when predicting extubation failure.

Consequently, a more granular analysis of feature importance will likely yield diminishing returns in terms of actionable insights. Instead, we focus on the broader findings gained and, more importantly, the capability of the models developed to offer a level of transparency, albeit limited at this stage. The ability to order features by importance across patients is beyond the day-to-day understanding of clinicians. The complexity of interpretation reflects the inherent complexity of clinical decision-making, where a multitude of interconnected factors contribute to a final assessment. Features employed across different sets are clinically relevant and have been investigated in various studies across the literature. However, while essential, the novelty of our approach makes interpretability challenging but undoubtedly achievable with further work. The consistent use of SHAP in the literature possibly reflects ML convention rather than an adapted strategy to the complexity of the data and, consequently, the model at hand.

## 6.4  Clinical relevance of our proposed prediction system

In this study, we have outlined the development of an end-to-end prediction system to evaluate the likelihood of a patient failing extubation.

Our system comprises three distinctive components: data pre-processing, model development, and interpretation. Within each of these, our work has significantly contributed to advancing the field through innovative, data-driven, and clinically relevant approaches, as outlined below.

- **Data Pre-processing**

  - **Clinically Informed Feature Selection:** We introduced a systematic approach for feature selection, balancing popularity in precedent studies and real-world clinical considerations with the WAVE study used as a benchmark

  - **Clinically relevant Time Window:** We prioritised a 6-hour window before extubation for data extraction to reflect clinical practice in extubation decision-making. This confirmed clinically relevant time frame is surprisingly not used in previous work

  - **Stratified Train/Test Split:** We stratified the train/test split based on the proportion of synthetic data required. This ensured the model was not biased toward ill-representative synthetic patterns, an approach not seen in previous studies

  - **Novel Resampling and Interpolation:** We created a bespoke resampling strategy, splitting features into subsets based on the average observed frequency to try and minimise the creation of synthetic data patterns, which would otherwise lead to model bias. Linear interpolation was used to better represent temporal patterns where forward fill was traditionally used

  - **Masking:** Where features had no values, we filled them with NaNs and implemented masking to preserve data integrity. This allowed models to handle missing data more effectively and minimised synthetic data creation

- **Model Development**

  - **Exploration of Temporal Architectures:** Our work is the first to explore using a Temporal Convolutional Network (TCN) to predict extubation failure. We compared the performance of two inherently different architectures to assess their strengths and weaknesses

  - **Fused Decision Architecture:** Guided by the data requirements, we developed Fused LSTM and TCN models to integrate multiple feature subsets resampled to different rates. This novel approach enabled us to handle various inputs

  - **Static data integration with FFNN:** Although not the first to implement this for static data, a FFNN has not been explicitly used when predicting extubation failure. This enabled us to handle the fundamental difference between tabular and time-series data while ensuring static context was captured

  - **Bespoke Hyperparameters to Handle Class Imbalance:** To address the inherent imbalance in the data, we implemented a bespoke sampling method and weighted loss parameters to be tuned during optimisation. This moved beyond the traditional application of one technique or the other, providing a more dynamic data-driven solution

- **Model Interpretability**

  - **Challenges with traditional interpretability methods:** Our work highlighted the limitations of popular interpretability techniques (SHAP and LIME) when applied to complex, temporally dependent and synthetic-data contingent models. This is crucial foundation for future work

- **Feature Ablation with Clinical Insights:** We pivoted to implement feature ablation as a practical alternative, which has not been used before in previous work. This showcased an avenue towards model transparency that can be justified or questioned through clinical insights

More fundamentally, our work highlights the requirement for the integration of clinical input at all stages of data pre-processing, model development and interpretability. Our study prioritised the end-user, ensuring all choices made were clinically relevant and justifiable as far as possible.

Clinical deployment is challenging in the healthcare space. While the application of ML has automated and considerably improved certain areas within medicine, the acute requirements and considerable consequences of predictions extend beyond the typical deployment pipeline [144]. As such, the development of assistive tools must be rigorous and comprehensive. Our current models, while demonstrating a degree of transparency and improvement over random chance, are not yet sufficiently accurate or interpretable enough to warrant clinical deployment.

However, this study should not be viewed through the narrow lens of immediate clinical implementation. Instead, it serves as a robust platform for future work, with insights that shed light onto several critical areas that, without question, must be addressed in any study looking to predict extubation failure.

Most importantly, we have outlined the pivotal role of synthetic data and its potential to introduce biases but also presented potential solutions that need to be further refined. We have also underscored the importance of a data-driven approach, emphasising that model architecture must be guided by the nature of the data rather than pre-conceived machine learning conventions, which may not be clinically relevant.

## 6.5   Limitations and Future Work

Our study has several limitations. We focus on the fundamental limitations, primarily on the novel aspects of our end-to-end approach, while also presenting potential avenues for future work to address these. The limitations and future work pertaining to pipeline components and the general study are outlined in Tables 6.1, 6.2, 6.3 and 6.4.

## 6.6   Legal, Social, Ethical and Professional Considerations

Ethical considerations for this project were reviewed in Appendix Figure E.1. The current study was done retrospectively, and secondary data was used; hence, there were no relevant ethical concerns. No human participants were involved, and all patient data in the MIMIC database was wholly deidentified. This is also the case for any data provided regarding the WAVE study. Despite the data being sensitive electronic health records, no identifying characteristics were provided. Informed consent was received from the relevant parties to use both datasets. Furthermore, both studies received the respective ethical review board approval; hence, Imperial College London did not require one for this project.

We accept that there can always be abuse of ML when models are deployed in a clinical setting. Our research aimed to improve patient outcomes and inform decisions around extubation failure.

In this paper, we acknowledge several risks of misclassification, potential causes and offer some avenues of mitigation. Patient populations are naturally diverse; thus, the choice of training data, feature selection and model development are vital when ensuring equitable outcomes.

From a social perspective, the perception of machine learning depends on the level of understanding. The inherent complexity of the temporal models used in this study (LSTMs and TCNs) further complicates the level of transparency. To address these perceptions, we prioritised transparency and explainability throughout, which we argue must be maintained in any ML research in medicine. Wider acceptance of ML could be improved by considering data privacy issues, model bias, and human oversight.

All machine learning experts predicting extubation failure have a duty to collaborate closely with clinicians to ensure clinical relevance throughout the investigation and, eventually, to train them to use the developed assistive tool. It is crucial that the final decision must remain with healthcare professionals. Entrusting a patient's life to an ML model is ethically incompetent, and it should be ensured that any models deployed are merely a supportive tool rather than a replacement.

As we have argued, AI has the potential to revolutionise clinical practice. However, realising this potential necessitates a responsible and ethical approach that prioritizes patient safety, data privacy, and equitable predictions while ensuring that medical values are upheld.

Table 6.1: Data and Data Pre-processing limitations and future work

| Data and Data Pre-processing | | |
| --- | --- | --- |
| **Topic** | **Limitation** | **Future Work** |
| **Reliance on a single dataset** | This study was solely based on the MIMIC-IV database. The dataset was imbalanced despite reflecting real-world distributions of extubation failure. Reliance on a single dataset also means our findings are not generalisable to other ICU settings. Moreover, we only used the WAVE dataset as the sole benchmark of real-world features. While the WAVE features are clinically meaningful, one single study cannot inherently represent all clinically available features. Classifying features as clinically available or unavailable likely distorts the complexity of data collection. As such, some features may be misplaced within the feature sets. | Future work must utilise large data sources, whether primary or secondary, where possible. Aggregating datasets together is challenging as the context and data recorded may not align (as with MIMIC and WAVE). Studies should explore the use of publicly available databases with a substantial number of patients other than MIMIC-IV, where applicable, such as eICU [145]. Features from multiple datasets enriched with clinician intensivists should be cross-referenced during feature selection to better capture the diversity of real-world ICUs. |
| **Poor quality/missing data** | MIMIC, as with medical data, more generally, had a large amount of missing data, and thus, there is likely a ceiling up to which model performance can be optimised through pre-processing. Data points were not sampled at consistent frequencies, forcing us to develop a novel resampling and interpolation approach. Our models would likely perform better on cleaner, more consistent data. | Future studies would benefit from exploring diverse ICU databases to acquire higher-quality data. An alternative would be to carry out primary research where data recording would be more within the authors' control and can be tailored to their objectives. <br><br> Improved data imputation techniques could be employed to handle missing data, with one option leveraging generative models for data augmentation. However, care would need to be taken to ensure this does not propagate the instance of synthetic data patterns. |
| **Pre-processing was all based on Feature Set 1** | Preprocessing was not adapted to the unique characteristics of each feature set. The threshold and sub-setting strategies were solely based on Feature Set 1, potentially throttling performance on Feature Sets 2 and 3 and explaining the similarity in performance across the feature sets. | Pre-processing strategies need to be unique to each feature set. The thresholds, number of subsets, and resampling rates all need to be tailored to the unique features of the set and its performance and then validated across multiple scenarios. In this study, the threshold values and resampling rates were set with clinical guidance, but given this has not been done before, a wide range of values should be investigated. Ideally, the strategy should be made to dynamically adjust the criteria to the observed data. ML models could even be deployed to learn optimal subset sizes, numbers, and threshold values for a given dataset. Then, they can be adaptively applied when new data is introduced. |
| **Setting missing values to NaN** | Although used to maintain data integrity when masking, setting values to NaN in lower-represented features led to limited data for those features, potentially affecting model performance and underrepresenting certain features. | More comprehensive data must be acquired to minimise missing data and lessen reliance on NaN masking. Admittedly, this is difficult to achieve in the medical domain. More advanced imputation methods could also be developed to preserve feature relevance. |
| **Non-patient-specific imputation** | At times, average values were used for imputation, which would not represent individual patient nuances and affect the discriminatory ability of the models. | Studies should investigate more patient-specific imputation techniques that rely possibly on historical data for that patient prior to the 6-hour window used in this study. |
| **Subjectivity in static feature selection** | Static features were mainly selected with clinical guidance. While relevant, our study may not capture the variety of static features available to ICU intensivists and introduces a degree of subjectivity to the study. | Consensus studies should be implemented involving several clinicians to get a better idea of common static features and reduce subjectivity. <br><br> A more quantitative approach, such as observing static feature usage patterns across electronic health records, can be used to complement clinical opinions. |

Table 6.2: Model development limitations and future work

| Model development | | |
|---|---|---|
| **Topic** | **Limitation** | **Future work** |
| **Only two temporal architectures were used** | While implementing a TCN model to predict extubation failure is novel, it might be that the issues we were experiencing regarding model bias are isolated to LSTMs and TCNs. Better-suited architectures may exist for this data. | A wide range of temporal architectures should be explored, including Temporal Fusion Transformers and even more hybrid approaches that combine different ensembles of temporal architectures for potentially boosted predictive performance. |
| **Models developed can only handle three feature subsets** | Our models were explicitly designed to handle our pre-processed data by having three LSTM/TCN layers, one for each subset. This architecture is fixed and was not designed to be adaptable to potentially more or fewer subsets. This would limit its transferability to other datasets. | Model architecture should adapt to the number of feature subsets, likely passing the number of subsets as a parameter to guide the required number of LSTM/TCN layers. |
| **Adding a FFNN is computationally expensive** | Using a hybrid LSTM/TCN-FFNN model to handle static and dynamic data is novel. However, incorporating an extra neural network on top of a deep temporal network significantly added to the computational resources required. This expense may not be applicable in all scenarios. | More computationally efficient alternatives should be investigated to handle static data, possibly integrating Gradient Boosting Machines such as XGBoost or CatBoost, designed to handle mixed data types. These are also substantially easier to interpret and could aid in model explainability. |
| **Choosing to employ global hyperparameters for the LSTM/TCN layers** | Using global hyperparameters to represent each LSTM/TCN layer may have led to suboptimal performance. | Studies should explore local hyperparameter tuning for different model layers. At the same time, this may increase the computational resource requirements for cross-validation; thus, efficient methods like Bayesian Optimisation must continue to be employed. |
| **Model training time was not considered** | In this study, we did not consider the time required to train our models. This is essential for any ML study with limited computational resources and time. Generally, training deep learning architectures such as LSTMs and TCNs takes extensive time due to the significant number of parameters. | Measures should be built into the model code to record the time taken to train. The trade-off between model complexity, performance and training time should be analyzed to identify the most economical models for real-world deployment. |
| **Only LightGBM was used as a baseline** | In literature, typically, several models were employed as non-temporal baselines. For example, in Zeng *et al.* [80], logistic regression, support vector machine, MLP, random forest and XGBoost were used as baseline models. | Additional baseline models should be introduced, replicating the variety employed in the literature for a more comprehensive comparison. |
| **Integration of FFNN and LSTM/TCN outputs** | FFNN and LSTM/TCN outputs were only concatenated in the final step in our forward propagation. This may not represent the most optimal means of introducing static context to temporal data. | Various integration approaches should be investigated, and model performance should be compared to find the optimal strategy. One alternative is combining the FFNN output at each LSTM/TCN sequence time step. This would facilitate dynamic interaction of the static context with the time-series data throughout the prediction process. |

Table 6.3: Interpretability limitations and future work

| Interpretability | | |
| --- | --- | --- |
| **Topic** | **Limitation** | **Future work** |
| **We were only able to implement feature ablation for model interpretability** | Due to the inherent complexity of our approach, traditional post-hoc interpretability packages were rendered incompatible. As a result, we could only visualize the importance of features through ablation, which has inherent limitations. Clinicians cannot trust a model they cannot interpret; hence, the work completed in this study is not viable for clinical deployment at this stage. | Several other interpretability techniques exist that could be compatible and prove more successful. Di Martino *et al.* [140] outlined that attention mechanisms are the primary ante-hoc method for RNN-related models. Attention assigns different weights to different input elements, and these can be visualized to interpret which parts of the input sequence the model focuses on when making a prediction. Clinicians can then see which features and time points most influenced the model's prediction. Moreover, temporal attention can be implemented to see how the importance of certain features changes over time. This can be vital information for a clinician when assessing patient trajectories, so further work should focus on this. For TCN interpretability, gradient-based methods are popular [140]. Although we tried using Integrated Gradients, alternative techniques such as Grad-CAM should be investigated to see whether more refined insights can be extracted. |

Table 6.4: General study limitations and future Work

| General study | | |
|---|---|---|
| **Topic** | **Limitation** | **Future work** |
| **Lack of external validation** | A primary limitation of this study was the lack of external validation. Robust external validation in any study looking into extubation failure is critical to establishing model generalisability and clinical efficacy. However, this was not possible given the time and resource constraints of this thesis. | Inspiration should be taken from the study by Zhao *et al.* [49], where external validation was carried out on over 500 patients from a different hospital. Similarly, future studies should identify an alternative dataset or real-world patients for external validation. This should be common practice in any study looking into extubation failure. |
| **Not all decisions were clinically validated** | Due to limited time with clinicians, we had to prioritise specific areas requiring domain knowledge. As such, certain decisions, such as outlier removal, were not clinically informed, which could skew the data towards anomalous results and confound model predictions. | Where possible, future studies must involve clinicians in all aspects of data pre-processing, model development, and interpretability while being considerate of their demanding schedules. |
| **Limited computational resources** | The entirety of this project was carried out using Google Colab. Since this is a third-party source, the computational resources are restricted in the free version, meaning we had to optimise computational training time where possible. This may have prevented us from exploring a broader range of hyperparameters when mitigating model overfitting. | Future studies should leverage high-performance computing or cloud-based platforms where possible. This would facilitate broader hyperparameter tuning and exploration of more computationally demanding model architectures that could be more apt for predicting extubation failure. |

# Conclusion

In this thesis, we have established a strong foundation for the creation of an end-to-end system for predicting extubation failure to assist clinicians. We pioneered innovative, clinically relevant strategies to pre-process medical data to minimise the influence of synthetic patterns and pave the way for more robust and reliable models. To the best of our knowledge, we introduced the first use of Temporal Convolutional Networks (TCNs) to predict extubation failure, not to mention the application of fused decision architectures guided by the nature of the data. We also attempted to make our models interpretable to increase transparency and earn clinicians' trust.

Our approach, while not without limitation, represents a significant step forward in handling the challenges of real-world clinical data for temporal modelling. We unearthed and addressed significant areas of consideration often overlooked in the literature, particularly the inescapable issue of synthetic data bias. Our work underscores the importance of adapting data pre-processing to the specific characteristics of the data and clinical context rather than relying on established machine learning conventions.

Deciding to extubate a patient from the ventilator can be a life-or-death decision. With the profound cost of an ill-informed decision, the need for reliable and interpretable assistive tools cannot be understated. Where the majority of previous work fails to recognise the gravity of the situation, our approach thrives by focusing on clinically informed data pre-processing, employing relevant architectures to capture temporal patient trajectories and model transparency. We intend for this study to be viewed as a robust platform to inspire future work to predict extubation failure, making those reading cognisant of the potential challenges and proposing means of addressing them, ultimately with the view of improving patient outcomes.

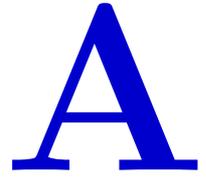

# Appendix: Literature Review

Table A.1: Overview of typical studies that use machine learning to predict extubation failure. Note, the definition of extubation failure is implied to be re-intubation within the highlighted period. Table adapted from Igarashi *et al.* [17]

| Author (year) | Kuo (2015) [44] | Hsieh (2018) [45] | Chen (2019) [50] | Fabregat (2021) [54] | Otaguro (2021) [42] | Zhao (2021) [49] | Fleuren (2021) [56] | Huang (2023) [55] |
|---|---|---|---|---|---|---|---|---|
| Most accurate model | ANN | ANN | Light GBM | SVM | Light GBM | CatBoost | XGBoost | RF |
| Dataset (Train and test) (N, patients) | Two hospitals (N = 121) | Single hospital (N = 3,602) | MIMIC-III (N = 3,636) | Single hospital (N = 1,108) | Single hospital (N = 117) | MIMIC-IV (N = 16,189) | COVID-19 database (N = 883) | Single hospital (N = 233) |
| Dataset (External validation) (N, patients) | No | No | No | No | No | Single hospital (N = 502) | No | No |
| Number of features | 8 | 37 | 68 (36 compact) | 20 | 58 | 89 (19 compact) | 40 | 6 |
| Definition and rate of extubation failure | <48 h, 26% | <72 h, 5% | <48 h, 17% | <48 h, 9% | <72 h, 11% | <48 h, 17% | <48 h, 13% | <48 h, 12% |
| n-fold cross-validation | 5 | 10 | 5 | 7 | 5 | N/A | 5 | 10 |
| Accuracy | 0.80 | N/A | 0.8023 | 0.946 | 0.9265 | N/A | N/A | 0.940 |
| Sensitivity / Recall | 0.82 | 0.822 | 0.7485 | N/A | 0.9602 | 0.72 | N/A | 0.875 |
| Precision | N/A | 0.939 | N/A | N/A | 0.9146 | N/A | N/A | N/A |
| Specificity | 0.73 | N/A | 0.8327 | N/A | N/A | 0.78 | N/A | 0.967 |
| F1 score | N/A | 0.867 | N/A | N/A | 0.9369 | 0.77 | N/A | 0.958 |
| AUROC | 0.83 | 0.850 | 0.8130 | 0.983 | 0.9502 | 0.835 (0.803 ext. validation) | 0.7 | 0.976 |
| Interpretability | No | No | SHAP | No | No | SHAP | No | SHAP |

# B

# Appendix: Data Pre-processing

Table B.1: Detailed rationale and implementation of inclusion criteria to obtain the initial patient cohort. All criteria were developed in collaboration with Dr Murali for clinical relevance

| Criteria | Rationale | Implementation |
| --- | --- | --- |
| **Needs to be in the ICU** | We will only consider patients undergoing mechanical ventilation in the ICU. While forms of ventilation are possible outside the ICU, IMV is typically administered in the ICU due to the level of care required and the risks involved. | Only data from the icu folder of MIMIC-IV was considered. |
| **Needs to have undergone invasive mechanical ventilation** | Positive-pressure ventilation can be invasive or non-invasive. In this study, we will only consider IMV given the focus on predicting extubation necessitates invasive ventilation. | Extracted patients from the *procedureevents* file that had an event recorded where the itemid was Invasive ventilation. |
| **Needs to have been extubated** | It is not guaranteed that all patients undergoing IMV have a recorded extubation event at the end, given the inconsistency sometimes present in healthcare data. As we are trying to predict extubation failure, it was essential that any patient must have a recorded extubation event such that we could perform data annotation (i.e. extubation failure/success) for supervised model training. | Once the cohort of patients had been derived after the first two criteria, we had to ensure that the ventilation events for these patients had an associated extubation attempt, successful or not. The *procedureevents* file logged extubation events with the corresponding itemid. All events in *procedureevents* have start and end times. To match a ventilation event with its corresponding extubation, we applied the following logic derived with Dr Murali: If the start time of extubation is within 10 mins after the end time of ventilation where they both relate to the same patient and the same hospital stay, this is taken to be ventilation with attempted extubation. This logic considers that extubation may not necessarily be immediately after the recorded end of ventilation as further preparations may be required for the patient to be extubated. |

Table B.2: Full rationale and implementation of exclusion criteria to obtain the final patient cohort. All criteria were developed in collaboration with Dr Murali for clinical relevance

| Criteria | Rationale | Implementation |
|---|---|---|
| **Age <18 and >89 – excluding paediatric and extremely elderly patients** | Paediatric patients have different physiological responses and treatment protocols compared to adults. Similarly, elderly patients may have multiple comorbidities (conditions) that could confound analysis trying to predict extubation failure in the general adult ICU population. Specific age bounds were recommended by Dr Murali. | Since age is deidentified, we took the provided anchor_age as the best estimate for the patient's actual age and filtered out any that met the criterion. |
| **Duration of ventilation <24 hours or >30 days** | Ventilation duration is a vital observation of patient condition and treatment progression. Extremes in duration can represent distinct clinical scenarios. Longer durations might indicate chronic conditions or complications that are not the focus of our study, while short durations might indicate procedural ventilation (i.e. after surgery) rather than true respiratory failure. Procedural ventilation was taken to be ventilation with a duration of less than 24 hours, and prolonged ventilation was taken to be longer than 30 days following clinical considerations. Focusing on patients in a typical duration range ensures a homogeneous study group and helps the model learn patterns relevant to the majority of ICU patients. | Ventilation duration was taken as the difference between the start and end time of the logged ventilation event. |
| **Excluding repeated admissions** | Repeated admissions often reflect a deteriorating clinical course. We only focused on the first ICU admission of a patient, as repeated admissions might introduce variability related to different stages of illness or treatment response, potentially confounding analysis. | The first ventilation event was taken as the logged IMV event for that patient with the earliest start time. All further ventilation events for that patient were excluded. |
| **Patients under palliative care and end-of-life patients** | Patients under palliative care or nearing end-of-life often have several comorbidities that can influence extubation outcomes, potentially confounding analysis. Including these patients would introduce bias into the model, as treatment plans may not be geared towards successful extubation. | Palliative or end-of-life care was not explicitly recorded, but we took any reference to palliative care or end-of-life care for a patient to be indicative. This referred to any patients with a diagnoses ICD code pertaining to "Encounter for palliative care". For complete exclusion, we also removed any patient who had an entry in the *chartevents* file relating to palliative care. |
| **Patients that died during ventilation** | These cases do not provide information about extubation outcomes and would skew the analysis. | We excluded patients where the recorded time of death was before the end time of ventilation. |
| **Patients who had major head or neck surgeries** | Head and neck surgeries can directly exacerbate respiratory function and the ability to breathe independently post-extubation. These surgeries can lead to complications such as airway oedema, increasing the risk of extubation failure. Including these patients would introduce confounding factors that would prevent the model from being able to predict on a general ICU population. | We excluded any patient who had a recorded procedure in the *procedures_icd* file that related to head, neck, larynx, trachea, or respiratory system surgeries, where the procedure was recorded to have taken place before the end time of ventilation. |

Table B.3: Cross-reference Set 1 static and dynamic features outlining those present in MIMIC-IV. Cross-reference Set 1 refers to features popular in the literature and clinically available (present in WAVE). Features highlighted in blue were unavailable in MIMIC and, thus, could not be used in further analysis. Key: Y = available in MIMIC-IV, Y-calculate = components available in MIMIC-IV to calculate feature, N = not available in MIMIC-IV

| Cross-reference Set 1 | | | |
|---|---|---|---|
| **Feature** | **Static/Dynamic** | **Available in MIMIC** | **MIMIC label** |
| **Demographic** | | | |
| Age | Static | Y | - |
| Gender | Static | Y | - |
| Body mass index | Static | Y - calculate | - |
| Weight | Static | Y | - |
| Height | Static | Y | - |
| Ethnicity | Static | Y | - |
| **Vital Signs** | | | |
| Respiratory rate | Dynamic | Y | Respiratory Rate |
| $SpO_2$ | Dynamic | Y | $O_2$ saturation pulseoxymetry |
| $SpO_2{:}FiO_2$ | Dynamic | Y - calculate | - |
| End-tidal carbon dioxide | Dynamic | Y | $EtCO_2$ |
| **Laboratory results** | | | |
| $PaCO_2$ | Dynamic | Y | Arterial $CO_2$ Pressure |
| Arterial pH | Dynamic | Y | PH (Arterial) |
| $PaO_2$ | Dynamic | Y | Arterial $O_2$ pressure |
| Haemoglobin count | Dynamic | Y | Hemoglobin |
| P/F ratio ($PaO_2$ to $FiO_2$) | | Y - calculate | - |
| **Ventilator features** | | | |
| $FiO_2$ | Dynamic | Y | Inspired $O_2$ Fraction |
| Tidal volume | Dynamic | Y | Tidal Volume (observed) |
| Peak inspiratory pressure | Dynamic | Y | Peak Insp. Pressure |
| Minute volume | Dynamic | Y | Minute Volume |
| RSBI | Dynamic | N | - |
| Plateau pressure | Dynamic | Y | Plateau Pressure |
| Ventilator mode | Dynamic | Y | Ventilator Mode |
| Maximum inspiratory pressure | Dynamic | Y | Negative Insp. Force |
| Spontaneous breathing trial success times | Dynamic | N | - |

Table B.4: Cross-reference Set 2 static and dynamic features outlining those present in MIMIC-IV and their label. Cross-reference Set 2 refers to features popular in the literature but not clinically available (not present in WAVE). Features highlighted in blue were unavailable in MIMIC and, thus, could not be used in further analysis. Key: Y = available in MIMIC-IV, Y-calculate = components available in MIMIC-IV to calculate feature, N = not available in MIMIC-IV

|  | Cross-reference Set 2 | | |
| --- | --- | --- | --- |
| Feature | Static/Dynamic | Available in MIMIC | MIMIC label |
| **Demographic** | | | |
| Past medical history | | N | - |
| Reasons for respiratory failure | | N | - |
| Weight loss | | N | - |
| Charlson Index | | Y - calculate | - |
| **Vital Signs** | | | |
| Heart rate | Dynamic | Y | Heart Rate |
| Glasgow coma scale | Dynamic | Y | GCS - Motor Response + GCS - Eye Opening |
| Mean arterial pressure | Dynamic | Y | Arterial Blood Pressure mean |
| Body temperature | Dynamic | Y | Temperature Fahrenheit |
| Richmond agitation-sedation scale | Dynamic | Y | Richmond-RAS Scale |
| Systolic pressure | Dynamic | Y | Arterial Blood Pressure systolic |
| Diastolic pressure | Dynamic | Y | Arterial Blood Pressure diastolic |
| **Laboratory results** | | | |
| White blood cell count | Dynamic | Y | WBC |
| Creatine | Dynamic | Y | Creatinine (serum) |
| Platelet | Dynamic | Y | Platelet Count |
| Na+ | Dynamic | Y | Sodium (serum) |
| K+ | Dynamic | Y | Potassium (serum) |
| Lactate | Dynamic | Y | Lactic Acid |
| Glucose | Dynamic | Y | Glucose (serum) |
| Total bilirubin | Dynamic | Y | Total Bilirubin |
| Hematocrit | Dynamic | Y | Hematocrit (serum) |
| Ca+ | Dynamic | Y | Ionized Calcium |
| **Ventilator features** | | | |
| Mean airway pressure | Dynamic | Y | Mean Airway Pressure |
| Hours since last controlled mode | Static | N | - |
| Maximum expiratory pressure | Dynamic | N | - |

Table B.5: Cross-reference Set 3 static and dynamic features outlining those present in MIMIC-IV and their label if so. Cross-reference Set 3 refers to features not popular in the literature but clinically available (present in WAVE). Features highlighted in blue were unavailable in MIMIC and, thus, could not be used in further analysis. Key: Y = available in MIMIC-IV, Y-calculate = components available in MIMIC-IV to calculate feature, N = not available in MIMIC-IV

| Cross-reference Set 3 | | | |
|---|---|---|---|
| Feature | Static/Dynamic | Available in MIMIC | MIMIC label |
| **Demographic** | | | |
| - | | | |
| **Vital signs** | | | |
| - | | | |
| **Laboratory results** | | | |
| $SaO_2$ | Dynamic | Y | Arterial $O_2$ Saturation |
| $PaCO_2$ | Dynamic | Y | Arterial $CO_2$ Pressure |
| $PvCO_2$ | Dynamic | Y | Venous $CO_2$ Pressure |
| pHv | Dynamic | Y | PH (Venous) |
| Base excess | Dynamic | Y | Arterial Base Excess |
| $SvO_2$ | Dynamic | Y | Mixed Venous $O_2$ % Sat |
| $PvO_2$ | Dynamic | Y | Venous $O_2$ Pressure |
| DPG | Dynamic | N | - |
| Methemoglobin | Dynamic | N | - |
| Carboxyhemoglobin | Dynamic | N | - |
| **Ventilator features** | | | |
| Fraction of end tidal oxygen | Dynamic | N | - |
| Dead space ventilation | Dynamic | Y | Vd/Vt Ratio |
| Cardiac output | Dynamic | Y | Cardiac Output (CCO) |
| Carbon dioxide production | Dynamic | Y | $CO_2$ production |
| Energy expenditure | Dynamic | Y | Resting Energy Expenditure |
| Compliance | Dynamic | Y | Compliance |
| Alveolar ventilation | Dynamic | N | - |

Table B.6: Baseline characteristics of static and dynamic features for the extracted patient cohort categorised by extubation outcome prior to pre-processing. Unless otherwise stated, the mean and standard deviation are calculated for all features. Categorical features of gender, age and ethnicity are represented by the count and Charlson score, GCS, RAS and Ventilator Mode are defined by the mode. Where values are 'na' reflect outlier compromisation that was dealt with in later pre-processing. The p-value is calculated between success and failure cohorts where relevant

| Feature | Total Patients (n = 4,701) mean ± SD or count | Extubation Success (n = 3,157) mean ± SD or count | Extubation Failure (n = 1,544) mean ± SD | P-value |
|---|---|---|---|---|
| Weight (kg) | 83.56 ± 24.20 | 83.29 ± 23.61 | 84.13 ± 25.36 | 0.259 |
| Height (cm) | 169.20 ± 11.88 | 169.34 ± 12.11 | 168.91 ± 11.40 | 0.238 |
| BMI | na | na | 29.81 ± 12.35 | - |
| Male | 2728 | 1839 | 889 | - |
| Female | 1973 | 1318 | 655 | - |
| Age ≤44 | 612 | 459 | 153 | - |
| Age 45-54 | 685 | 464 | 221 | - |
| Age 55-64 | 1077 | 741 | 336 | - |
| Age 65-74 | 1148 | 750 | 398 | - |
| Age ≥75 | 1179 | 743 | 436 | - |
| Ethnicity (Asian) | 47 | 33 | 14 | - |
| Ethnicity (Black) | 483 | 367 | 116 | - |
| Ethnicity (Hispanic) | 126 | 91 | 35 | - |
| Ethnicity (White) | 2859 | 1894 | 965 | - |
| Ethnicity (Other) | 1186 | 772 | 414 | - |
| Charlson Score (Mode) | 2 | 2 | 2 | - |
| Inspired $O_2$ Fraction | 43.4 ± 11.8 | 42.4 ± 10.6 | 45.6 ± 13.8 | <0.001 |
| Tidal Volume (observed) | 474.0 ± 163.9 | 475.3 ± 164.0 | 471.8 ± 163.7 | 0.382 |
| Tidal Volume (spontaneous) | 522.2 ± 1221.6 | 520.7 ± 1071.4 | 525.7 ± 1512.3 | 0.879 |
| Minute Volume | 8.8 ± 9.3 | 8.7 ± 11.1 | 9.1 ± 3.0 | 0.053 |
| Peak Insp. Pressure | 14.1 ± 30.2 | 13.6 ± 36.6 | 15.1 ± 6.9 | 0.051 |
| Mean Airway Pressure | 7.5 ± 3.2 | 7.1 ± 2.7 | 8.2 ± 3.9 | <0.001 |
| EtCO₂ | 37.9 ± 7.6 | 38.2 ± 7.3 | 37.1 ± 8.2 | 0.047 |
| Heart Rate | 85.0 ± 18.2 | 84.1 ± 17.7 | 86.6 ± 19.1 | <0.001 |
| Respiratory Rate | 19.1 ± 14.4 | 18.7 ± 17.0 | 20.0 ± 6.3 | <0.001 |
| GCS - Eye Opening (Mode) | 4 | 4 | 4 | - |
| GCS - Motor Response (Mode) | 6 | 6 | 6 | - |
| $O_2$ saturation pulseoxymetry | 97.5 ± 3.0 | 97.8 ± 2.7 | 97.0 ± 3.5 | <0.001 |
| Richmond-RAS Scale (Mode) | -1 | -1 | -1 | - |
| Ventilator Mode (Mode) | 11 | 11 | 11 | - |
| Arterial Blood Pressure systolic | 123.3 ± 24.4 | 124.3 ± 23.6 | 121.6 ± 25.8 | <0.001 |
| Arterial Blood Pressure diastolic | 60.5 ± 50.8 | 61.3 ± 62.2 | 58.8 ± 14.0 | 0.002 |
| Arterial Blood Pressure mean | 81.5 ± 18.8 | 82.3 ± 18.4 | 80.1 ± 19.5 | <0.001 |
| Temperature Fahrenheit | 98.9 ± 2.0 | 98.9 ± 1.8 | 98.9 ± 2.2 | 0.990 |
| Hematocrit (serum) | 28.7 ± 4.7 | 28.7 ± 4.8 | 28.6 ± 4.6 | 0.736 |
| Sodium (serum) | 140.3 ± 5.8 | 140.1 ± 5.5 | 140.7 ± 6.5 | 0.094 |
| Potassium (serum) | 4.1 ± 0.6 | 4.0 ± 0.5 | 4.2 ± 0.6 | 0.001 |
| Arterial $O_2$ pressure | 112.7 ± 38.2 | 114.8 ± 36.4 | 108.7 ± 41.2 | <0.001 |
| Arterial $CO_2$ Pressure | 41.0 ± 8.7 | 40.5 ± 8.2 | 42.0 ± 9.6 | <0.001 |
| PH (Arterial) | 7.4 ± 0.1 | 7.4 ± 0.1 | 7.4 ± 0.1 | <0.001 |
| Arterial Base Excess | 406.9 ± na | 612.8 ± na | 0.9 ± 5.7 | 0.476 |
| Arterial $O_2$ Saturation | 96.4 ± 2.5 | 96.5 ± 2.6 | 96.1 ± 2.3 | 0.014 |
| Ionized Calcium | 1.1 ± 0.1 | 1.1 ± 0.1 | 1.1 ± 0.1 | 0.021 |
| Lactic Acid | 2.0 ± 2.5 | 1.6 ± 1.8 | 2.7 ± 3.3 | <0.001 |
| Hemoglobin | 9.6 ± 1.7 | 9.6 ± 1.7 | 9.5 ± 1.6 | 0.384 |
| WBC | 12.0 ± 6.9 | 11.7 ± 6.9 | 12.8 ± 6.8 | 0.015 |
| Creatinine (serum) | 1.5 ± 1.4 | 1.5 ± 1.4 | 1.6 ± 1.3 | 0.088 |
| Glucose (serum) | 137.8 ± 49.7 | 135.9 ± 47.1 | 142.1 ± 55.1 | 0.040 |
| Platelet Count | 168.7 ± 122.2 | 165.1 ± 114.8 | 178.3 ± 139.8 | 0.110 |
| $CO_2$ production | 186.1 ± 61.0 | 190.5 ± 66.8 | 177.9 ± 47.3 | 0.028 |
| Compliance | 84.0 ± 46.0 | 90.3 ± 46.2 | 73.8 ± 44.0 | <0.001 |
| Plateau Pressure | 18.4 ± 5.1 | 17.3 ± 4.5 | 20.0 ± 5.4 | <0.001 |
| Mixed Venous $O_2$ % Sat | 64.3 ± 9.4 | 63.9 ± 8.8 | 65.0 ± 10.4 | 0.486 |
| PH (Venous) | 7.4 ± 0.1 | 7.4 ± 0.1 | 7.4 ± 0.1 | 0.068 |
| Venous $CO_2$ Pressure | 48.8 ± 10.5 | 47.4 ± 9.4 | 51.2 ± 11.9 | 0.022 |
| Venous $O_2$ Pressure | 54.3 ± 30.9 | 56.7 ± 33.3 | 50.3 ± 25.8 | 0.180 |
| Cardiac Output (CCO) | 5.4 ± 1.7 | 5.4 ± 1.7 | 5.5 ± 1.6 | 0.356 |
| Total Bilirubin | 3.2 ± 5.5 | 3.1 ± 4.9 | 3.6 ± 6.7 | 0.418 |
| Negative Insp. Force | -34.1 ± 10.3 | -36.2 ± 9.6 | -31.7 ± 10.7 | 0.173 |

Table B.7: Average feature sampling frequency for Feature Set 2 in the training set. Where features are greyed out reflects their removal having fallen beneath the threshold of 0.5

| Feature | Average Train Set Sampling Frequency |
|---|---|
| Heart Rate | 6.648 |
| Respiratory Rate | 6.615 |
| $O_2$ saturation pulseoxymetry | 6.611 |
| Arterial Blood Pressure mean | 3.761 |
| Arterial Blood Pressure diastolic | 3.754 |
| Arterial Blood Pressure systolic | 3.754 |
| Inspired $O_2$ Fraction | 2.104 |
| GCS - Eye Opening | 1.633 |
| GCS - Motor Response | 1.628 |
| Tidal Volume (observed) | 1.600 |
| Minute Volume | 1.598 |
| Mean Airway Pressure | 1.568 |
| Peak Insp. Pressure | 1.524 |
| Temperature Fahrenheit | 1.446 |
| Ventilator Mode | 1.374 |
| Tidal Volume (spontaneous) | 1.368 |
| Richmond-RAS Scale | 1.279 |
| PH (Arterial) | 0.535 |
| Arterial $CO_2$ Pressure | 0.525 |
| Arterial $O_2$ pressure | 0.525 |
| Sodium (serum) | 0.307 |
| Potassium (serum) | 0.306 |
| Glucose (serum) | 0.271 |
| Creatinine (serum) | 0.269 |
| Hematocrit (serum) | 0.265 |
| Ionized Calcium | 0.251 |
| Hemoglobin | 0.247 |
| Platelet Count | 0.231 |
| WBC | 0.221 |
| Lactic Acid | 0.193 |
| $EtCO_2$ | 0.189 |
| Plateau Pressure | 0.160 |
| Total Bilirubin | 0.076 |
| Negative Insp. Force | 0.008 |

Table B.8: Average feature sampling frequency for Feature Set 3 in the training set. Where features are greyed out reflects their removal having fallen beneath the threshold of 0.15

| Feature | Average Train Set Sampling Frequency |
|---|---|
| Heart Rate | 6.648 |
| Respiratory Rate | 6.615 |
| $O_2$ saturation pulseoxymetry | 6.611 |
| Arterial Blood Pressure mean | 3.761 |
| Arterial Blood Pressure diastolic | 3.754 |
| Arterial Blood Pressure systolic | 3.754 |
| Inspired $O_2$ Fraction | 2.104 |
| GCS - Eye Opening | 1.633 |
| GCS - Motor Response | 1.628 |
| Tidal Volume (observed) | 1.600 |
| Minute Volume | 1.598 |
| Mean Airway Pressure | 1.568 |
| Peak Insp. Pressure | 1.524 |
| Temperature Fahrenheit | 1.446 |
| Ventilator Mode | 1.374 |
| Tidal Volume (spontaneous) | 1.368 |
| Richmond-RAS Scale | 1.279 |
| PH (Arterial) | 0.535 |
| Arterial $O_2$ pressure | 0.525 |
| Arterial $CO_2$ Pressure | 0.525 |
| Arterial Base Excess | 0.525 |
| Sodium (serum) | 0.307 |
| Potassium (serum) | 0.306 |
| Glucose (serum) | 0.271 |
| Creatinine (serum) | 0.269 |
| Hematocrit (serum) | 0.265 |
| Cardiac Output (CCO) | 0.252 |
| Ionized Calcium | 0.251 |
| Hemoglobin | 0.247 |
| Platelet Count | 0.231 |
| WBC | 0.221 |
| Arterial $O_2$ Saturation | 0.198 |
| Lactic Acid | 0.193 |
| $EtCO_2$ | 0.189 |
| Plateau Pressure | 0.160 |
| Compliance | 0.143 |
| $CO_2$ production | 0.106 |
| Total Bilirubin | 0.076 |
| PH (Venous) | 0.045 |
| Venous $O_2$ Pressure | 0.039 |
| Mixed Venous $O_2$ % Sat | 0.039 |
| Venous $CO_2$ Pressure | 0.038 |
| Negative Insp. Force | 0.008 |

Table B.9: Mean and standard deviation for all extracted dynamic train and test set features. Features shown are from Feature Set 3, which includes all features from Feature Sets 1 and 2

| Feature | Train mean | Train std | Test mean | Test std |
|---|---|---|---|---|
| Heart Rate | 84.67 | 16.13 | 84.94 | 16.75 |
| Respiratory Rate | 19.09 | 7.17 | 18.86 | 4.68 |
| $O_2$ saturation pulseoxymetry | 97.57 | 2.41 | 97.51 | 2.19 |
| Arterial Blood Pressure mean | 81.81 | 15.15 | 82.31 | 14.18 |
| Arterial Blood Pressure diastolic | 60.50 | 18.25 | 61.31 | 11.80 |
| Arterial Blood Pressure systolic | 123.70 | 19.76 | 123.01 | 20.57 |
| Inspired $O_2$ Fraction | 42.97 | 9.73 | 42.92 | 8.58 |
| GCS - Eye Opening | 3.19 | 0.92 | 3.15 | 0.92 |
| GCS - Motor Response | 5.34 | 1.25 | 5.22 | 1.40 |
| Tidal Volume (observed) | 469.48 | 146.23 | 481.56 | 153.01 |
| Minute Volume | 8.78 | 6.85 | 8.69 | 2.49 |
| Mean Airway Pressure | 7.53 | 2.78 | 7.46 | 2.70 |
| Peak Insp. Pressure | 14.14 | 21.79 | 13.78 | 5.26 |
| Temperature Fahrenheit | 98.80 | 2.51 | 98.83 | 1.01 |
| Ventilator Mode | 18.21 | 12.00 | 18.51 | 12.43 |
| Tidal Volume (spontaneous) | 497.29 | 490.31 | 552.23 | 965.52 |
| Richmond-RAS Scale | -1.10 | 1.48 | -1.14 | 1.56 |
| PH (Arterial) | 7.41 | 0.07 | 7.41 | 0.07 |
| Arterial $O_2$ pressure | 112.83 | 35.24 | 112.95 | 33.84 |
| Arterial $CO_2$ Pressure | 41.15 | 8.49 | 41.23 | 8.81 |
| Arterial Base Excess | 361.94 | 13425.49 | 1.41 | 5.08 |
| Sodium (serum) | 140.19 | 5.52 | 141.15 | 6.50 |
| Potassium (serum) | 4.08 | 0.57 | 4.07 | 0.56 |
| Glucose (serum) | 137.45 | 48.69 | 139.67 | 54.90 |
| Creatinine (serum) | 1.55 | 1.30 | 1.58 | 1.60 |
| Hematocrit (serum) | 28.82 | 4.65 | 28.77 | 5.11 |
| Cardiac Output (CCO) | 5.38 | 1.35 | 5.87 | 2.34 |
| Ionized Calcium | 1.12 | 0.08 | 1.12 | 0.08 |
| Hemoglobin | 9.58 | 1.63 | 9.57 | 1.72 |
| Platelet Count | 171.77 | 124.01 | 168.56 | 118.28 |
| WBC | 12.19 | 7.00 | 11.29 | 6.47 |
| Arterial $O_2$ Saturation | 96.51 | 1.92 | 95.96 | 3.27 |
| Lactic Acid | 1.87 | 2.30 | 2.13 | 2.64 |
| $EtCO_2$ | 37.71 | 7.14 | 37.52 | 7.58 |
| Plateau Pressure | 18.36 | 5.10 | 18.05 | 4.69 |

Table B.10: Bounds used to remove outliers for numerical features. The table lists all features from Feature Set 3, which includes all features from sets 1 and 2. The MIMIC provided lower and upper bounds were used where available; otherwise, the mean $\pm$ 3 standard deviations were used. Where negative values were not meaningful for a feature and the mean $-$ 3 standard deviations was negative, zero was set as the lower bound. The bounds for Arterial Base Excess were set in discussion with Dr Murali, as the imputed bound values were not within a clinically meaningful range

| Feature | Lower bound | Upper bound |
| --- | --- | --- |
| Inspired $O_2$ Fraction | 7.678 | 79.189 |
| Tidal Volume (observed) | 299.000 | 750.000 |
| Tidal Volume (spontaneous) | 299.000 | 750.000 |
| Minute Volume | 0.000 | 12.100 |
| Peak Insp. Pressure | 0.000 | 115.092 |
| Mean Airway Pressure | 0.000 | 17.102 |
| EtCO$_2$ | 17.000 | 59.109 |
| Heart Rate | 30.659 | 139.152 |
| Respiratory Rate | 0.000 | 66.676 |
| O$_2$ saturation pulseoxymetry | 88.535 | 100.000 |
| Arterial Blood Pressure systolic | 90.000 | 140.000 |
| Arterial Blood Pressure diastolic | 60.000 | 90.000 |
| Arterial Blood Pressure mean | 25.026 | 137.893 |
| Temperature Fahrenheit | 92.586 | 105.190 |
| Hematocrit (serum) | 15.200 | 42.701 |
| Sodium (serum) | 123.147 | 157.080 |
| Potassium (serum) | 2.700 | 5.797 |
| Arterial O$_2$ pressure | 16.000 | 227.632 |
| Arterial CO$_2$ Pressure | 16.000 | 66.916 |
| PH (Arterial) | 7.190 | 7.580 |
| Arterial Base Excess | -10.000 | 10.000 |
| Arterial O$_2$ Saturation | 90.388 | 100.000 |
| Ionized Calcium | 0.884 | 1.354 |
| Lactic Acid | 0.500 | 8.909 |
| Hemoglobin | 5.100 | 14.485 |
| WBC | 0.100 | 33.118 |
| Creatinine (serum) | 0.200 | 5.413 |
| Glucose (serum) | 23.000 | 282.720 |
| Platelet Count | 6.000 | 539.929 |
| Plateau Pressure | 2.943 | 31.000 |
| Cardiac Output (CCO) | 4.000 | 8.000 |

| Paper | Database | Objective | Inclusion criteria | Exclusion criteria | Extubation failure definition |
|---|---|---|---|---|---|
| Chen et al. | MIMIC-III | Develop a LightGBM model to predict extubation failure | • N/A | • Patients who did not intubate and had unclear extubation states<br>• Age < 18<br>• Admissions that the patient was dead before extubation | Re-intubation within 48 hours of extubation |
| Zhao et al. | MIMIC-IV | Develop and validate an accurate machine-learning model to predict EF in intensive care units (ICUs) | • Filter patients with ICU stays with an extubation code | • Age < 18 years<br>• Unplanned extubation<br>• Not the first extubation during the hospital stay<br>• No MV records before extubation | The need for ventilatory support (NIV or re-intubation) or death within 48 hours following planned extubation |
| Jia et al. | MIMIC-III | Employ Convolutional Neural Networks (CNNs) to predict extubation readiness at each patient state, with a 1 h time step | • Admissions who underwent invasive mechanical ventilation | • Age < 18<br>• Not successfully discharged i.e. died in hospital (fatalities can be caused by factors beyond the weaning process<br>• If patient record had missing values after processing<br>• Patients who had ventilation support for less than 8 hours<br>• Patients who experienced extubation failure (as predicting readiness) | N/A |
| Prasad et al. | MIMIC-III | To develop a decision support tool that leverages available patient information in the data-rich ICU setting to alert clinicians when a patient is ready for initiation of weaning, and to recommend a personalized treatment protocol | • Patients undergoing IMV | • Not adult<br>• Exclude those under ventilator support for less than 24 hours<br>• Admissions where patient not successfully discharged by the end of admission | N/A |
| Liu et al. | MIMIC-IV | To develop a practical model to predict weaning in patients with sepsis | • Patients with invasive ventilation | • Age < 18<br>• Repeated ICU admission | The definition of weaning success (WS) was as follows: (a) no intubation or invasive ventilation within 48 h after weaning, (b) no death within 48 h after weaning, and (c) non-invasive ventilation time was shorter than 48 h after weaning |
| Kim et al. | MIMIC-IV | To develop machine learning models that predict the probability of successful weaning within 14 days of intubation | • Patients with "Intubation" and "Invasive ventilation" codes appearing at least once in the "procedure event" or "chart event"<br>• Patients with "Ventilator type" and "Ventilator mode" codes appearing 5 times or more within 24 hours after the first code | • Aged <18 or >100 years,<br>• Previous tracheostomy, and<br>• Missing Sequential Organ Failure Assessment (SOFA) score and Simplified Acute Physiology Score II (SAPS II) | MV discontinuation without death |
| Mikhno et al. | MIMIC-III | Present machine learning techniques used to locate features relevant to EF, and to develop a model for predicting extubation failure or success 2 hours prior to the extubation attempt in neonates | • Patients were 23 to 31 weeks of gestational age<br>• Intubated within 24 hours of birth in the same hospital<br>• The duration of intubation was at least one day<br>• Assigned ICD-9 code 769, indicating a diagnosis of Respiratory Distress Syndrome | • Nursing notes<br>• Unclear extubation events<br>• Self-extubation by infant | Any patient that was re-intubated within 48 hours of the first extubation attempt |

Table B.11: Summary of inclusion and exclusion criteria used for select extubation-related studies that used the MIMIC database. Definitions of extubation failure are included where relevant. Studies outlined are Chen [50], Zhao [49], Jia [146], Prasad [147], Liu [148], Kim [149] and Mikhno [150].

| Features | Kuo | Hsieh | Chen | Fabregat | Otaguro | Zhao | Fleuren | Lee | Pai | Jia | Xia | Kim | Huang | Zeng | Total occurrences |
|---|---|---|---|---|---|---|---|---|---|---|---|---|---|---|---|
| **Demographic** | | | | | | | | | | | | | | | |
| Age | X | X | X | X | X | X | X | X | | X | | X | | X | 11 |
| Gender | | X | X | X | X | X | X | X | | X | X | X | | X | 11 |
| Body mass index | | X | | X | X | X | X | | | | | | | X | 6 |
| Weight | | | X | | X | | | | | X | | X | | X | 5 |
| Height | | | X | | X | | | X | | | | X | | X | 5 |
| Ethnicity | | | | | | X | | | | X | | X | | X | 4 |
| Past medical history | | X | X | | | X | X | | | | | | | | 4 |
| Reasons for respiratory failure | X | | | | | | | | | X | | X | | X | 4 |
| Weight loss | | | X | | | | | | | | | | | | 1 |
| Charlson index | | | | | | X | | | | | | | | | 1 |
| **Vital Signs** | | | | | | | | | | | | | | | |
| Heart rate | | X | X | X | X | X | X | X | X | X | X | | | X | 11 |
| Respiratory rate | | X | X | X | X | X | X | X | | X | X | | X | X | 11 |
| Glasgow coma scale | | | X | X | X | X | X | X | X | | | X | | X | 9 |
| Mean arterial pressure | | X | X | | X | X | | X | | X | | | | X | 7 |
| Body temperature | | | X | | X | X | | | | | | X | | X | 5 |
| Richmond agitation-sedation scale | | | X | | | | X | | X | X | | | | X | 5 |
| SpO2 | | | X | | | X | | X | | | X | | | X | 5 |
| Systolic blood pressure | | | X | | X | | | | | X | | | | X | 4 |
| Diastolic blood pressure | | | X | | X | | | | | X | | | | X | 4 |
| O2 saturation to inspired fraction ratio | | | | X | | X | | | | | | | | | 2 |
| End-tidal carbon dioxide | | | | | | X | | | | | | | | | 1 |
| Number of premature ventricular contraction | | | | | | X | | | | | | | | | 1 |
| **Laboratory results** | | | | | | | | | | | | | | | |
| PaCO2 | | X | X | | X | X | X | X | | X | X | | | X | 9 |
| Arterial pH | | X | X | | X | X | | | | X | X | X | | X | 8 |
| White blood cell | | | X | | X | X | X | | | | X | X | | X | 7 |
| PaO2 | | X | X | | X | X | | | | X | X | | | X | 7 |
| Creatinine | | X | X | | X | X | X | | | | | X | | X | 7 |
| Hemoglobin | | X | X | | X | X | | | | | | X | | X | 6 |
| Platelet | | | X | | X | X | X | | | | | X | | X | 6 |
| Na+ | | X | X | | X | X | | | | | | X | | X | 6 |
| K+ | | X | X | | X | X | | | | | | X | | X | 6 |
| Lactate | | | X | | X | X | | | | | X | X | | X | 6 |
| Glucose | | X | X | | | X | X | | | | X | | | X | 6 |

| Features | Kuo | Hsieh | Chen | Fabregat | Otaguro | Zhao | Fleuren | Lee | Pai | Jia | Xia | Kim | Huang | Zeng | Total occurrences |
|---|---|---|---|---|---|---|---|---|---|---|---|---|---|---|---|
| **Laboratory results** | | | | | | | | | | | | | | | |
| Total bilirubin | | X | X | | X | X | | | | | | X | | X | **6** |
| Hematocrit | | X | | | X | X | X | | | | | | | X | **5** |
| P/F ratio | | X | | | X | X | X | | | | | | | X | **5** |
| Ca+ | | X | X | | X | X | | | | | | | | X | **5** |
| Cl- | | X | | | X | X | | | | | | | | X | **4** |
| Anion gap | | | | | X | X | | | | | | X | | X | **4** |
| Creatine phosphokinase | | | X | | X | X | | | | | | | | X | **4** |
| Amylase | | | X | | | X | | | | | | X | | X | **4** |
| Prothrombin time | | | X | | X | X | | | | | | | | X | **4** |
| Activated partial thromboplastin time | | | X | | X | X | | | | | | | | X | **4** |
| Red blood cell | | | | | X | X | | | | | | X | | | **3** |
| SaO2 | | | X | | X | X | | | | | | | | | **3** |
| Blood urea nitrogen (BUN) | | | X | | | | | | | | | X | | X | **3** |
| Troponin | | | X | | X | | | | | | | | | X | **3** |
| B-type natriuretic peptide | | | X | | X | | X | | | | | | | | **3** |
| Aspartate aminotransferase | | | X | | X | X | | | | | | | | | **3** |
| Albumin | | | | | X | | | | | | | X | | X | **3** |
| Base excess | | | | | X | X | | | | | | | | | **2** |
| HCO3- | | | | | X | X | | | | | | | | | **2** |
| Methemoglobin | | | | | X | X | | | | | | | | | **2** |
| Alveolar-arterial oxygen gradient | | | X | | | | | | | | | | | X | **2** |
| Central venous oxygen saturation | | | X | | | X | | | | | | | | | **2** |
| Total protein | | | X | | | | | | | | | | | X | **2** |
| C-reactive protein | | | X | | X | | | | | | | | | | **2** |
| Alanine transaminase | | | | | X | X | | | | | | | | | **2** |
| Lactate dehydrogenase (LDH) | | | | | X | X | | | | | | | | | **2** |
| PT/INR | | | | | X | X | | | | | | | | | **2** |
| Bicarbonate | | | | | | | | | | | | X | | X | **2** |
| Mean corpuscular volume | | | | | | X | | | | | | | | | **1** |
| Mean corpuscular hemoglobin | | | | | | X | | | | | | | | | **1** |
| Red cell distribution width | | | | | | X | | | | | | | | | **1** |
| P+ | | X | | | | | | | | | | | | | **1** |
| Carboxyhemoglobin | | | | | X | | | | | | | | | | **1** |
| Alkaline phosphatase | | | | | X | | | | | | | | | | **1** |
| Fibrinogen | | | | | | | | | | | | | | X | **1** |
| Neutrolphil or lymphocyte ratio | | | | | | | | | | | | X | | | **1** |
| Globulin | | | | | | | | | | | | | | X | **1** |
| Indirect/Direct bilirubin | | | | | | | | | | | | | | X | **1** |
| INR (International Normalised Ratio) | | | | | | | | | | | | | | X | **1** |
| D-Dimer | | | | | | | | | | | | | | X | **1** |

| Features | Kuo | Hsieh | Chen | Fabregat | Otaguro | Zhao | Fleuren | Lee | Pai | Jia | Xia | Kim | Huang | Zeng | Total occurrences |
|---|---|---|---|---|---|---|---|---|---|---|---|---|---|---|---|
| **Ventilator information** | | | | | | | | | | | | | | | |
| Time under mechanical ventilation (Tmv) | X | X | X | X | X | X | X | | | | X | | | X | 9 |
| Fraction of inspired oxygen | | X | | | X | X | X | X | | X | | | X | X | 8 |
| Tidal volume | X | | X | | X | X | | X | | X | | | X | X | 8 |
| Peak inspiration pressure (Pmax) | | | X | X | X | | X | | | X | | | X | X | 7 |
| PEEP | | X | | | | X | X | | | X | X | | X | X | 7 |
| Mean airway pressure | | | X | | X | X | | | | X | X | | X | | 6 |
| Minute volume | | X | X | | X | | | X | | | | | | X | 5 |
| Plateau pressure | | | | X | | X | | | | X | | | | X | 4 |
| Positive end-expiratory pressure | | X | | | X | X | X | | | | | | | | 4 |
| Spontaneous breathing trial success times | | | | | | X | | | | X | | | | X | 3 |
| Ventilation mode | | | | X | | | | | | X | | | | | 2 |
| Hours since last controlled mode | | | | | | | X | | | | | | | | 1 |
| Tidal volume per kg ideal body weight | | | | | | | X | | | | | | | | 1 |
| Maximum inspiratory pressure | | X | | | | | | | | | | | | | 1 |
| Maximum expiratory pressure | | X | | | | | | | | | | | | | 1 |
| Airway occlusion pressure | | | | | | | X | | | | | | | | 1 |
| Ventilatory ratio | | | | | | | X | | | | | | | | 1 |
| Inspiratory time | X | | | | | | | | | | | | | | 1 |
| Expiratory time | X | | | | | | | | | | | | | | 1 |
| O2 Flow | | | | | | | | | | X | | | | | 1 |
| Pressure support ventilation level | | | | | | | | | | | X | | | | 1 |
| Inspiratory flow rate | | | | | | | | | | | | | | X | 1 |
| **Other** | | | | | | | | | | | | | | | |
| Rapid shallow breathing index | | X | X | X | | X | | | X | | | | | | 5 |
| Sedatives and analgesics dose | | | | X | | | X | | | | | | | X | 3 |
| Vasopressor | | | X | | | X | | | | | X | | | | 3 |
| Urine output | | | | | | X | | | X | | | | | X | 3 |
| Fluid balance | | | | | | | X | | X | | | | | | 2 |
| Crystalloid and colloid amount | | | | | | X | | | | | | | | X | 2 |
| Sequential organ failure assessment | | | X | | | | | | | | X | | | | 2 |
| Acute physiology and chronic health evaluation-II | | | | | X | | X | | | | | | | | 2 |
| SEMICYUC code | | | | | X | X | | | | | | | | | 2 |
| Hospital stay | X | | | | | | | | | | | | | | 1 |
| ICU stay | X | | | | | | | | | | | | | | 1 |
| Sedation day | | | X | | | | | | | | | | | | 1 |
| Total cumulative dose (sedatives and analgesics) | | | | | | | X | | | | | | | | 1 |
| Antibiotic type (ABX) | | | | | | X | | | | | | | | | 1 |
| Continuous renal replacement therapy | | | | | | X | | | | | | | | | 1 |
| Transfusion (RBC, FFP, PLT) | | | | | | X | | | | | | | | | 1 |
| Hours since last proning session | | | | | | X | | | | | | | | | 1 |
| Central venous pressure | | | | | | X | | | | | | | | | 1 |
| Simplified acute physiology score | | | | | | | | | X | | | | | | 1 |
| OASIS | | | | | | | | | X | | | | | | 1 |
| COPD diagnosis | | | | | | | X | | | | | | | | 1 |

Table B.12: Analysis of features used in extubation-related literature. Across all studies, where a feature was present in the set used to train the model, an 'X' is present. The total occurrence for each feature is then summed across all studies to evaluate popularity. Studies shown are: Kuo [44], Hsieh [45], Chen [50], Fabregat [54], Otaguro [42], Zhao [49], Fleuren [56], Lee [151], Pai [152], Jia [146], Xia [153], Kim [149], Huang [55] and Zeng [80]

| Category | Sub-categories | Feature | Data type | Units | Indicates |
|---|---|---|---|---|---|
| **Patient and Unit Identifiers** | | UnitId | Categorical | - | ID of ventilator unit |
| | | PatientCardId | Categorical | - | Patient ID |
| | | Ptype | Categorical | - | Type of Patient |
| **Ventilation Mode and Settings** | Ventilation Mode | VentilationMode | Categorical | - | Type of ventilation being administered |
| | | VentModeSet | Categorical | - | Mode settings on ventilator |
| | Oxygen Concentration | FiO2Set | Numerical | % | Fraction of inspired oxygen set on the ventilator |
| | Tidal Volume | VtSet | Numerical | mL | Tidal volume set on the ventilator |
| | Pressure Settings | PcSet | Numerical | cmH20 | Pressure control setting on the ventilator |
| | | PsSet | Numerical | cmH20 | Pressure support setting on the ventilator |
| | | PEEPSet | Numerical | cmH20 | Positive end-expiratory pressure set on the ventilator |
| | Respiratory Rate | RfSet | Numerical | bpm | Respiratory rate set on the ventilator |
| | I:E Ratio | IESet | Numerical | Ratio | Inspiratory to expiratory ratio set on the ventilator |
| **Blood Measurements** | Arterial Oxygenation | SaO2 | Numerical | % | Arterial oxygen saturation |
| | | PaO2 | Numerical | kPa | Partial pressure of oxygen in arterial blood |
| | Ventilation Effectiveness | PaCO2 | Numerical | kPa | Partial pressure of carbon dioxide in arterial blood |
| | | PvCO2 | Numerical | kPa | Partial pressure of carbon dioxide in venous blood |
| | Acid-Base Balance | pHa | Numerical | pH | Arterial pH |
| | | pHv | Numerical | pH | Venous pH |
| | | be | Numerical | mmol/L | Base excess in the blood |
| | Oxygen Delivery and Utilisation | SvO2 | Numerical | % | Venous oxygen saturation |
| | | PvO2 | Numerical | mmHg | Partial pressure of oxygen in venous blood |
| | | Hb | Numerical | g/dL | Hemoglobin concentration |
| | | DPG | Numerical | mmol/L | Diphosphoglycerate level |
| | Methemoglobin and Carboxyhemoglobin | FMetHb | Numerical | % | Fraction of methemoglobin |
| | | FCOHb | Numerical | % | Fraction of carboxyhemoglobin |
| **Respiratory Mechanics** | Tidal Volume and Respiratory Rate | Vtac | Numerical | mL | Actual tidal volume |
| | | Vt | Numerical | mL | Tidal volume |
| | | Rf | Numerical | bpm | Respiratory rate |
| | Gas Exchange Efficiency | SpO2 | Numerical | % | Peripheral oxygen saturation |
| | | FetO2 | Numerical | % | Fraction of end-tidal oxygen |
| | | FetCO2 | Numerical | % | Fraction of end-tidal carbon dioxide |
| | | Vd | Numerical | mL | Dead space ventilation - air not participating in gas exchange |
| | Metabolic and Cardiac Function | CO | Numerical | L/min | Cardiac output - amount of blood pumped |
| | | VO2 | Numerical | mL/min | Oxygen consumption |
| | | VCO2 | Numerical | mL/min | Carbon dioxide production |
| | | EE | Numerical | kcal/day | Energy expenditure |
| | Lung Mechanics | Pplat | Numerical | cmH2O | Plateau pressure - pressure applied to small airways and alveoli |
| | | Ppeak_meas | Numerical | cmH20 | Measured peak inspiratory pressure - max pressure during inhalation |
| | | Pplat_VC_meas | Numerical | cmH20 | Measured plateau pressure in volume control mode |
| | | Comp | Numerical | mL/cmH20 | Compliance |
| | Indicies and Ratios | RSBI | Numerical | Ratio (bpm/L) | Rapid shallow breathing index - respiratory rate / tidal volume |
| | | RSBIWA | Numerical | Ratio (bpm/L) | Rapid shallow breathing index weighted average |
| | | VA | Numerical | mL/min | Alveolar ventilation |
| | | MV | Numerical | L/min | Minute ventilation |
| **Weaning Indicators and Trial Status** | Spontaneous Breathing Trials | SBTRunning | Binary | - | Indicate if spontaneous breathing trial is running |
| | | SBTPassed | Binary | - | Indicate if spontaneous breathing trial was passed |
| | Intubation Status | Intubated | Binary | - | Whether the patient is intubated |

Table B.13: Summary of features available in the WAVE study [154]

# C

# Appendix: Model Development

Table C.1: Optimal hyperparameters derived from Random Searches across 50 iterations per run

| Random Search | Optimal parameters |
|---|---|
| Run 1 | 'weight_decay': 0.0001, 'sampling_method': 'oversample', 'num_epochs': 20, 'loss': 'normal', 'learning_rate': 0.0005, 'layer_dim': 1, 'hidden_dim': 32, 'dropout_prob': 0.25, 'batch_size': 32 |
| Run 2 | 'weight_decay': 0.001, 'sampling_method': 'undersample', 'num_epochs': 20, 'loss': 'normal', 'learning_rate': 0.001, 'layer_dim': 2, 'hidden_dim': 32, 'dropout_prob': 0.25, 'batch_size': 64 |
| Run 3 | 'weight_decay': 0.001, 'sampling_method': 'undersample', 'num_epochs': 20, 'loss': 'normal', 'learning_rate': 0.001, 'layer_dim': 2, 'hidden_dim': 32, 'dropout_prob': 0.25, 'batch_size': 64 |

Table C.2: Subset groupings for Feature Set 2. Features are split into low, medium and high subsets based on their average observation frequency. Low Frequency Subset (average frequency $< 1$ in 6-hour window). Medium Frequency Subset ($1 <$ average frequency $< 3$ in 6-hour window). High Frequency Subset (average frequency $> 3$ in 6-hour window)

| Subset | Features |
|---|---|
| Low frequency | PH (Arterial), Arterial $O_2$ pressure, Arterial $CO_2$ Pressure |
| Medium frequency | Inspired $O_2$ Fraction, GCS - Eye Opening, GCS - Motor Response, Tidal Volume (observed), Minute Volume, Mean Airway Pressure, Peak Insp. Pressure, Temperature Fahrenheit, Tidal Volume (spontaneous), Richmond-RAS Scale |
| High frequency | Heart Rate, $O_2$ saturation pulseoxymetry, Respiratory Rate, Arterial Blood Pressure mean, Arterial Blood Pressure diastolic, Arterial Blood Pressure systolic |

Table C.3: Subset groupings for Feature Set 3. Features are split into low, medium and high subsets based on their average observation frequency. Low Frequency Subset (average frequency $< 1$ in 6-hour window). Medium Frequency Subset ($1 <$ average frequency $< 3$ in 6-hour window). High Frequency Subset (average frequency $> 3$ in 6-hour window)

| Subset | Features |
|---|---|
| Low frequency | PH (Arterial), Arterial $O_2$ pressure, Arterial $CO_2$ Pressure, Arterial Base Excess, Sodium (serum), Potassium (serum), Glucose (serum), Creatinine (serum), Hematocrit (serum), Cardiac Output (CCO), Ionized Calcium, Hemoglobin, Platelet Count, WBC, Arterial $O_2$ Saturation, Lactic Acid, $EtCO_2$, Plateau Pressure |
| Medium frequency | Inspired $O_2$ Fraction, GCS - Eye Opening, GCS - Motor Response, Tidal Volume (observed), Minute Volume, Mean Airway Pressure, Peak Insp. Pressure, Temperature Fahrenheit, Tidal Volume (spontaneous), Richmond-RAS Scale |
| High frequency | Heart Rate, $O_2$ saturation pulseoxymetry, Respiratory Rate, Arterial Blood Pressure mean, Arterial Blood Pressure diastolic, Arterial Blood Pressure systolic |

Table C.4: Fused LSTM-FFNN hyperparameter grid

| Hyperparameter | Values |
| --- | --- |
| Hidden dimension | 32, 64, 128, 256, 512 |
| Layer dimensions | 1, 2, 3, 4 |
| Dropout probability | 0.0, 0.25, 0.5, 0.75 |
| Optimiser learning rate | 0.01, 0.001, 0.0001, 0.00001 |
| Batch size | 16, 32, 64, 128, 256 |
| Number of epochs | Range 10 to 100 inclusive in steps of 10 |
| Sampling method | 'normal', 'undersample', 'oversample' |
| Loss type | 'normal', 'weighted' |
| L2 weight decay | 0.01, 0.001, 0.0001, 0.00001 |
| FFNN layers | 1, 2, 3 |
| FFNN units | 32, 64, 128, 256 |
| FFNN activation | 'relu', 'tanh', 'sigmoid', 'leaky_relu' |
| FFNN dropout | 0.0, 0.25, 0.5, 0.75 |

Table C.5: Fused TCN-FFNN hyperparameter grid

| Hyperparameter | Values |
| --- | --- |
| Number of channels | '16,32', '32,64', '64,128', '128,256', '16,32,64', '32,64,128', '64,128,256' |
| Kernel size | Range 2 − 11 |
| Dropout probability | 0.0, 0.25, 0.5, 0.75 |
| Optimiser learning rate | 0.01, 0.001, 0.0001, 0.00001 |
| Batch size | 16, 32, 64, 128, 256 |
| Number of epochs | Range 10 to 100 inclusive in steps of 10 |
| Sampling method | 'normal', 'undersample', 'oversample' |
| Loss type | 'normal', 'weighted' |
| L2 weight decay | 0.01, 0.001, 0.0001, 0.00001 |
| FFNN layers | 1, 2, 3 |
| FFNN units | 32, 64, 128, 256 |
| FFNN activation | 'relu', 'tanh', 'sigmoid', 'leaky_relu' |
| FFNN dropout | 0.0, 0.25, 0.5, 0.75 |

Table C.6: LightGBM hyperparameter grid

| Hyperparameter | Values |
| --- | --- |
| Number of leaves | 20, 30, 40, 50, 60, 70, 80, 90, 100 |
| Max depth | 3, 4, 5, 6, 7, 8, 9, 10 |
| Min data in leaf | 20, 30, 40, 50, 60, 70, 80, 90, 100 |
| Learning rate | 0.1, 0.01, 0.001, 0.0001 |
| Sampling method | 'normal', 'undersample', 'oversample' |
| L2 weight decay | 0.1, 0.01, 0.001, 0.0001 |

# D

## Appendix: Model Results

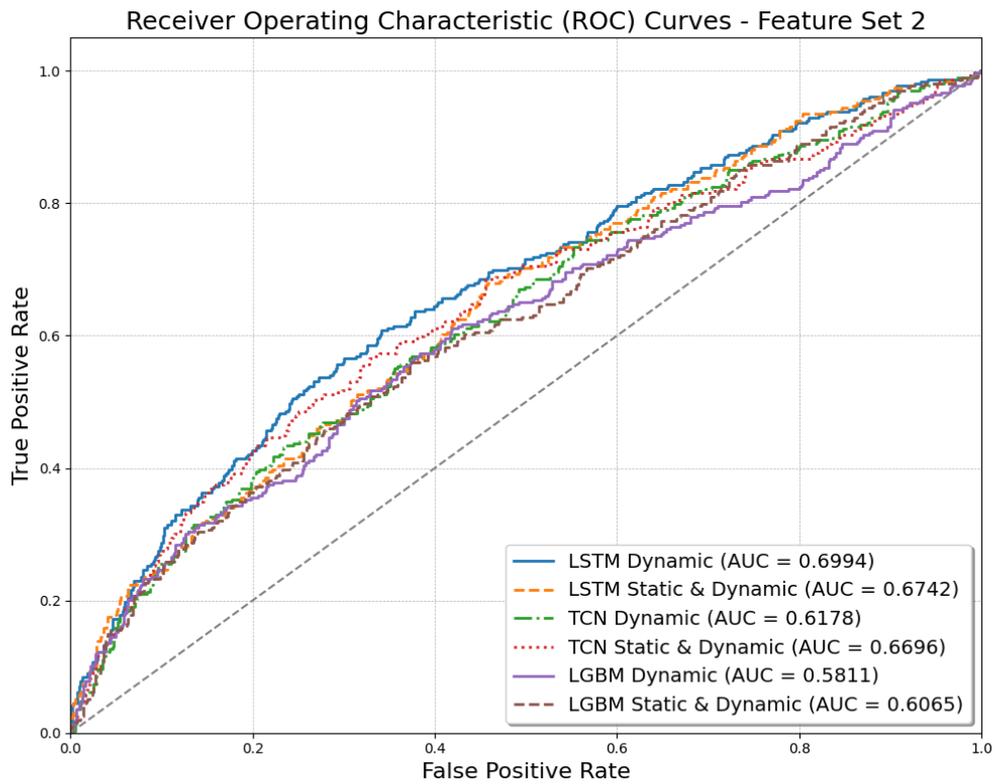

Figure D.1: ROC curves for LSTM, TCN and LightGBM models trained on Feature Set 2

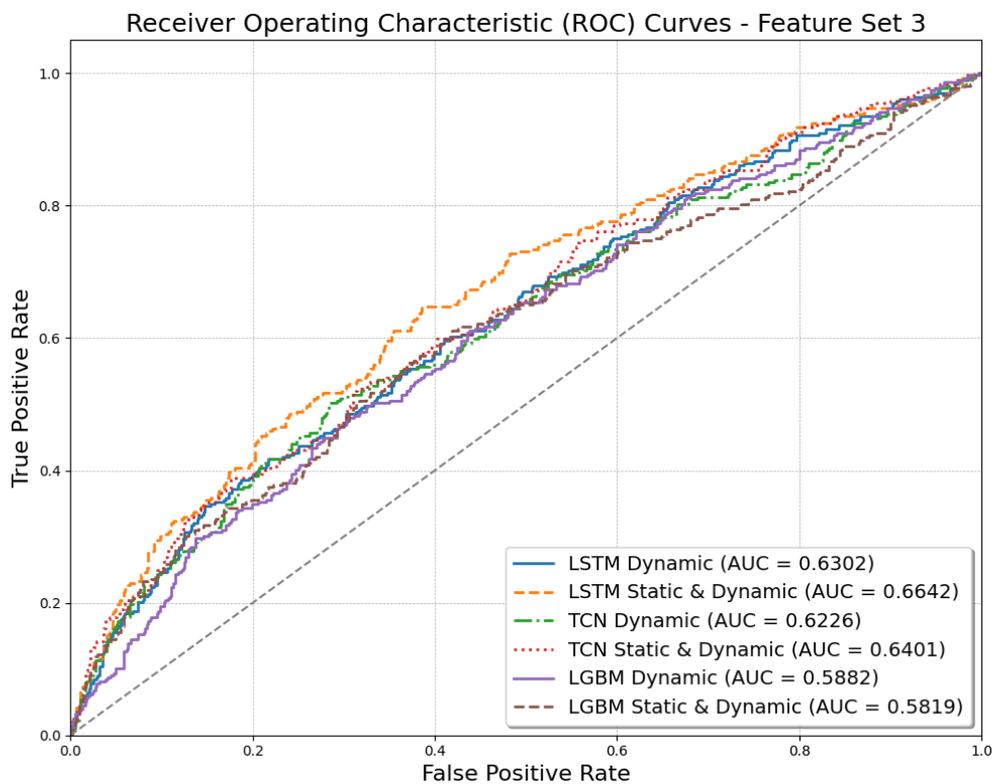

Figure D.2: ROC curves for LSTM, TCN and LightGBM models trained on Feature Set 3

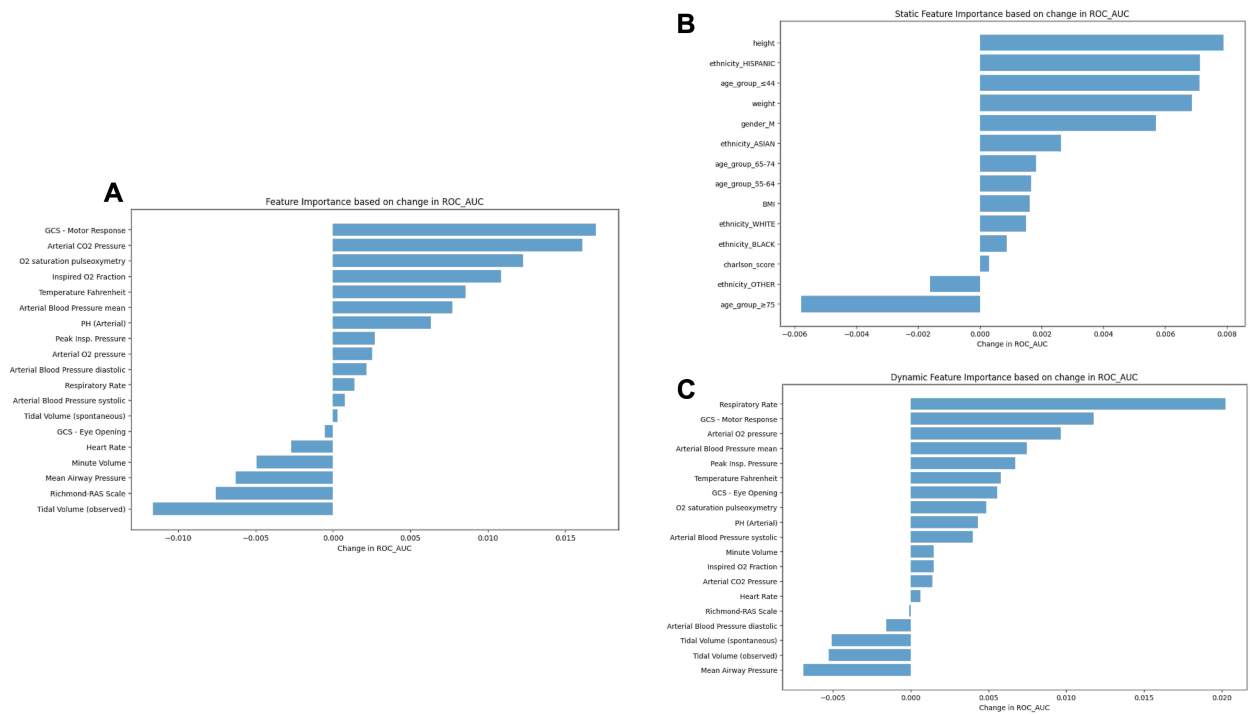

Figure D.3: Feature ablation values for Fused LSTM on dynamic only data (A) and Fused LSTM-FFNN on static (B) and dynamic (C) features from Feature Set 2. The scores represent the change in AUC-ROC when that feature is ablated

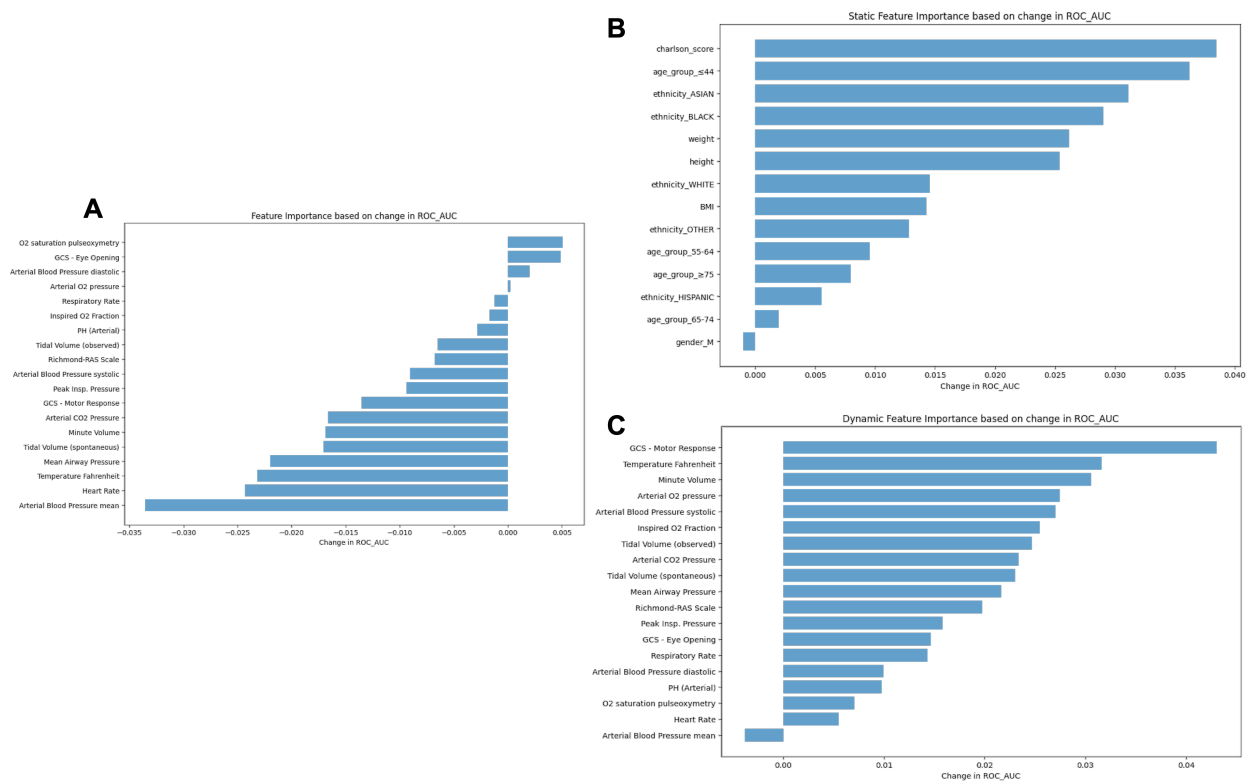

Figure D.4: Feature ablation values for Fused TCN on dynamic only data (A) and Fused TCN-FFNN on static (B) and dynamic (C) features from Feature Set 2. The scores represent the change in AUC-ROC when that feature is ablated

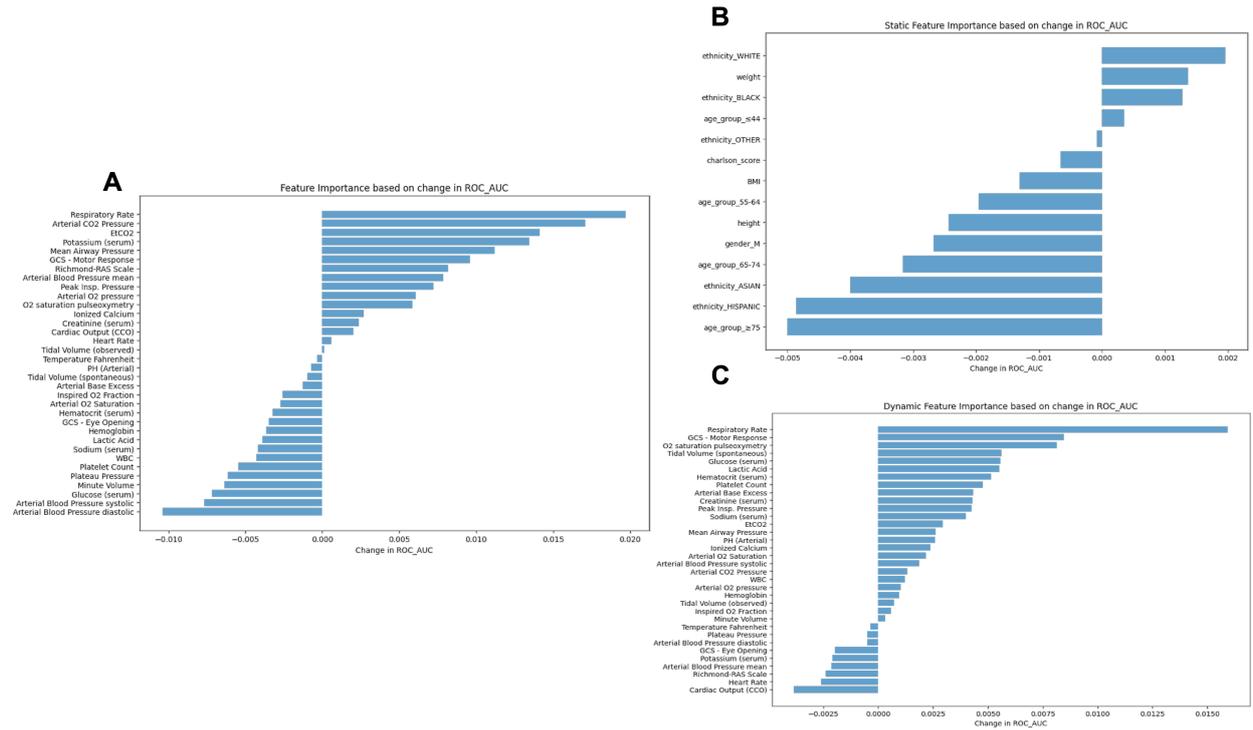

Figure D.5: Feature ablation values for Fused LSTM on dynamic only data (A) and Fused LSTM-FFNN on static (B) and dynamic (C) features from Feature Set 3. The scores represent the change in AUC-ROC when that feature is ablated

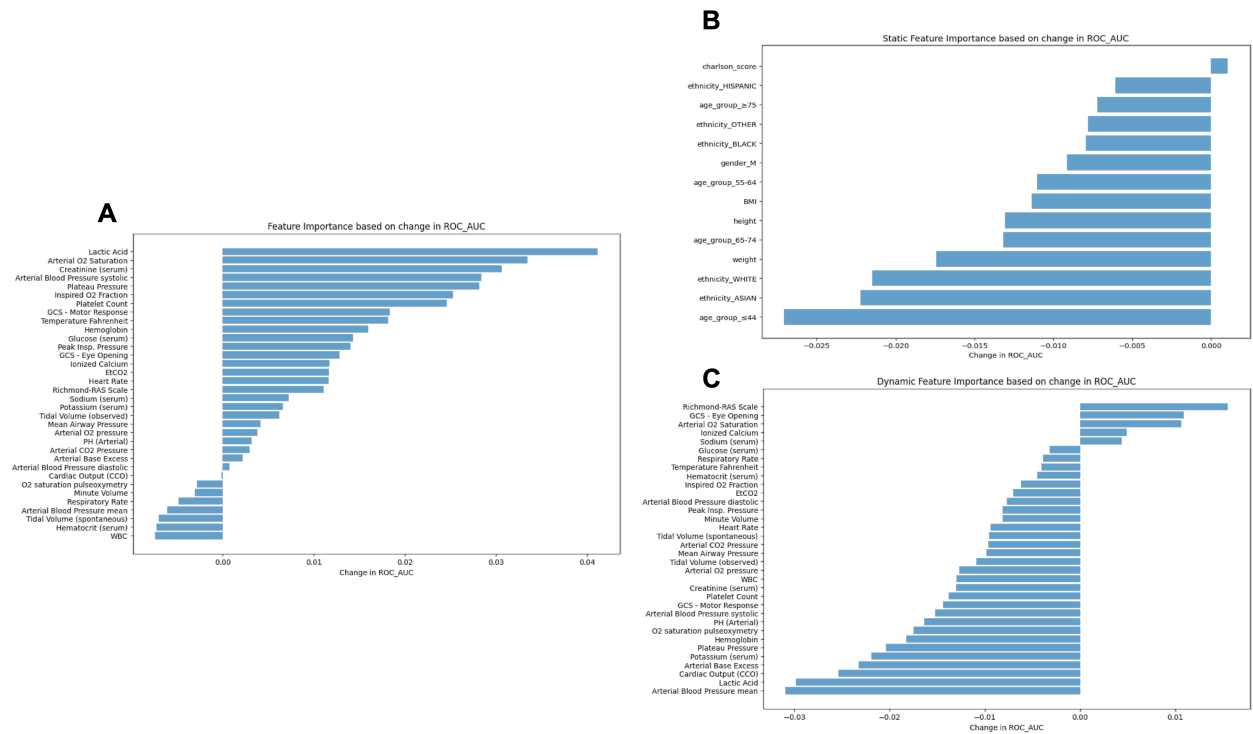

Figure D.6: Feature ablation values for Fused TCN on dynamic only data (A) and Fused TCN-FFNN on static (B) and dynamic (C) features from Feature Set 3. The scores represent the change in AUC-ROC when that feature is ablated

# E
## Appendix: Ethics

Table E.1: Ethics Checklist from Imperial College London

| Ethics checklist | Yes | No |
|---|---|---|
| **Section 1:  Humans** | | |
| Does your project involve human participants? | | No |
| **Section 2:  Protection of personal data** | | |
| Does your project involve personal data collection and/or processing? | | No |
| Does it involve the collection and/or processing of sensitive personal data (e.g. health, sexual lifestyle, ethnicity, political opinion, religious or philosophical conviction)? | | No |
| Does it involve processing of genetic information? | | No |
| Does it involve tracking or observation of participants? It should be noted that this issue is not limited to surveillance or localization data. It also applies to WAN data such as IP address, MACs, cookies etc. | | No |
| Does your project involve further processing of previously collected personal data (secondary use)? For example Does your project involve merging existing data sets? | | No |
| **Section 3:  Animals** | | |
| Does your project involve animals? | | No |
| **Section 4:  Developing Countries** | | |
| Does your project involve developing countries? | | No |
| If your project involves low and/or lower-middle income countries, are any benefit-sharing actions planned? | | No |
| Could the situation in the country put the individuals taking part in the project at risk? | | No |
| **Section 5:  Environmental Protection and Safety** | | |
| Does your project involve the use of elements that may cause harm to the environment, animals or plants? | | No |
| Does your project involve the use of elements that may cause harm to humans, including project staff? | | No |
| **Section 6:  Dual Use** | | |
| Does your project have the potential for military applications? | | No |
| Does your project have an exclusive civilian application focus? | | No |
| Will your project use or produce goods or information that will require export licenses in accordance with legislation on dual-use items? | | No |
| Does your project affect current standards in military ethics – e.g., global ban on weapons of mass destruction, issues of proportionality, discrimination of combatants and accountability in drone and autonomous robotics developments, incendiary or laser weapons? | | No |
| **Section 7:  Misuse** | | |
| Does your project have the potential for malevolent/criminal/terrorist abuse? | | No |
| Does your project involve information on/or the use of biological-, chemical-, nuclear/radiological-security sensitive materials and explosives, and means of their delivery? | | No |
| Does your project involve the development of technologies or the creation of information that could have severe negative impacts on human rights standards (e.g. privacy, stigmatization, discrimination), if misapplied? | | No |
| Does your project have the potential for terrorist or criminal abuse e.g. infrastructural vulnerability studies, cybersecurity related project? | | No |
| **Section 8:  Legal Issues** | | |
| Will your project use or produce software for which there are copyright licensing implications? | | No |
| Will your project use or produce goods or information for which there are data protection, or other legal implications? | | No |
| **Section 9:  Other Ethics Issues** | | |
| Are there any other ethics issues that should be taken into consideration? | | No |